\documentclass[11pt]{article}  %

 \usepackage[letterpaper, left=1in, right=1in, top=1in, bottom=1in]{geometry}

\usepackage[colorlinks=true, linkcolor=blue!70!black, citecolor=blue!70!black]{hyperref}
\usepackage{microtype}

\usepackage{algorithm}

\usepackage[noend]{algpseudocode} 
 
\algblockdefx[class]{Class}{EndClass}[1]{\textbf{object} \textsc{#1}:}{\textbf{end object}}
\makeatletter  
\ifthenelse{\equal{\ALG@noend}{t}}%
  {\algtext*{EndClass}} 
  {}%
\makeatother

\usepackage{amsthm}
\usepackage{mathtools}
\usepackage{amsmath}
\usepackage{bbm}
\usepackage{amsfonts}
\usepackage{amssymb}

\usepackage{xpatch}
\usepackage{pifont}
\usepackage{array}
\usepackage{booktabs}
\usepackage{blindtext}
\usepackage{caption}

\usepackage[nameinlink,capitalize]{cleveref} 

\usepackage{natbib}
\usepackage{amsfonts}

\usepackage{amsfonts,amssymb,mathrsfs}
\bibpunct{(}{)}{;}{a}{,}{,}
\usepackage{etoolbox}

\usepackage{microtype}      %
\usepackage{times}
\usepackage{booktabs}
\usepackage{comment} 
\usepackage{enumitem}

\title{The Space Complexity of Learning-Unlearning Algorithms} 
\date{}

\usepackage[suppress]{color-edits}    
\addauthor{as}{purple}    
\addauthor{nv}{red} 
\addauthor{sg}{purple}
\addauthor{yc}{purple} 
\addauthor{at}{blue} 

\usepackage{times}

\usepackage{mathtools}
\usepackage{amsmath}
\usepackage{amsxtra}
\usepackage{enumitem}      
\usepackage{cleveref}
\usepackage{mathtools}
\usepackage{bbm}
\usepackage{amsfonts}

\usepackage{MnSymbol} %
\usepackage{amssymb}
\usepackage{textcomp}
\usepackage{nicefrac}  

\usepackage{xpatch}

\newtheorem{theorem}{Theorem}[section]  
\newtheorem{definition}[theorem]{Definition}
\newtheorem{lemma}[theorem]{Lemma} 
\newtheorem{claim}[theorem]{Claim}
\newtheorem{informaltheorem}{Informal Theorem} 
\newtheorem{proposition}[theorem]{Proposition}

\newtheorem{corollary}[theorem]{Corollary}

\newtheorem{remark}[theorem]{Remark}
\newtheorem{opproblem}{Open Problem}

\newtheorem*{theorem*}{Theorem}

\usepackage{prettyref}
\newcommand{\pref}[1]{\prettyref{#1}}

\newcommand{\savehyperref}[2]{\texorpdfstring{\hyperref[#1]{#2}}{#2}}
\newrefformat{eq}{\savehyperref{#1}{\textup{(\ref*{#1})}}}
\newrefformat{eqn}{\savehyperref{#1}{Equation~\ref*{#1}}}
\newrefformat{con}{\savehyperref{#1}{Conjecture~\ref*{#1}}}
\newrefformat{lem}{\savehyperref{#1}{Lemma~\ref*{#1}}}
\newrefformat{def}{\savehyperref{#1}{Definition~\ref*{#1}}}
\newrefformat{line} {\savehyperref{#1}{line~\ref*{#1}}}
\newrefformat{thm}{\savehyperref{#1}{Theorem~\ref*{#1}}}
\newrefformat{theorem}{\savehyperref{#1}{Theorem~\ref*{#1}}}
\newrefformat{corr}{\savehyperref{#1}{Corollary~\ref*{#1}}}
\newrefformat{sec}{\savehyperref{#1}{Section~\ref*{#1}}}
\newrefformat{subsec}{\savehyperref{#1}{Section~\ref*{#1}}}
\newrefformat{app}{\savehyperref{#1}{Appendix~\ref*{#1}}}
\newrefformat{ass}{\savehyperref{#1}{Assumption~\ref*{#1}}}
\newrefformat{ex}{\savehyperref{#1}{Example~\ref*{#1}}}
\newrefformat{fig}{\savehyperref{#1}{Figure~\ref*{#1}}}
\newrefformat{alg}{\savehyperref{#1}{Algorithm~\ref*{#1}}}
\newrefformat{rem}{\savehyperref{#1}{Remark~\ref*{#1}}}
\newrefformat{conj}{\savehyperref{#1}{Conjecture~\ref*{#1}}}
\newrefformat{prop}{\savehyperref{#1}{Proposition~\ref*{#1}}}
\newrefformat{proto}{\savehyperref{#1}{Protocol~\ref*{#1}}}
\newrefformat{prob}{\savehyperref{#1}{Problem~\ref*{#1}}}
\newrefformat{claim}{\savehyperref{#1}{Claim~\ref*{#1}}}
\newrefformat{exp}{\savehyperref{#1}{Experiment~\ref*{#1}}}  
\newrefformat{op}{\savehyperref{#1}{Open Problem~\ref*{#1}}}

\allowdisplaybreaks

\DeclarePairedDelimiter{\abs}{\lvert}{\rvert} 

\DeclarePairedDelimiter{\crl}{\{}{\}}
\DeclarePairedDelimiter{\prn}{(}{)}

\DeclarePairedDelimiter{\floor}{\lfloor}{\rfloor}

\let\Pr\undefined

\DeclareMathOperator{\Pr}{Pr}

\DeclareMathOperator*{\argmin}{argmin} %

\newcommand{\eps}{\epsilon}
\newcommand{\veps}{\varepsilon}

\newcommand{\ldef}{\vcentcolon=}

\newcommand{\mc}[1]{\mathcal{#1}}
\newcommand{\wt}[1]{\widetilde{#1}}

\newcommand{\mb}[1]{\boldsymbol{#1}}

\def\ddefloop#1{\ifx\ddefloop#1\else\ddef{#1}\expandafter\ddefloop\fi}
\def\ddef#1{\expandafter\def\csname bb#1\endcsname{\ensuremath{\mathbb{#1}}}}
\ddefloop ABCDEFGHIJKLMNOPQRSTUVWXYZ\ddefloop
\def\ddefloop#1{\ifx\ddefloop#1\else\ddef{#1}\expandafter\ddefloop\fi}
\def\ddef#1{\expandafter\def\csname b#1\endcsname{\ensuremath{\mathbf{#1}}}}
\ddefloop ABCDEFGHIJKLMNOPQRSTUVWXYZ\ddefloop
\def\ddef#1{\expandafter\def\csname c#1\endcsname{\ensuremath{\mathcal{#1}}}}
\ddefloop ABCDEFGHIJKLMNOPQRSTUVWXYZ\ddefloop
\def\ddef#1{\expandafter\def\csname h#1\endcsname{\ensuremath{\widehat{#1}}}}
\ddefloop ABCDEFGHIJKLMNOPQRSTUVWXYZabcdefghijklmnopqrsuvwxyz\ddefloop    %
\def\ddef#1{\expandafter\def\csname hc#1\endcsname{\ensuremath{\widehat{\mathcal{#1}}}}}
\ddefloop ABCDEFGHIJKLMNOPQRSTUVWXYZ\ddefloop
\def\ddef#1{\expandafter\def\csname t#1\endcsname{\ensuremath{\widetilde{#1}}}}
\ddefloop ABCDEFGHIJKLMNOPQRSTUVWXYZ\ddefloop
\def\ddef#1{\expandafter\def\csname tc#1\endcsname{\ensuremath{\widetilde{\mathcal{#1}}}}}
\ddefloop ABCDEFGHIJKLMNOPQRSTUVWXYZ\ddefloop

\usepackage{tikz}

\newcommand{\poly}{{\normalfont \text{poly}}}

\newcommand{\yes}{{\normalfont \textsf{Yes}}} 
\newcommand{\no}{{\normalfont \textsf{No}}} 
\newcommand{\Sc}{S_{\text{code}}}

\newcommand{\conv}{\textsf{conv}}
\newcommand{\supp}{\textsf{supp}}
\newcommand{\ip}[2]{\left\langle #1, #2 \right\rangle}
\newcommand{\lbrb}[1]{\left\{#1\right\}}
\newcommand{\lprp}[1]{\left(#1\right)}

\newcommand{\TiLU}{{\normalfont \text{TiLU}}\xspace}
\newcommand{\LU}{{\normalfont \text{LU}}\xspace}  

\newcommand{\Secref}[1]{\hyperref[#1]{Section \ref*{#1}}}
\newcommand{\Appref}[1]{\hyperref[#1]{Appendix \ref*{#1}}}

\newcommand{\starno}{\mathfrak{s}}
\newcommand{\hstarno}{\mathfrak{s}_{\circ}}
\newcommand{\eluder}{\mathfrak{e}} 

\newcommand{\aux}{{\normalfont \textsf{aux}}\xspace} 
\newcommand{\learn}{{\normalfont \textsf{Learn}}}
\newcommand{\unlearn}{{\normalfont \textsf{Unlearn}}}
\newcommand{\ticket}{{\normalfont \textsf{tx}}}
\newcommand{\ERM}{{\normalfont \text{ERM}}\xspace} 

\newcommand{\mcX}{\mathcal{X}}
\newcommand{\mcH}{\mathcal{H}}
\newcommand{\mcY}{\mathcal{Y}}
\newcommand{\mcS}{\mathcal{S}}
\newcommand{\mcG}{\mathcal{G}}
\newcommand{\mcI}{\mathcal{T}}

\newcommand{\Encode}{{\normalfont  \textsf{Encode}}} 
\newcommand{\Decode}{{\normalfont \textsf{Decode}}}
\newcommand{\Merge}{{\normalfont \textsf{Merge}}} 
\newcommand{\bit}{\crl{0, 1}} 

\newcommand{\enc}{{\normalfont \textsf{Enc}}}
\newcommand{\dec}{{\normalfont \textsf{Dec}}}

\newcommand{\VCd}{{\normalfont \textsf{VC}}}

\newcommand{\Eld}{\mathfrak{e}}
\newcommand{\Idd}{\mathfrak{m}}
\newcommand{\mis}{\Idd}
\newcommand{\Ent}{{\large \mathtt{H}}}  

\newcommand{\vZ}{\bZ}
\newcommand{\vX}{\bX}
\newcommand{\vY}{\bY}
\newcommand{\vD}{\bD}
\newcommand{\vL}{\bL}
\newcommand{\vN}{\bN}
\newcommand{\vU}{\bU}
\newcommand{\vA}{\textsf{aux}\xspace} 
\newcommand{\vT}{\textsf{tx}\xspace}
\newcommand{\Ber}{\text{Ber}}

\newcommand{\I}{{\large \mathtt{I}}}

\newcommand{\sorted}{\textsf{LexicographicallySort}}

\newcommand{\oldDec}{\normalfont \textsf{Dec}_{\texttt{vs}}}
\newcommand{\oldEnc}{\normalfont \textsf{Enc}_{\texttt{vs}}} 

\newcommand{\s}{s}

\definecolor{Gred}{RGB}{219, 50, 54}
\definecolor{Ggreen}{RGB}{60, 186, 84}
\definecolor{Gblue}{RGB}{72, 133, 237}
\definecolor{Gyellow}{RGB}{247, 178, 16}
\definecolor{ToCgreen}{RGB}{0, 128, 0}
\definecolor{myGold}{RGB}{231,141,20}
\definecolor{myBlue}{rgb}{0.19,0.41,.65}
\definecolor{myPurple}{RGB}{175,0,124}

\newcommand{\litl}{\mathfrak{l}}

\newcommand{\canonical}{{\normalfont \textsf{Canonical}}} 

\newcommand{\Poisson}{\textsf{Poisson}} 

\newcommand{\VC}{{\normalfont \text{VC}}\xspace}  

\newcommand{\Hparity}{\mcH_{{\normalfont \text{parity}}}} 

\renewcommand{\cref}{\pref}
\renewcommand{\Cref}{\pref}

\usepackage{multirow}

\usetikzlibrary{positioning, arrows.meta, fit, decorations.pathreplacing, automata, trees}
\usepackage{bbold} 

  \author{
    Yeshwanth Cherapanamjeri\thanks{Authors are listed in alphabetical order of their last names.} \\ 
    {\small\texttt{MIT}} \and
    Sumegha Garg\\
    {\small\texttt{Rutgers University}} 
    \and Nived Rajaraman \\ 
    {\small\texttt{UC Berkeley}} \and Ayush Sekhari \\   
    {\small\texttt{Boston University}} \and Abhishek Shetty \\ 
    {\small\texttt{MIT}} 
 }

\begin{document} 

\maketitle 

\begin{abstract}%

We study the memory complexity of machine unlearning algorithms that provide strong data deletion guarantees to the users. Formally, consider an algorithm for a particular learning \emph{task} that initially receives a training dataset. Then, after learning, it receives data deletion requests from a subset of users (of arbitrary size), and the goal of unlearning is to perform the  \emph{task} as if the learner never received the data of deleted users. In this paper, we ask how many bits of storage are needed to be able to delete certain training samples at a later time. We focus on the task of realizability testing, where the goal is to check whether the remaining training samples are realizable within a given hypothesis class $\mcH$. 

Toward that end, we first provide a negative result showing that the VC dimension, a well-known combinatorial property of \(\cH\) that characterizes the amount of information needed for learning and representing the ERM hypothesis in the standard PAC learning task---is not a characterization of the space complexity of unlearning. In particular, we provide a hypothesis class with constant VC dimension (and Littlestone dimension), but for which any unlearning algorithm for realizability testing needs to store \(\Omega(n)\)-bits, where \(n\) denotes the size of the initial training dataset. In fact, we provide a stronger separation by showing that for any hypothesis class \(\cH\), the amount of information that the learner needs to store, so as to perform unlearning later, is lower bounded by the \textit{eluder dimension} of \(\cH\), a combinatorial notion always larger than the VC dimension. We complement the lower bound with an upper bound in terms of the star number of the underlying hypothesis class, albeit in a stronger ticketed-memory model proposed by \cite{ghazi2023ticketed}. We show that for any class $\mcH$ with bounded star number, there exists a ticketed scheme that uses only \(\widetilde{O}(\text{StarNo}(\cH))\) many bits of storage and these many sized tickets. Since the star number for a hypothesis class is never larger than its Eluder dimension, our work highlights a fundamental separation between central and ticketed memory models for machine unlearning.

Lastly, we consider the setting where the number of deletions is \emph{bounded} and show that, in contrast to the unbounded setting, there exist unlearning schemes with sublinear 
(in $n$) storage for hypothesis classes with bounded \emph{hollow star number}, a notion of complexity that is always smaller than the star number and the eluder dimension. 
\end{abstract}

\clearpage 

\section{Introduction}    
A fundamental right that has emerged through recently proposed regulations such as European Union's \citep{GDPR16}, the California Consumer Privacy Act \citep{ccpa}, and Canada's proposed Consumer Privacy
Protection Act (CPPA),  is the ``right to be forgotten''. This principle allows individuals to demand removal of their personal data from a trained ML system. Though, this seems like a simple task at first glance, formalizing what it means to remove is challenging. In a typical large-scale ML system, the data is not only stored in its raw form in various training datasets, but can be memorized by the ML models in complicated ways, and can thus influence the behavior of the model towards other users, inadvertently leaking private information \citep{shokri2017membership,carlini2022membership,carlini2022quantifying, carlini2023extracting, ippolito2022preventing}.

Given the importance of this problem, considerable efforts have been dedicated in recent years to formalize what ``forgetting'' really means for interactions between individuals and data holders  \citep{cohen2020towards,cohen2023control,garg2020formalizing, cao2015towards}. In the context of machine learning, this led to the notion of \textit{machine unlearning}, first proposed by \citet{cao2015towards} where the goal is that,  post-deletion, the ML model behaves as if the deleted data was never included in the ML model, to begin with.
In order to circumvent the massive overhead of ``retraining from scratch'' every time a deletion is requested, there has been considerable research in more feasible strategies for machine unlearning (see \citet{nguyen} and references therein).

While the training time aspect of machine unlearning has garnered significant attention in recent literature ~\citep{ginart2019making,wu2020deltagrad,Golatkar_2020_CVPR,golatkar2020forget,BourtouleCCJTZLP21,izzo2021,NeelRS21,sekhari2021remember,jang2022knowledge,huang2023tight,wang2023kga}, both from the theoretical and practical perspective, our understanding of its memory complexity remains rudimentary.
A few recent works have considered the memory footprint of machine unlearning, and the associated tradeoffs due to memory constraints \citep{sekhari2021remember, ghazi2023ticketed}. 
 \cite{ghazi2023ticketed} introduced the notion of central and ticketed learning-unlearning schemes, and mergeable hypothesis classes, and provided space-efficient unlearning schemes for various specific hypothesis classes. 
 But there is a lack of a characterizations of memory complexity of unlearning.
Toward bridging this gap, in this paper, we ask 
\begin{quotation}
    \begin{center}
\textit{Can natural combinatorial notions of the complexity of a class $\mcH$ be used to bound the information that needs to be retained for unlearning?}
\end{center}
    \end{quotation}

This question is natural, given the rich literature on characterizations for learnability in the standard machine learning setting (i.e.~no data removal) by combinational notations e.g.~VC dimension for PAC learning \citep{Valiant84, vcdim}, Littlestone dimension for online learning \citep{littlestone1988learning}, and the star number and eluder dimension for active learning and sequential decision making \citep{hanneke2015minimax, russo2013eluder}.

    In addition to theoretical interest, practically, in many modern ML settings, the vast size of training datasets often makes storage capacity a critical bottleneck.  
In this work, we focus on the memory complexity of machine unlearning for the task of realizability testing under deletions, where the goal is to just answer whether the remaining training data is consistent with the given hypothesis class or not. This simpler task serves as a useful theoretical primitive that captures various interesting aspects of the full-fledged unlearning problem---where the learner needs to return a hypothesis within a given hypothesis class---and the tools we develop for understanding space complexity for realizability testing are useful even when considering unlearning schemes for empirical risk minimization (See \Cref{app:reduction} for formal white-box reductions).

\subsection{Our Contributions} 

We study the memory complexity of realizability testing under two models of unlearning: the central model and the ticketed model. In the central model, all the necessary information for unlearning is stored in a centralized memory (\pref{def:LU}).
On the other hand, in the ticketed model, this necessary information is spread between a central memory and users corresponding to data points in the set (\pref{def:TiLU}). 

\smallskip
\noindent We summarize our contributions in \Cref{tab:results}. 

\begin{table}[] 
    \small
\center 
 \resizebox{\textwidth}{!}{\begin{tabular}{c c c c c c}
&& \multicolumn{3}{c}{\textbf{\large Arbitrary $\#$ Deletions}}  & \textbf{\(k\)-Deletions} \\ 
&& Bounded VC Dim & Bounded Littlestone Dim & Bounded Star No/Eluder Dim.  & \\\hline 
\parbox[t]{0.3cm}{\multirow{2}{*}{\rotatebox[origin=c]{90}{\centering {\large \textbf{Central} } }}}
& \small{Upper:}
& \parbox[t]{4cm}{\centering $O(n \log(\abs{\cZ}))$ \\\small{Trivial; store the dataset}} & \parbox[t]{4cm}{\centering $O(n \log(\abs{\cZ}))$ \\\small{Trivial; store the dataset}}  
& \parbox[t]{3cm}{\centering \pref{op:op1} }   
& \parbox[t]{3cm}{\centering $O\prn*{\hstarno(\cH)^k k \log(\abs*{\cZ})}$\\\small{(\pref{thm:bdd_deletions_star_no})}} 
\rule[-20pt]{0pt}{37pt}\\ ~\\
& \small{Lower:}
& \parbox[t]{4cm}{\centering \textcolor{black}{$\Omega\prn*{n}$ for 1D thresholds} \\\small{\citep{ghazi2023ticketed}}}  
& \parbox[t]{4cm}{\centering $\Omega\prn*{n}$ for some \(\cH\) \\\small{(\pref{theorem:VClb})}}
& \parbox[t]{3cm}{\centering $\Omega\prn*{\eluder(\cH)}$\\\small{(\pref{thm:lbeluder})}} 
& \parbox[t]{4cm}{\centering $\Omega(d^k)$ \\ for \(d\)-dim linear seps. \\\small{( \pref{sec:halfspaces}, \pref{thm:linear_central_lb})}} %
\rule[-24pt]{0pt}{28pt}\\ \hline
\parbox[t]{0.3cm}{\multirow{2}{*}{\rotatebox[origin=c]{90}{\parbox[t]{2.345cm}{\centering {\large \textbf{Ticketed}} }}}} 
& \small{Upper:}
& \parbox[t]{4cm}{\centering $O(n \log(\abs{\cZ}))$ \\\small{Trivial; store the dataset}} 
& \parbox[t]{4cm}{\centering $O(n \log(\abs{\cZ}))$ \\\small{Trivial; store the dataset}} 
& \parbox[t]{4.5cm}{\centering $O\prn*{\starno(\cH) \log(\abs*{\cZ}) \log(n) }$\\\small{(\pref{thm:uneluder})} \\ \vspace{1mm} 
\scriptsize{{ \citep[\(O(d\log(n))\) for \(d \)-bit mergeable $\mathcal{H}$]{ghazi2023ticketed}}}
} 
& \parbox[t]{4cm}{\centering \textcolor{black}{Same as central\\ }} 
\rule[-24pt]{0pt}{41pt}\\~\\
& \small{Lower:}
& \parbox[t]{4cm}{\centering $\Omega\prn*{\epsilon n^{1 - \epsilon}}$ for some \(\cH\) with VC-dim $O(1/\epsilon)$ \\\small{(\pref{theorem:VClb})}} 
& \parbox[t]{4cm}{\centering $\Omega\prn*{\epsilon n^{1 - \epsilon}}$ for some \(\cH\) with LS-dim $O(1/\epsilon)$ \\\small{(\pref{theorem:VClb})}}  
& \parbox[t]{3cm}{not possible to get \\ \(\Omega(\text{poly}(\eluder(\cH)))\)  \centering \\\small{(\pref{thm:tiluUB})}}  
& \parbox[t]{4cm}{ \centering $\Omega(d^k)$ \\ for \(d\)-dim linear seps.\\\small{(\pref{sec:halfspaces}, \pref{thm:linear_central_lb})}} 
\rule[-24pt]{0pt}{26pt}\\\hline
\vspace{2mm}
\end{tabular}} 
\caption{Summary of our main results for realizability testing, ignoring the dependence on \(\veps\) and \(\delta\). Here, $\hstarno$ refers to the hollow star number (\pref{def:hollow_star}), $\starno$ refers to the  star number (\pref{def:star_no}), \(\eluder\) refers to the Eluder dimension of \(\cH\) (\pref{def:eluder}), and $\cZ$ to the domain of the hypothesis class $\mathcal{H}$.  Note that for \(\cH\) denoting the set of all linear separators in \(d\)-dimensions, \(\hstarno(\cH) \leq d + 2\), but \(\starno(\cH) = \infty\) and \(\eluder(\cH) = \infty\).}     
\label{tab:results}
\end{table}

\paragraph{Sample Compression Schemes Do Not Suffice.} 
Sample compression schemes \citep{littlestone1986relating,floyd1995sample,moran2016sample} are the standard notion of compression studied in learning theory. Here, one asks for a small subset of the dataset from which one can infer the labels of the entire dataset. This is intimately related to combinatorial dimensions such as the VC dimension and the Littlestone dimension. We show, in \Cref{subsec:vcnotenough}, that even the  Littlestone dimension is not sufficient for small memory learning-unlearning schemes for realizability testing. Since the Littlestone dimension is always larger than the VC dimension and both complexities imply sample compression schemes,  this shows that sample compression schemes are not sufficient for non-trivial learning-unlearning schemes for realizability testing.

\begin{informaltheorem}[Corresponds to \Cref{theorem:VClb}]
    There is a hypothesis class with bounded Littlestone dimension (and thus a bounded sample compression scheme) such that any learning-unlearning scheme for a data set of size $n$, in the central as well as the ticketed model, requires nearly linear in \(n\) memory.  
\end{informaltheorem}

\paragraph{Version Space Compression.} 
In \cref{sec:compression}, we consider a notion of compression known as \emph{version space compression}, which corresponds to an encoding from which one can recover the entire set of hypotheses consistent with a given data set. This notion was previously considered in \cite{hanneke2009theoretical} motivated by the fact that keeping track of realizable hypotheses suffices (even in the central model) when there are only insertions to the data set. We extend this for deletions and show that version space compression schemes suffice for ticketed unlearning schemes (\cref{lem:vstomergeable} and \cref{thm:mergeablealg}).  Furthermore, surprisingly, version space compression schemes are not sufficient for the central model (\cref{thm:lbeluder} and \cref{sec:Eluder}). This formalizes the intuition that for realizability testing, handling deletions from the data set is harder than handling additions to that dataset. In particular, we show any class with a $C$-bit version space compression has a ${O}( C \log n )$-bit ticketed LU scheme on data sets of size $n$. Further, for the class of linear thresholds in one dimension, there is a version space compression storing $2$ data points, while any central LU schemes needs linear (in $n$) memory.   
        
\begin{informaltheorem}[Corresponds to \cref{thm:lbeluder}, \cref{thm:uneluder} and \cref{lem:vstomergeable}]
Any class with a $C$-bit version space compression has a ${O}( C \log n )$-bit ticketed LU scheme on data sets of size $n$.
For the class of linear thresholds in one dimension, there is a version space compression storing $2$ data points, while any central learning unlearning schemes needs linear (in $n$) memory.
\end{informaltheorem}

\paragraph{Star Number, Eluder Dimension, and Version Space Compression.} 
We then explore version space compressions further by connecting them to the \emph{star number}, a notion of complexity arising from the study of active learning and sequential decision making. It is well know that classes with bounded star number have version space compression schemes of linear size in the star number (\cref{thm:eludercompression}; \cite{hanneke2015minimax}). That is, any class with star number $\starno$ has a version space compression that takes $ \tilde{O}( \starno)$ bits (c.f. \cref{thm:eludercompression}). Along with the previously mentioned connection between version space compression and ticketed learning unlearning schemes, this implies that:

\begin{informaltheorem}[Corresponds to \cref{thm:uneluder}]
Any class with star number $\starno$ has a $\widetilde{O}(\starno \log(n))$-bit ticketed LU scheme on data sets of size n. 
\end{informaltheorem}
        
 Furthermore, we complement this upper bound by providing a lower bound in the central model to any hypothesis class in terms of its eluder dimension. \emph{Eluder dimension}, is another notion of complexity arising in various sequential decision making problems \cite{russo2013eluder}. In particular, we show that: %

\begin{informaltheorem}[Corresponds to \cref{thm:lbeluder}]
Any central memory learning unlearning scheme for any class with eluder dimension $\eluder$ requires $ \Omega(  \min\{ n , \eluder \} ) $ bits of memory. 
\end{informaltheorem}

Since, for any hypothesis class, its eluder dimension is always larger than its star number (\pref{lem:relate_dimension}).  Taken together, the above two results show a novel separation between the central and ticketed memory models for machine unlearning.

\paragraph{Bounded Deletions.}
Motivated by circumventing the memory lower bounds in the central model (that considered arbitrary number of deletions), we study the setting where the number of deletions is bounded by a prespecified number  $k$. Here, we provide a central memory learning-unlearning scheme with space complexity that grows with the \emph{hollow star number} of $\mcH$ (\cref{thm:bdd_deletions_star_no}). 

That is, for a class with hollow star number $\hstarno$, there is a central memory LU scheme that handles $k$ deletions  and requires $ \tilde{O}( \hstarno^k)$ bits of memory. The hollow star number is always smaller than the eluder dimension and typically smaller than the Littlestone dimension. A class that witnesses this gap is the class of halfspaces in $d$ dimension, which has hollow star number $O(d)$ but infinite eluder and Littlestone dimensions. In fact, for this natural class, we show that this theorem is essentially tight (\cref{sec:halfspaces}). For the class of halfspaces in $d$ dimensions, any LU scheme (central or ticketed) that handles $k$ deletions requires $ \Omega(d^{k-1})$ bits of memory. 

\begin{informaltheorem}[Corresponds to \cref{thm:bdd_deletions_star_no}, \cref{thm:linear_central_lb} and \cref{thm:linear_tilu_lb}]
For a class with hollow star number $\hstarno$, there is a central memory LU scheme that handles $k$ deletions  and requires $ \tilde{O}( \hstarno^k)  $ bits of memory. For the class of halfspaces in $d$ dimensions, any LU scheme (central or ticketed) that handles $k$ deletions requires $ \Omega(d^{k-1})$ bits of memory.
\end{informaltheorem}

\paragraph{Connections to Empirical Risk Minimization (ERM).} 
We further show that the tools we develop for designing space complexity lower/ upper bounds for \LU schemes (and later, \TiLU schemes) for realizability testing are useful even when considering unlearning schemes for empirical risk minimization. \Cref{app:reduction} is dedicated to understanding these connections in more detail and presents new upper and lower bounds in the context of unlearning schemes for \ERM.

\subsection{Technical Overview}

\paragraph{Memory Complexity Lower Bounds.}
The space lower bounds for learning-unlearning schemes in the central-memory as well as the ticketed model follow a common recipe: construct a 1) collection of datasets indexed by a large secret/ string, and 2) set of unlearning requests such that outputs to these requests reveal the secret. We then argue that the auxiliary information stored for facilitating these unlearning requests can be used as a compression scheme for the secret; thus, the size of the secret becomes a lower bound for the memory needed for learning-unlearning schemes. The lower bounds for approximate-unlearning (see \Cref{def:LU}) needs a bit more work as the secret is not revealed completely with the unlearning queries.

A key challenge arises while showing memory lower bounds in the ticketed model. Since we are aiming to bound the size of each ticket, an unlearning request with a large number of deletions inherently allows the algorithm more power (i.e. large effective memory). To show a lower bound through compression argument, we additionally need to ensure that all the deletion requests come from a small set of users/datapoints.
In \Cref{theorem:VClb}, we construct a hypotheses class with constant Littlestone dimension, such that any learning-unlearning scheme, even in the ticketed model, requires essentially the memory required to store the entire data set.

\paragraph{Version-Space Compression.}

One of the key contributions of this work is to connect the notion of version space compression schemes for a hypothesis class $\mcH$ to machine unlearning. Given a dataset, a version space represents the set of all hypotheses from the $\mcH$ that are consistent with the dataset. A version space compression scheme is an algorithm that takes as input a data set $S$ and outputs an encoding from which one can recover the version space corresponding to $S$. We show that version space compression schemes are essentially equivalent to a notion of compression known as mergeable compression \citep{ghazi2023ticketed}, which corresponds to encoding of data sets from which one can infer whether the data set is realizable or not, with an additional property that encoding of the union of two data sets can be computed from the encoding of the two data sets. 

Using this equivalence, we show that there is a ticketed unlearning scheme for any class with a version space compression scheme. We further use this perspective to expand upon the connection to star number. It is well know that classes with bounded star number have bounded version space compression schemes, where the compression of dataset $S$ stores a star set of $S$. 

To complement this upper bound through version-space compression schemes, we also show that for classes that satisfy an additional structural assumption of having small Minimum Identification Set (MIS)---a small set such that knowing the labels of the points in the set uniquely identifies the hypothesis, one can convert a low memory exact-unlearning scheme in the central model to a version space compression scheme (\pref{thm:vscentral}). This result can be used to show that for the class of parities over $\{0,1\}^d$ any unlearning scheme in the central model must require $\Omega(d^2)$ bits of memory even for data sets of size $O(d)$.

\paragraph{Bounded Deletions.}   
All the lower bounds discussed above requires unlearning requests where the number of deletions grow polynomially with the size of the (original) training dataset. An interesting question remain as to whether the memory complexity of unlearning can be reduced when the number of deletions is bounded.
    
We show that this is indeed the case; the memory complexity in the case of constant deletions is upper bounded by the hollow star number of the hypothesis class, even in the central memory model. The hollow star number is a combinatorial parameter that captures size of the largest unrealizable data set such that removing any one point from the data set makes it realizable. Towards building an intuition for the results, we first note that for any unrealizable data set, there is a subset of size at most the hollow star number such that it is also unrealizable. This can be seen by greedily removing points from the data set until the removal of any point makes the data set realizable and the size of the set thus obtained is at most the hollow star number. Given this observation, note that the number of points whose removal from a data set makes it realizable is at most the hollow star number since the point must lie in the unrealizable set of size at most the hollow star number constructed above. This gives a simple central memory scheme that handles one deletion: store all the points whose removal makes the data set realizable. 

This idea can be applied recursively to show that the number of subsets of size $k$ whose removal from a data set makes it realizable (we call such subset $k$-critical) is at most the hollow star number to the power of $k$. We complement this upper bound by showing that for the class of linear separators in $d$ dimension (whose hollow star number is $O(d)$), the memory complexity of unlearning for $k$ deletions is lower bounded by $\Omega(d^k)$. In fact, we even show this lower bound in the ticketed model. The idea again is to construct a large family of data sets, such that the index/identity of the true underlying dataset can be inferred via making deletion queries of small support.

\section{Formalization of Learning-Unlearning}

\textbf{Basic Notation.} We will  use $[m]$ to denote the set $\{ 1,2,\cdots,m \}$ for any $m \in \mathbb{N}$. 
For any set $S$ and $p \le |S| \in \mathbb{N}$,~$\binom{S}{p} = \{ T \subseteq S : |T| = p \}$ denotes the $p$-sized subsets of $S$. 
In this paper, we will focus on the setting of binary classification.
Let $\mathcal{X}$ be a set of feature vectors. Unless stated otherwise, we study the case where $\cX$ is finite.
A hypothesis $h$ over the domain $\mathcal{X}$ is a function $\mathcal{X} \to \{ 0,1 \}$. 
A collection of hypotheses is known as a hypothesis class, which we will denote by $\mathcal{H}$.
A dataset $D\in (\cX\times \{0,1\})^{\star}$ is a sequence of ordered pairs $\{ (x_i , y_i )\} $, where $x_i \in \mathcal{X}$ and $y_i \in \{0,1 \}$. 
We will call $y_i$ the label of $x_i$. For ease of notation, we will use $\cZ$ to denote the set $(\cX\times \{0,1\})$. Let \(\cW\) be an arbitrary set. For a distribution $\bP:\cW\rightarrow [0,1]$, we  use $\bP(\cW')$ to denote the probability mass over $\cW'\subseteq\cW$, that is, $\bP(\cW')=\sum_{w\in\cW'}\bP(w)$. We formalize the notion of a learning task and unlearning below.

\begin{definition}[Learning task] 
\label{def:learning_task}
Let \(\cW\) be a set of possible outcomes. 
A learning task with respect to a hypothesis class $\mcH$ is denoted by function $f : \cZ^\star \to 2^\cW$ that takes in a dataset as input,  and outputs a solution in range $\cW$.
\end{definition} 

We will focus on two learning tasks in this paper: realizability testing and Empirical Risk Minimization (ERM). For the realizability testing problem, the goal is to determine whether there exists a hypothesis that labels all the points in the dataset correctly, as formalized next.

\begin{definition}[Realizability Testing]
Let $\mathcal{H}$ be a hypothesis class.
A dataset $D = \{ (x_i , y_i ) \}_{i=1}^n$ is said to be $\mathcal{H}$-realizable if there exists a hypothesis $h \in \mathcal{H}$ such that $y_i = h(x_i)$ for all $i \in [n]$. 

\smallskip
\noindent Realizability testing corresponds to the learning task $f$ of deciding whether an input dataset $D = \{ (x_i,y_i) \}_{i=1}^n$ is $\mathcal{H}$-realizable or not. In particular, $\cW = \{ \yes, \no \}$ and $f(D) = \{\yes\}$ if $D$ is $\mathcal{H}$-realizable and $f(D) = \{\no\}$ otherwise. 
\end{definition}

\noindent 
We defer the definition of the learning task of ERM to \Cref{app:reduction} since the majority of the main body of the paper focuses on realizability testing (however, many results trivially extend to the ERM setting; cf.~\Cref{app:reduction} for details). Note that for the task of computing ERMs, the set \(\cW\) referred to in \pref{def:learning_task} denotes the set of Empirical Risk Minimizers within the class \(\cH\) for a given dataset \(S\)\asedit{, w.r.t. \(0\)-\(1\)-loss.}

\begin{definition}[Central-memory $(\veps,\delta)$-Learning-Unlearning ($\LU$) scheme for a learning task] \label{def:LU}
Let $\veps, \delta\in (0, 1)$. For a learning task $f : \cZ^\star \to 2^{\cW}$ with respect to a hypothesis class $\mathcal{H}$, an $(\veps,\delta)$-\LU scheme corresponds to a pair of randomized algorithms $(\learn,\unlearn)$. For any dataset $D = \{ (x_i,y_i) \}_{i=1}^n$ and index set $I \subseteq [n]$, the scheme guarantees that: 
\begin{itemize}
    \item $\learn(D)$ outputs a solution in $f(D)\subseteq \cW$, and some auxiliary information of the dataset $\aux(D) \in \{0, 1 \}^\star$. Let $\bP_D$ denote the distribution over solutions outputted by $\learn(D)$. 
    \item For any unlearning query denoted by a subset of dataset,
    $D_I = \{ (x_i,y_i) : i \in I \}$, $\unlearn (D_I, \aux)$ outputs a solution in $\cW$ such that, for all subsets $\cW'\subseteq\cW$,
    \[\Pr[\unlearn (D_I, \aux(D))\in \cW']\le e^{\veps}\bP_{D\setminus D_I}(\cW')+\delta.\]
\end{itemize}
Here, the randomness is over the auxiliary information outputted by the $\learn$ algorithm and random bits used in the $\unlearn$ algorithm.

The space complexity of an $(\veps,\delta)$-\LU scheme is said to be $\s (n)$ if for all $n$, on any input dataset $D$ of size at most $n$, the learning-unlearning scheme satisfies $|\aux| \le \s (n)$, that is, the number of bits of auxiliary information stored is at most $\s (n)$. 
\end{definition}

Throughout the paper, when $\veps$, $\delta$ are not important or are clear from the context, we will use the phrase $\LU$ scheme to refer to a central-memory $(\veps,\delta)$-\LU scheme. Additionally, we will overload the notation \LU scheme to indicate \LU schemes for the learning task of realizability testing, which is the main focus of this paper.
We call an $\LU$ scheme \emph{non-trivial} if it uses space that scales sublinearly in the size of the input dataset.\footnote{It is natural for $\s (n)$ to have at least some mild dependence on the size of the dataset $n$ since just keeping track of the number of points in the dataset requires $\log(n)$ memory $n$.} Unless stated otherwise, we assume that $n<< \min(|\cX|,|\cH|)$.  When the space complexity is allowed to scale linearly with any of these parameters, a learner can use the following trivial strategies to implement a central-memory $(0,0)$-\LU scheme.
\begin{itemize} 
\item The learner stores the entire training dataset in $\aux$. For any unlearning query, the unlearner recomputes the solution to the learning task on the remainder of the dataset. This requires $O(n\log |\cX|)$ bits of storage in the central memory.
\item For every $x\in \cX$, the learner stores how many times $(x,0)$ and $(x,1)$ appear in the dataset. For any unlearning query, the unlearner can again deduce the remaining dataset. This requires $O(|\cX|\log n)$ bits of storage.
\item For every hypothesis $h\in\mcH$, the learner stores the count of data points that disagree with $h$. For any unlearning query, the unlearner recomputes this count for every hypothesis, and checks if any count is $0$. This requires $O(|\cH|\log n)$ bits of storage.
\end{itemize}

We call a \LU scheme to be \emph{space-efficient} if its space complexity scales polynomially with $\log n$, $\log {|\mathcal{H}|}$  and $\log{\abs{\cX}})$.  In this paper, we focus on schemes using $\poly(\log n, d(\mathcal{H}), \log |\cX|)$ bits of memory, where $d(\mathcal{H})$ is a combinatorial complexity measure of the hypothesis class $\mathcal{H}$. Surprisingly, sublinear-memory \LU schemes often do not exist, even for classes with low complexity. As shown in \Cref{theorem:VClb}, there are classes with constant VC and Littlestone dimensions that do not admit sublinear-memory $(\veps,\delta)$-\LU schemes for any $\delta < 1/2$. To circumvent  these limitations, we also examine the ticketed model of unlearning from \citet{ghazi2023ticketed}, which generalizes central-memory schemes by allowing each data point (tied to a client) to store additional per-client metadata.  

\begin{definition}[Ticketed $(\veps,\delta)$-Learning-Unlearning (\TiLU) scheme] \label{def:TiLU}
For a learning task $f : \cZ^\star \to 2^{\cW}$ $w.r.t.$~a hypothesis class $\mathcal{H}$, an $(\veps,\delta)$-\TiLU scheme corresponds to algorithms $(\learn, \unlearn)$ such that for any dataset $D = \{ (x_1,y_1),\dots,(x_n,y_n)\}$ and index set $I\subseteq[n]$, 
\begin{enumerate}[label=\(\bullet\)] 
    \item $\learn (D)$ returns a solution in $f(D)\subseteq \cW$, and some auxiliary information of the dataset $(\aux, \{ \ticket_i \}_{i \in [n]} )$. Here, $\aux \in \{ 0,1 \}^{\star}$ is stored in the central memory, while the ticket $\ticket_i \in \{ 0,1 \}^\star$, associated with the sample $(x_i,y_i)$, is stored with the $i$-th user. Let $\bP_D$ denote the distribution over solutions output by $\learn(D)$. 
    \item For any unlearning query given by a subset of the dataset, $D_I = \{ (x_i,y_i) \}: i\in I\}$, using the auxiliary information in central memory and the tickets from users in $I$,
    $\unlearn$ outputs a solution in $\cW$ such that, for all $\cW'\subseteq\cW$, $$\Pr[\unlearn (D_I, \aux(D),\{ \ticket_i \}_{i \in I})\in \cW']\le e^{\veps}\bP_{D\setminus D_I}(\cW')+\delta.$$
\end{enumerate} 

Here, the randomness is over the random bits of $\learn$ and $\unlearn$ algorithms.
The space complexity of an $(\veps,\delta)$-\TiLU scheme is said to be $\s (n)$ if on any input dataset $D$ of size at most $n$, the learning-unlearning scheme satisfies $\max \{ |\aux|, \max_{i \in [n]} |\ticket_i| \} \le \s (n)$, i.e., \asedit{both} the ticket size and number of auxiliary bits at the server are at most $\s (n)$. 
\end{definition}

The tickets are additional information the learner provides to each user, in order to reduce the memory requirement at the server. In particular, to unlearn some set of points indexed by $I \subseteq [n]$, the learner can aggregate tickets from all the users in $I$ to carry out unlearning. The motivation for this model is that the learner has access to more information to unlearn larger subsets of the dataset. 
 We will see both general upper bounds and strong lower bounds on the memory complexity of a \TiLU scheme.

\paragraph{Further Notation.} We will use information theoretic quantities to prove the space complexity lower bounds. We use capital bold letters such $\vX, \vY,\vZ$, etc., to denote random variables and $x, y, z,$, etc., to denote the values these random variables take. Given a probability distribution $\bP:\mcX\rightarrow [0,1]$, we use the notation $x\sim \bP$ when value $x$ is sampled according to distribution $\bP$.   Similarly, we use the notation $z\sim\vZ$ to denote the process that $\vZ$ takes value $z$ with probability $\Pr[\vZ=z]$. We use $\Ber(q)$ to denote the Bernoulli distribution which takes value $1$ with probability $q$ and $0$ with probability $1-q$. 
We use notations $\bbE[\vZ]$  to denote the expectation of random variable $\vZ$. $\bbE[\vZ \mid{} \vY=y]$ denote the expectation of random variable $\vZ$ conditioned on the event $\vY=y$. 

\paragraph{Basics of Information Theory.} 
Given a random variable $\vZ$, $\Ent(\vZ)$ denotes the Shannon entropy of $\vZ$, that is, $\Ent(\vZ)=\sum_{z}\Pr(\vZ=z)\log_2  (1/\Pr(\vZ=z))$. We also use $\Ent(\bP)$ to denote the entropy of a probability distribution $\bP$. Overloading the notation, we will denote the entropy of $\Ber(q)$ by $\Ent(q)$. 

\smallskip
$\I(\vX;\vY\mid{}\vZ)$ represents the mutual information between $\vX$ and $\vY$ 
conditioned on the random variable $\vZ$. $\I(\vX;\vY\mid{}\vZ)=\Ent(\vX\mid{}\vZ)-\Ent(\vX\mid{}\vY,\vZ)$, where $\Ent(\vX\mid{}\vY)=\bbE_{y\sim \vY}\Ent(\vX\mid{}\vY=y)\le \Ent(\vX)$. 
Next, we describe some of the properties of mutual information used in the paper. 
\begin{enumerate}
\item \label{itemmi1} (Chain Rule) $\I(\vX;\vY,\vZ)=\I(\vX;\vY)+\I(\vX;\vZ\mid{}\vY)$. 
\item \label{itemmi3} If $\I(\vX;\vY) = 0$, then $\I(\vX;\vZ\mid{}\vY)\ge \I(\vX;\vZ)$.
\item \label{itemmi2} If $\I(\vX;\vY\mid{}\vZ) = 0$, then $\I(\vX;\vZ) \ge \I(\vX;\vZ\mid{}\vY)$.

\end{enumerate}
Property \ref{itemmi1} follows from the chain rule for Shannon entropy ($\Ent$).  Properties \ref{itemmi3} and \ref{itemmi2} follow from the observation that 
\[\I(\vX;\vY\mid{}\vZ)+\I(\vX;\vZ)=\I(\vX;\vY,\vZ)=\I(\vX;\vY)+\I(\vX;\vZ\mid{}\vY).\]
As mutual information is non-negative, if $\I(\vX;\vY\mid{}\vZ) = 0$, then $\I(\vX;\vZ)\ge \I(\vX;\vZ\mid{}\vY)$  and if $\I(\vX;\vY)=0$, then $\I(\vX;\vZ\mid{}\vY)\ge \I(\vX;\vZ)$.

\section{Does Bounded VC Dimension Imply Unlearning?} \label{sec:VCnotenough}  

It is natural to ask whether classes $\mathcal{H}$ with bounded VC dimension (or Littlestone dimension) support sublinear-memory unlearning schemes for $\mathcal{H}$-realizability testing. 
These complexity measures are known to be related to sample compression schemes for learning \citep{floyd1995sample, moran2016sample}.  We start with a formal definition of the VC dimension.

\begin{definition}[\VC dimension; \cite{vcdim}] 
A set of points $\wt \cX \subseteq \mathcal{X}$ is said to be shattered if every possible signing on the points can be realized across hypotheses in $\mathcal{H}$ i.e. for all functions $ t : \wt \cX \to \{0,1\}$, there exists a function $h \in \mathcal{H}$ such that for all $x \in \wt \cX$, $t(x) = h(x)$. 
The \VC dimension of a hypothesis class $\mathcal{H}$, denoted by $\VCd(\mcH)$, is the cardinality of the largest subset of its domain which is shattered by the hypothesis class. 
\end{definition}

We next define the Littlestone dimension of a hypothesis class. 

\begin{definition}[Littlestone dimension; \cite{littlestone1988learning}] 
    \label{def:littlestone} 
Given a hypothesis class \(\cH\) that maps elements from a domain \(\cX\) to \(\crl{0, 1}\), the  littlestone tree is a rooted binary tree, where each node is labeled by a point \(x \in \cX\),  the left edge is labeled by \(y = 0\), and the right edge is labeled by \(y = 1\). The tree is said to be shattered by \(\cH\) if, for every branch in the tree there exists a concept \(h \in \cH\) which is consistent with the edges of the branch (i.e.~for every \(x\) and the decision \(y\) on this branch, we have \(h(x) = y\)). 

The Littlestone dimension of the hypothesis class \(\cH\)  is the largest \(\ell \in \bbN \cap \crl{0}\), for which there exists a complete Littlestone tree of depth \(\ell\) all of whose branches are shattered by \(\cH\). 
\end{definition} 

\subsection{Unlearning with Sublinear Memory Implies Bounded VC Dimension} \label{subsec:vclb} 

We first show that any class, for which there exists a central-model learning-unlearning algorithm with non-trivial space complexity, must have bounded VC dimension, that is, it must also be learnable under the PAC learning model \citep{vcdim}. This result follows as a corollary of the following theorem, which shows that any bounded VC class needs a certain number of bits in the central memory for any \LU scheme.

\begin{theorem} \label{theorem:LU=>VC} Let $\veps\in[0,1]$ and $\delta\in[0,1/2)$ be constants. For all hypothesis classes $\mcH$, any $(\veps,\delta)$-\LU scheme for $\mcH$-realizability testing must use $\VCd(\mcH)\cdot (1-\Ent(\delta))$ bits of storage, even for datasets of size $n=O(\VCd(\mcH))$.
\end{theorem} 

\begin{corollary}[\LU scheme $\implies$ Bounded VC dimension] \label{corr:1}
 Let $\veps\in[0,1]$ and $\delta\in[0,1/2)$ be constants. Consider any hypothesis class $\mathcal{H}$ for which there exists a $C \in \mathbb{N}$ and an $(\veps,\delta)$-\LU scheme with space complexity of at most $\s(n) = C \log(n)$ bits on datasets of size at most $n$. Then, the {\normalfont VC} dimension of $\mathcal{H}$ is at most $O_\delta(C \log(C))$. \end{corollary} 
\noindent 

The above corollary is the contrapositive of \Cref{theorem:LU=>VC}; indeed, if the VC dimension of $\mathcal{H}$ were some $d \gg C$, then by \Cref{theorem:LU=>VC}, there must exist a dataset of size $2d$ such that any \LU scheme requires a space complexity of $\Omega(d)$ bits. On this dataset, the \LU scheme in \Cref{corr:1} has a space complexity upper bounded by $m = C \log (2d)$.  
We get a contradiction for $d=O(C\log(C))$. This result can be extended to the case where the space complexity of the schemes scales as sublinearly with $n$; for a hypothesis class $\mcH$, if there exists a \LU scheme with space complexity at most $C f(n)$, where $f = (\cdot)^\alpha$ for a constant $\alpha \in (0,1)$, then $\mcH$ has VC dimension at most $O(C^{1/(1-\alpha)})$. 

\subsection{Finite VC Dimension Does Not Imply Unlearning with Sublinear Memory} 
\label{subsec:vcnotenough}

\Cref{corr:1} shows that the existence of a sublinear \LU scheme implies that the hypothesis class must have finite VC dimension. In this section, we show that the converse is not true.
The main result of this section is the following surprising lower bound: there exist hypothesis classes that have constant Littlestone dimension (and thereby constant VC dimension), but any \LU scheme that supports unlearning a constant number of  points requires a space complexity of $\Omega (n)$ bits. Moreover, these results extends to a near-linear space complexity lower bound for \TiLU schemes.

\begin{theorem} \label{theorem:VClb}
For any $\beta \in (0,1)$, there exists a hypothesis class \(\cH\) with Littlestone dimension (and thereby, VC dimension) at most $1/\beta+1$, such that 
\begin{itemize}
  \item For any $(\epsilon, \delta )$-\LU scheme for $\cH$-realizability testing, the space complexity is at least \\ $\Omega ( (1- \Ent(\delta)) \cdot  n)$ bits, 
  \item For any $(\epsilon, \delta )$-\TiLU scheme for $\cH$-realizability testing, the space complexity is at least \\ $\Omega ( \beta (1- \Ent(\delta)) \cdot  n^{1-\beta})$ bits.
\end{itemize}
\end{theorem}

Thus, the VC dimension (or Littlestone dimension) does not characterize whether there exists a \LU or \TiLU scheme with space complexity scaling as $\poly(\log(n), \log(|\cX|))$. In the next section, we tune into a property of every hypothesis class that generalizes this lower bound. %

\section{Separation Between Central and Ticketed Memory Models via Combinatorial Dimensions}    
\label{sec:Eluder}

\noindent 
The key idea behind the lower bound in \Cref{theorem:VClb} is to show that it is possible to find a sequence of points such that under an appropriate unlearning query, the presence or absence of the $i^{\text{th}}$ point in the sequence decides whether the dataset is separable or not. In particular, consider the sequence of points $\{ (x_i,0) : i \in [m] \}$ in the dataset. For any $i \in [m]$, consider $S = \{ (x_j,0): j \le i \}$; there exist hypotheses $h,h' \in \mathcal{H}$ which agree with all the labels in $S$, but don't agree on their labels of the next point, i.e., $0 = h(x_{i+1}) \ne h' (x_{i+1}) = 1$. By choosing the unlearning query carefully, we force $h'$ to be the only function that potentially agrees with the remaining data, i.e. $\{ (x_i,0) : i \in [m] \}$ and thereby, realizability testing can be used to assess whether $(x_{i+1},0)$ exists in the dataset or not. 

Such sequences have been considered in learning theory under the name of ``eluder sequences'' \citep{russo2013eluder, hanneke2024eluder}.   
The presence of such a long sequences is a problem for an unlearning scheme - knowing which of $(x_1,0),\cdots,(x_i,0)$ exist in the dataset provides no indication as to whether $(x_{i+1},0)$ exists in the dataset or not, and the unlearning scheme ends up having to store information about all such points to be able to answer the unlearning queries.

We show that the existence of a long eluder sequence for a hypothesis class implies that any \LU scheme for the class has high space complexity on worst-case datasets. To this end, we define the eluder dimension of a class \citep{russo2013eluder, hanneke2024eluder}.
First, we recall the definition of the version space.

\begin{definition}[Version Space]
    Let $\mathcal{H}$ be a hypothesis class over a domain $\mathcal{X}$.
    Let $D$ be a data set of labeled points.
    The version space of $D$ with respect $\mathcal{H}$, denoted by $\mathcal{H} (D)  $, is defined as the set of hypotheses consistent with these datapoints i.e. $\mathcal{H}(D) = \left\{ h \in \mathcal{H}\mid h(x) = y, ~ \forall (x,y) \in D \right\}. $

    \end{definition}

\begin{definition}[Eluder Dimension; \cite{russo2013eluder}]  
\label{def:eluder} 
Let $\mathcal{H}$ be a hypothesis class over a domain $\mathcal{X}$.
A sequence of labeled points $(x_1,y_1),\dots,(x_\ell,y_\ell)$ is said to be an eluder sequence if for each $i \le \ell-1$, the version space of $\{ (x_1,y_1),\dots,(x_i,y_i) \}$ contains 
hypotheses $h$, $h'$ such that $h (x_{i+1}) \ne h' (x_{i+1})$. 
The eluder dimension of $\mathcal{H}$ is defined as the largest value of $\ell$ with an eluder sequence of length $\ell$.  
\end{definition}

Bounds on the eluder dimension for various function classes are well known, e.g. when \(\cF\) is finite, \(\Eld(\cH) \leq \abs{\cH} - 1\), and when \(\cH\) is the set of \(d\)-dimensional function with bounded norm, then \(\Eld(\cH) = O(d)\). We refer the reader to \cite{russo2013eluder, mou2020sample, li2022understanding} for more examples. Our main lower bound in this section is as follows: 
\begin{theorem} \label{thm:lbeluder}
Let $\veps\in[0,1]$ and $\delta\in[0,1/2)$ be constants. 
For any hypothesis class $\cH$ with eluder dimension $\eluder(\cH)$, the space complexity of any $(\veps,\delta)$-\LU scheme on datasets of size $n$, is at least $\Omega(\min(n,\eluder(\cH) ))$ (where $\Omega$ hides a constant depending on $\delta$).
\end{theorem}

\subsection{Ticketed Learning-Unlearning Schemes Under Bounded Star Number} 

We next proceed to an upper bound in the ticketed model. In particular, we show that for classes with bounded star number, there is a space-efficient \TiLU scheme. We first define the star number: 

\begin{definition}[Star number; \cite{hanneke2015minimax}] 
\label{def:star_no} 
The star number of a class \(\cH\) over a domain \(\cX\), denoted by \(\starno(\cH)\), is defined as the largest \(\ell \in \bbN\) for which there exists a set of points \(\cS = \crl{(x_1, y_1), \dots, (x_{\ell}, y_{\ell})}\) such that \(\cS\) is  realizable via \(\cH\), and for every \(i \in \ell\), the one step neighbor of  \(\cS\) is realizable via \(\cH\), or equivalently, there exists a hypothesis \(h_0 \in \cH\) such that \(h_0(x_i) = y_i\) for \(i \in [\ell]\), and that for all \(i \in [\ell]\) there exists \(h_i\) such that \(h_i(x_i) = \bar y_i\) and \(h_i(x_j) = y_j\) for all \(j \neq i\), where \(\bar y_i\) denotes the complement of the  label \(y_i\). 
\end{definition}


Our key result of this section is the following upper bound.\footnote{Note that we provide deterministic learning-unlearning schemes in \pref{thm:uneluder}, and thus \(\eps = 0\)  and \(\delta = 0\), following \pref{def:TiLU}.} 

\begin{theorem}[TiLU Scheme under bounded eluder dimension] \label{thm:uneluder} For any hypothesis class \(\cH\) with star number \(\starno(\cH)\), there exists a \((0, 0)\)-\TiLU scheme with ticket size $O( \starno(\cH) \log (\abs{\cZ}) \log(n))$ and central memory size of \(1\) bit. %
\end{theorem}

Recall that for any hypothesis class $\cH$, the star number is always less than or equal to the eluder dimension; that is, $\starno (\cH) \leq \eluder(\cH)$ (see \pref{lem:relate_dimension}). Consequently, \pref{thm:uneluder}, together with \pref{thm:lbeluder}, demonstrates a separation between the central and ticketed models of unlearning. Specifically, while the lower bound for the central memory model in \pref{thm:lbeluder} scales with the eluder dimension, the upper bound for the ticketed model in \pref{thm:uneluder} scales with the star number, which can be significantly smaller.

The proof of \pref{thm:uneluder}  goes through compression schemes, which we define over this and next subsection.
The following definition is a straightforward extension of mergeable hypothesis classes in \cite{ghazi2023ticketed}, but for realizability testing. 

\begin{definition}[Mergeable hypothesis class for realizability testing] 
\label{def:mergeable_testing} 
A Hypothesis class $\cH \subseteq \crl{\cX \mapsto \cY}$ is said to be \emph{$C$-bit mergeable} for realizability testing if there exist methods 
$\Encode, \Merge$ and $ \Decode$ such that
\begin{itemize} 
\item $\Encode: (\cX \times \cY)^* \mapsto \bit^C $ is a permutation-invariant encoding of its input into $C$ bits. 
\item $\Decode: \bit^C  \mapsto \crl{\yes, \no}$ such that $\Decode(\Encode(S)) = \yes$ if \(S\) is $\cH$-realizable, and is  \(\no\) otherwise. 
\item $\Merge: \bit^C \times \bit^C   \mapsto \bit^C$ such that for all datasets  $S_1, S_2 \in (\cX\times \cY)^{*}$, it holds that $ \Merge(\Encode(S_1), \Encode(S_2))=\Encode(S_1 \cup S_2)$.
\end{itemize} 
\end{definition}

The following theorem shows that a mergeable class leads to a ticketed LU scheme.  
The proof of \Cref{thm:mergeablealg} is given in \pref{app:mergeablealg} for completeness.

\begin{theorem} \label{thm:mergeablealg} Suppose \(\cH\) is a \(C\)-bit mergeable hypothesis class, then it admits a \((0, 0)\)-\TiLU realizability testing scheme, with both the ticket size and auxiliary bits bounded by \(O(C\log n)\). 
\end{theorem}

 A key question is when mergeable compressions exist. The next lemma addresses this in terms of the eluder dimension.  
 The proof of the above lemma is deferred to \pref{sec:compression}. It goes through version-space compression, which we argue to be an essential as well as sufficient notion of compression for exact unlearning under the central and ticketed model respectively. 
 Precisely, the above lemma follows from \pref{lem:vstomergeable} and  \pref{thm:eludercompression}.

\begin{lemma} Any class \(\cH\) with star number \(\starno(\cH)\) is \(O(\starno(\cH)\log(\abs{\cZ}))\)-bit mergeable. 
\end{lemma} 

\paragraph{{\color{black} Can we hope to get a general lower bound for ticketed model?}} \pref{thm:lbeluder} shows that any central LU algorithm needs $\Omega (\Eld(\mcH)\log n)$ memory. On the other hand, in \pref{thm:uneluder}, we showed an unlearning scheme with tickets of size $\widetilde{O} (\starno(\mcH)\log n$). A natural question is whether the star number or the eluder dimension characterizes unlearning in the central or the ticketed unlearning model. We show that such a characterization impossible for the latter, while leaving the question open for the former. 

\begin{theorem}\label{thm:tiluUB}
There exists a hypothesis class $\mcH$ with VC dimension $d$ (hence, eluder dimension and star number at least $d$), admitting a \TiLU scheme with space complexity at most $O(\max \{ \log(d), \log(n), \log (|\mathcal{X}|) \})$. 
\end{theorem}

\begin{opproblem}
\label{op:op1}
Let \(\cH\) be a hypothesis class with eluder dimension \(\Eld(\mcH)\). Under the central memory model, does there exist a learning-unlearning scheme for \(\cH\) with space complexity \(\poly\prn*{\Eld(\cH), \log(\abs{\cX}), \log(n)}\)? 
\end{opproblem}

\subsection{A Compression Perspective} \label{sec:compression}  
As discussed in \pref{sec:VCnotenough}, both VC and Littlestone dimensions aren't sufficient to justify the hardness of memory-bounded unlearning. Therefore, sample compression schemes, well studied in the literature \citep{floyd1995sample,moran2016sample}, aren't sufficient for unlearning in the central as well as ticketed model.  In this section, we introduce the concept of version-space compression \citep{el2010foundations, el2012active, wiener2015compression, hanneke2015minimax} and provide evidence for why this is the right notion of compression for the problem of exact learning-unlearning.

\begin{definition}[Version-Space compression]  
\label{def:vs_compression}
 Let $\mcH$ be a hypothesis class over domain $\mcX$ and range $\mcY$. A $C$-bit version-space compression scheme for $\mcH$, represented by a pair of functions $(\enc,\dec)$, is defined as follows: 
\begin{itemize}
\item Given a set of samples $\mcS=\{(x_1,y_1),(x_1,y_2),\ldots,(x_r,y_r)\}$ as input, the encoding function $\enc:(\mcX\times \mcY)^*\rightarrow \{0,1\}^C$ outputs a $C$-bit compression $\enc(\mcS)$.
\item  On a $C$-bit string, the decoding function $\dec:\{0,1\}^C\rightarrow 2^{\mcH}$ outputs a subset of the hypothesis $\mcH$, such that for all sample sets $\mcS$, $\dec(\enc(\mcS))=\mcH(\mcS),$
where $\mcH(\mcS)=\{h\in\mcH\mid  h(x)=y \text{~for all~} (x,y)\in \mcS \}$. 
\end{itemize}
\end{definition}

First, we show that any version-space compression scheme implies a mergeable compression scheme and vice-versa. Combined with \Cref{thm:mergeablealg}, we will show that if a hypothesis class $\mcH$ admits a $C$-bit version-space compression scheme, then there exists a TiLU algorithm with ticket size at most $O(C\log n)$.

\begin{lemma}[Version-Space Compression $\implies$ Mergeable Compression] \label{lem:vstomergeable} 
Suppose a hypothesis class \(\cH\)  has a \(C\)-bit version-space compression scheme, given by the functions \(\oldEnc\) and \(\oldDec\), then \(\cH\) is a \(C\)-bit mergeable hypothesis class for realizability testing. 
\end{lemma}

Next, we show that any hypothesis class over a finite domain $\mcX$,  having finite star number admits a finite version-space compression scheme.

\begin{theorem}[{\cite{hanneke2015minimax}}]  \label{thm:eludercompression} 
Any hypothesis class \(\cH\) has a $O(\starno(\mcH)\log(|\cZ|)$-bit version-space compression scheme, where $\starno(\cH)$ denotes the star number of \(\cH\).  
\end{theorem}

\noindent 
The result of \pref{thm:eludercompression}  can be found in Section 7.3.1 of \citet{hanneke2015minimax}. We provide a proof in \pref{app:eludercompression_proof} for completeness.

Next, we show that version-space compression is necessary for ticketed scheme upper bounds that go through \pref{thm:mergeablealg}.

\begin{lemma}[Mergeable Compression $\implies$ Version-Space Compression]
\label{lem:mergeable_compression}
If a class $\mcH$ is $C$-bit mergeable for realizability testing, then it also admits a $C$-bit version-space compression scheme.
\end{lemma}
 \pref{lem:vstomergeable} along with \pref{thm:mergeablealg} implies that a version-space compression scheme for a hypothesis class $\mcH$ can be used to construct a \TiLU algorithm for $\mcH$. While  \pref{thm:tiluUB} was used to show that star number fails to completely characterize the memory complexity of \TiLU schemes, it also shows that a version-space compression scheme is not necessary for ticketed unlearning.  On the other hand, we show in \pref{thm:vscentral} in \pref{app:version_space_necessary}   that version-space compression is indeed necessary for unlearning in the central model.

\section{Unlearning Small Deletion Sets}  
\label{sec:bdd_deletions} 

In this section, we consider the setting where only a bounded number of points are included in a deletion request, and study the space complexity of an $\LU$ scheme for realizability testing, as a function of the number of points. Formally, a request provides a list of at most $k$-points to be deleted (where \(k\) is known beforehand) and the goal of $\unlearn$ function is to determine whether the remaining dataset is realizable. 

From \Cref{theorem:VClb}, we observe that even in the case of bounded deletions, the \VC-dimension of a hypothesis class fails to characterize whether there exist $\LU$ schemes with sublinear space complexity. Here, we adopt an alternative notion of complexity called hollow star number, and show that for function classes with hollow star number $\hstarno$, there exists an unlearning scheme with space complexity $O(\hstarno^k\log (|\cZ|))$. The hollow star number is closely related to the \textit{dual Helly dimension} \citep{bousquet2020proper}, and a formal definition is provided below:  

\begin{definition}[Hollow star number; \cite{hanneke2015minimax}] 
\label{def:hollow_star} 
The hollow-star number of a class \(\cH\) over a domain \(\cX\), denoted by \(\hstarno(\cH)\), is defined as the largest \(\ell \in \bbN\) for which there exists a set of points \(\cS = \crl{(x_1, y_1), \dots, (x_{\ell}, y_{\ell})}\) such that \(\cS\) is not realizable via \(\cH\), but for every \(i \in \ell\), the one step neighbor of  \(\cS\) is realizable via \(\cH\), or equivalently, there exists a hypothesis \(h_i\) such that \(h_i(x_i) = \bar y_i\) and \(h_i(x_j) = y_j\) for all \(j \neq i\), where \(\bar y_i\) denotes the complement of the  label \(y_i\). 
\end{definition} 

When clear from the context, we use \(\hstarno\) to denote the hollow star number of the class \(\cH\). Notably, the hollow star number is always upper bounded by the star number (\Cref{lem:hstar_star}), and for hypothesis classes such as linear separators, the hollow star number is bounded by its dimension (\Cref{lem:critic_set_size_linear}), while the star number and the eluder dimension is unbounded. Next, we show that for any unrealizable set, there exists an unrealizable subset of size bounded by the hollow star number.

\begin{lemma}
    \label{lem:bdd_unrealizable_set}
    Let $S = \{(x_1, y_1), \dots, (x_l, y_l)\}$ be an unrealizable dataset. Then, there exists a subset \(S' = \{(x'_1, y'_1), \dots, (x'_m, y'_m)\} \subset S\) which is also unrealizable and furthermore, $\abs{S'} \leq \hstarno$.  
\end{lemma}

Next, we define a class of subsets of an unrealizable set which form a core part of our unlearning procedure. These will corresponds to subsets of points whose removal from the dataset allows for realizable classification of the dataset. Furthermore, they satisfy the additional desirable property that \emph{any} subset whose removal renders the dataset realizable \emph{must} contain one of these critical sets. Hence, it suffices to simply store the set of $k$-critical sets. The main contribution of this section is showing that this quantity may be bounded in terms of the hollow star number.

\begin{definition}[$k$-critical sets]
    \label{def:kcritic_set}
    Let $\cX, \cH$ denote a domain space and a hypothesis class respectively. For a set $S = \{(x_1, y_1), \dots, (x_n, y_n)\}$ that is unrealizable with respect to $\cH$, we say that a size-$k$ subset, $Q \subset S$ ($\abs{Q} = k$), is $k$-critical if $S \setminus Q$ is realizable and furthermore, for any proper subset $Q' \subset Q$ ($\abs{Q'} < \abs{Q}$), $S \setminus Q'$ is unrealizable.
\end{definition}

In the proof of the following theorem, we provide a bound on the number of $l$-critical sets in terms of the hollow star number which implies the existence of a small-sized unlearning scheme.

\begin{theorem}
    \label{thm:bdd_deletions_star_no}
    Let $\cH$ be a hypothesis class over a domain $\cX$ and $k > 0$. Then, there exists a $(0,0)$-\LU-scheme for realizability testing, when deletion queries contain at most $k$-points, with space complexity at most $O(\hstarno^k \cdot k\log  (\abs{\cZ}))$.
\end{theorem}

Note that the upper bound is nontrivial because there are classes for which the star number and eluder dimension are unbounded, but the hollow star number is bounded (e.g.~halfspaces).

\begin{opproblem} 
\label{op:op2}
Suppose \(\cH\) is a hypothesis class with hollow star number \(\hstarno\). Does there exist a \TiLU scheme for \(\cH\), for arbitrary number of deletions, where both the ticket size and auxiliary memory size is bounded by \(\poly(\hstarno^{\hstarno}, \log(\abs{\cZ}, \log(\abs{n}))\)?
\end{opproblem}

\section{Learning-Unlearning Schemes for Halfspaces} 
\label{sec:halfspaces}
Next, we present a set of results for \emph{linear} classifiers which mirror the results of \Cref{sec:bdd_deletions}. The main contribution of this section is a lower bound which shows that the bound of \Cref{thm:bdd_deletions_star_no} is \emph{tight} for linear classifiers (\Cref{thm:linear_central_lb}). Hence, the exponential dependence on $k$ is \emph{necessary} for the relatively simple setting of linear classification. 

Before we proceed, note that the domain size for linear separators is uncountably infinite. To obtain finite bounds, we restrict the domain to a finite set, say the points on a grid. This incurs an additional overhead of $\log (\abs{\cX})$ in memory, which scales as $\approx d$ for natural choices of such a discretization. Our choice for the lower bound is specified subsequently.  First, we recall an upper bound on the hollow star number of linear separators \cite{hellynumber} which utilizes a result from \cite{bkms}. We present a proof in \pref{app:halfspace_proof} for convenience. 

\begin{lemma}[\cite{hellynumber}]
    \label{lem:critic_set_size_linear}
    For the hypothesis class of linear separators, $\mc{H}$, we have:
    \begin{equation*} 
        \hstarno (\mc{H}) \leq d + 2.
    \end{equation*}
\end{lemma}

With \Cref{lem:critic_set_size_linear}, we obtain the following corollary of \Cref{thm:bdd_deletions_star_no}.
\begin{corollary}
    \label{corr:bdd_deletions_linear}
    There exists a central $(0,0)$-\LU scheme with $O((d + 2)^{k+1} \log (\abs{\cX}))$ central memory size.
\end{corollary}

Next, we present the main result of the section which shows that the memory size guarantees of \Cref{corr:bdd_deletions_linear} are essentially \emph{optimal}. Furthermore, our results for a domain of size $2^d$ (hence, $\log (\abs{\mc{X}}) = d$) defined below with $e_i$ denoting the standard basis vectors:

\begin{equation}
    \label{eq:linear_lb_domain}
    \cX \coloneqq \bigcup_{i \in [d]} \cX_k \text{ where } \cX_i \coloneqq \lbrb{\sum_{i = 1}^d \alpha_i e_i: \alpha_i \in \lbrb{0, \frac{1}{i}} \text{ and } \sum_{i = 1}^d \alpha_i = 1}. 
\end{equation}

Our lower bound on the space complexity of an LU scheme for linear classifiers is presented below. As remarked previously, the theorem establishes the optimality of \Cref{corr:bdd_deletions_linear} even for the ubiquitous setting of \emph{linear} classification. 

\begin{theorem}
\label{thm:linear_central_lb}
    Let \(d \geq 1\) and consider any $2 \leq k \leq d-1$. For any $(\epsilon, \delta)$-\LU scheme for halfspaces over the domain $\cX$ defined in \cref{eq:linear_lb_domain} of size $d + \binom{n}{k}$, there exists a worst-case dataset of size at most $n = |\mathcal{X}|$ such that the space complexity of the scheme satisfies,
    \begin{align}
        \s (n) \ge (1 - \Ent(\delta)) \binom{d}{k} - 1.
    \end{align}
    Notably, this lower bound applies even when the scheme must only answer unlearning queries of size at most $k$.
\end{theorem}

As a corollary of the proof, we also obtain the following corollary for the \emph{ticketed} setting by observing that only the tickets of the \emph{positively} labelled points (that is, the tickets of $\{e_i\}_{i \in [d]}$), suffice for reconstructing the precise subset in $\mc{I}$. 

\begin{corollary} \label{thm:linear_tilu_lb}
$d \ge 1$ and $2 \leq k \leq d-1$. Consider any $(\epsilon, \delta)$-\TiLU scheme which supports unlearning queries of size at most $k$, for halfspaces over the domain defined in \cref{eq:linear_lb_domain} of size $d + \binom{n}{k}$. Then, there exists a worst-case dataset of size at most $n = |\mathcal{X}|$ such that the space complexity of the scheme satisfies,
    \begin{align}
        \s (n) = \Omega \left(\frac{ 1 - H( \delta) }{d} \cdot \binom{d}{k} \right).
    \end{align}
\end{corollary}

As remarked previously, the corollary follows from the observation that only the tickets of the points $(e_i, 1)$ are required for the reconstruction procedure defined in the proof of \Cref{thm:linear_central_lb}.

 \paragraph{Does additional structure help for halfspaces?}     

It may be conceivable that above lower bounds can be circumvented when the datasets are restricted to be structured or low-complexity in some manner. Two natural paradigms for halfspaces are, 
 \begin{enumerate}[label=\( \bullet \)]
     \item Assuming the \emph{margin} of the datasets considered are bounded, which functions as a notion of complexity.
     \item Assuming that the dataset is generated by i.i.d.~sampling points from some joint distribution over $\mathcal{X} \times \{ 0,1\}$.
 \end{enumerate}
 Under these assumptions, it is a-priori unclear whether the hardness of unlearning still persists to the same degree. However, as it turns out, the margin of the linear classifiers considered in the proof of \Cref{thm:linear_central_lb} scale polynomially in $d$, and it appears likely that restriction to \emph{constant} margin is required to obtain improved bounds on the space complexity of \LU schemes and their ticketed variants.

 \begin{remark}
     \label{rem:margin}
     The margin of the linear classifiers utilized in the proof of \Cref{thm:linear_central_lb} is $1 / (2d)$. 
 \end{remark}

Furthermore, at the cost of a slightly worse dependency on $k$ (the maximum size of unlearning queries), the same lower bounds can be translated to the case where the datasets are sampled i.i.d.~from an arbitrary distribution. We first introduce the definition of unlearning for such settings: 

\begin{definition}[$(\veps,\delta)$-Learning-Unlearning ($\LU$) scheme for a distributional learning task] \label{def:LU-dist}
Let $\veps, \delta\in (0, 1)$. For a learning task $f : \cZ^\star \to 2^{\cW}$ with respect to a hypothesis class $\mathcal{H}$, an $(\veps,\delta)$-\LU scheme corresponds to a pair of randomized algorithms $(\learn,\unlearn)$. For a known distribution $\mathcal{D}_n$ over datasets of size at most $n$, let $\vD \sim \mathcal{D}_n$. Then,
\begin{itemize}
    \item $\learn(\vD)$ outputs a solution in $f(\vD) \subseteq \cW$, and some auxiliary information of the dataset $\aux(\vD) \in \{0, 1 \}^\star$. Let $\bP_D$ denote the distribution over solutions outputted by $\learn(\vD)$. 
    \item Given a realization of $\vD = \{ (x_i,y_i) \}_{i=1}^{N}$, for any index set $I \subseteq [N]$, and corresponding unlearning query mapping to a subset of the dataset,
    $\vD_I = \{ (x_i,y_i) : i \in I \}$, $\unlearn (\vD_I, \aux)$ outputs a solution in $\cW$ such that, for all subsets $\cW'\subseteq\cW$,
    \[\Pr[\unlearn (\vD_I, \aux(\vD))\in \cW']\le e^{\veps}\bP_{\vD \setminus \vD_I}(\cW')+\delta.\]
\end{itemize}
Here, the randomness is over the auxiliary information outputted by the $\learn$ algorithm and random bits used in the $\unlearn$ algorithm.

The space complexity of an $(\veps,\delta)$-\LU scheme is said to be $\s (n)$ if the learning-unlearning scheme satisfies $\mathbb{E} [ |\aux(\vD)| ] \le \s (n)$, that is, the number of bits of auxiliary information stored is at most $\s (n)$ on average across datasets drawn from $\mathcal{D}_n$. For unlearning queries of size at most $k$, the bounded-query space complexity of the scheme is said to be $\s (n,k)$ if the learning-unlearning scheme satisfies $\mathbb{E} [ |\aux(\vD)| ] \le \s (n,k)$.
\end{definition}

\begin{theorem} 
\label{thm:distributional_halfspace}
Let $d \ge C$ for a sufficiently large constant, and $2 \le k' \le d-2$, and define $n = \binom{d}{k'}$. There exists a distribution $\rho$ supported on $\cZ = \cX \times \{ 0,1 \}$ with support size $n+d$, and a value of $n_0 \in [n/2,n]$ such that for a dataset $\vD$ of $n_0$ i.i.d. samples drawn from $\rho$, any $(\epsilon,\delta)$-\LU scheme which supports unlearning queries of size at most $k = \Theta( k' \log (d))$ must have expected space complexity $\mathbb{E} [|\aux (\vD)|] \ge \Omega ((c_1 - H(\delta)) n) $ for a universal constant $c_1 > 0$.  
\end{theorem}

 \section*{Conclusion} 
    In this paper, we studied the memory complexity of testing the realizability of a data set with respect to a hypothesis class under unlearning constraints. 
    We investigated the relationship between various complexity measures of the hypothesis class and the memory required for learning-unlearning schemes. 
    One interesting direction for future research is to investigate whether low-memory $\LU$ schemes exists for learning tasks beyond realizability testing and empirical risk minimization, especially when the learner is allowed to output a hypothesis within constant error of the ERM; for bounded deletions, algorithms from privacy literature may be useful for this setting. 
 
\section*{Acknowledgements}  
We thank Steve Hanneke for useful discussions, and for pointing out the connection between star number and version-space compressions. AS thanks Pasin Manurangsi, Pritish Kamath, and Sasha Rakhlin for helpful discussions. 

\clearpage

\bibliographystyle{plainnat}  
\bibliography{main.bib} 

\begin{thebibliography}{109}
\providecommand{\natexlab}[1]{#1}
\providecommand{\url}[1]{\texttt{#1}}
\expandafter\ifx\csname urlstyle\endcsname\relax
  \providecommand{\doi}[1]{doi: #1}\else
  \providecommand{\doi}{doi: \begingroup \urlstyle{rm}\Url}\fi

\bibitem[Alon et~al.(1996)Alon, Matias, and Szegedy]{alon1996space}
Noga Alon, Yossi Matias, and Mario Szegedy.
\newblock The space complexity of approximating the frequency moments.
\newblock In \emph{Proceedings of the twenty-eighth annual ACM symposium on
  Theory of computing}, pages 20--29, 1996.

\bibitem[Attiya et~al.(2024)Attiya, Bender, Farach-Colton, Oshman, and
  Schiller]{history_5}
Hagit Attiya, Michael~A Bender, Martin Farach-Colton, Rotem Oshman, and Noa
  Schiller.
\newblock History-independent concurrent objects.
\newblock \emph{arXiv preprint arXiv:2403.14445}, 2024.

\bibitem[Beame et~al.(2018)Beame, Gharan, and Yang]{beame2018time}
Paul Beame, Shayan~Oveis Gharan, and Xin Yang.
\newblock Time-space tradeoffs for learning finite functions from random
  evaluations, with applications to polynomials.
\newblock In \emph{Conference On Learning Theory}, pages 843--856. PMLR, 2018.

\bibitem[Belrose et~al.(2023)Belrose, Schneider-Joseph, Ravfogel, Cotterell,
  Raff, and Biderman]{belrose2023leace}
Nora Belrose, David Schneider-Joseph, Shauli Ravfogel, Ryan Cotterell, Edward
  Raff, and Stella Biderman.
\newblock Leace: Perfect linear concept erasure in closed form.
\newblock \emph{Advances in Neural Information Processing Systems}, 36, 2023.

\bibitem[Bender et~al.(2024)Bender, Farach-Colton, Goodrich, and
  Koml{\'o}s]{history_1}
Michael~A Bender, Mart{\'\i}n Farach-Colton, Michael~T Goodrich, and Hanna
  Koml{\'o}s.
\newblock History-independent dynamic partitioning: Operation-order privacy in
  ordered data structures.
\newblock \emph{Proceedings of the ACM on Management of Data}, 2\penalty0
  (2):\penalty0 1--27, 2024.

\bibitem[Blanchard(2024)]{optimization_4}
Moise Blanchard.
\newblock Gradient descent is pareto-optimal in the oracle complexity and
  memory tradeoff for feasibility problems, 2024.
\newblock URL \url{https://arxiv.org/abs/2404.06720}.

\bibitem[Blelloch and Golovin(2007)]{history_2}
Guy~E Blelloch and Daniel Golovin.
\newblock Strongly history-independent hashing with applications.
\newblock In \emph{48th Annual IEEE Symposium on Foundations of Computer
  Science (FOCS'07)}, pages 272--282. IEEE, 2007.

\bibitem[Bourtoule et~al.(2021)Bourtoule, Chandrasekaran, Choquette-Choo, Jia,
  Travers, Zhang, Lie, and Papernot]{BourtouleCCJTZLP21}
Lucas Bourtoule, Varun Chandrasekaran, Christopher~A Choquette-Choo, Hengrui
  Jia, Adelin Travers, Baiwu Zhang, David Lie, and Nicolas Papernot.
\newblock Machine unlearning.
\newblock In \emph{proceedings of the 42nd IEEE Symposium on Security and
  Privacy}, SP '21. IEEE Computer Society, 2021.

\bibitem[Bousquet et~al.(2020{\natexlab{a}})Bousquet, Hanneke, Moran, and
  Zhivotovskiy]{bousquet2020proper}
Olivier Bousquet, Steve Hanneke, Shay Moran, and Nikita Zhivotovskiy.
\newblock Proper learning, helly number, and an optimal svm bound.
\newblock In \emph{Conference on Learning Theory}, pages 582--609. PMLR,
  2020{\natexlab{a}}.

\bibitem[Bousquet et~al.(2020{\natexlab{b}})Bousquet, Hanneke, Moran, and
  Zhivotovskiy]{hellynumber}
Olivier Bousquet, Steve Hanneke, Shay Moran, and Nikita Zhivotovskiy.
\newblock Proper learning, helly number, and an optimal {SVM} bound.
\newblock In Jacob~D. Abernethy and Shivani Agarwal, editors, \emph{Conference
  on Learning Theory, {COLT} 2020, 9-12 July 2020, Virtual Event [Graz,
  Austria]}, volume 125 of \emph{Proceedings of Machine Learning Research},
  pages 582--609. {PMLR}, 2020{\natexlab{b}}.
\newblock URL \url{http://proceedings.mlr.press/v125/bousquet20a.html}.

\bibitem[Braverman et~al.(2019)Braverman, Kol, Moran, and Saxena]{bkms}
Mark Braverman, Gillat Kol, Shay Moran, and Raghuvansh~R. Saxena.
\newblock Convex set disjointness, distributed learning of halfspaces, and {LP}
  feasibility.
\newblock \emph{CoRR}, abs/1909.03547, 2019.
\newblock URL \url{http://arxiv.org/abs/1909.03547}.

\bibitem[Brophy and Lowd(2021)]{BrophyL21}
Jonathan Brophy and Daniel Lowd.
\newblock Machine unlearning for random forests.
\newblock In \emph{Proceedings of the 38th International Conference on Machine
  Learning}, ICML '21, pages 1092--1104. JMLR, Inc., 2021.

\bibitem[Buchbinder and Petrank(2003)]{history_6}
Niv Buchbinder and Erez Petrank.
\newblock Lower and upper bounds on obtaining history independence.
\newblock In \emph{Advances in Cryptology-CRYPTO 2003: 23rd Annual
  International Cryptology Conference, Santa Barbara, California, USA, August
  17-21, 2003. Proceedings 23}, pages 445--462. Springer, 2003.

\bibitem[Cao and Yang(2015)]{cao2015towards}
Yinzhi Cao and Junfeng Yang.
\newblock Towards making systems forget with machine unlearning.
\newblock In \emph{2015 IEEE symposium on security and privacy}, pages
  463--480. IEEE, 2015.

\bibitem[Carlini et~al.(2022)Carlini, Chien, Nasr, Song, Terzis, and
  Tramer]{carlini2022membership}
Nicholas Carlini, Steve Chien, Milad Nasr, Shuang Song, Andreas Terzis, and
  Florian Tramer.
\newblock Membership inference attacks from first principles.
\newblock In \emph{S \& P}, pages 1897--1914, 2022.

\bibitem[Carlini et~al.(2023{\natexlab{a}})Carlini, Hayes, Nasr, Jagielski,
  Sehwag, Tram{\`e}r, Balle, Ippolito, and Wallace]{carlini2023extracting}
Nicholas Carlini, Jamie Hayes, Milad Nasr, Matthew Jagielski, Vikash Sehwag,
  Florian Tram{\`e}r, Borja Balle, Daphne Ippolito, and Eric Wallace.
\newblock Extracting training data from diffusion models.
\newblock \emph{arXiv preprint arXiv:2301.13188}, 2023{\natexlab{a}}.

\bibitem[Carlini et~al.(2023{\natexlab{b}})Carlini, Ippolito, Jagielski, Lee,
  Tramer, and Zhang]{carlini2022quantifying}
Nicholas Carlini, Daphne Ippolito, Matthew Jagielski, Katherine Lee, Florian
  Tramer, and Chiyuan Zhang.
\newblock Quantifying memorization across neural language models.
\newblock In \emph{ICLR}, 2023{\natexlab{b}}.

\bibitem[Cauwenberghs and Poggio(2000)]{cauwenberghs2001incremental}
Gert Cauwenberghs and Tomaso Poggio.
\newblock Incremental and decremental support vector machine learning.
\newblock \emph{NIPS}, 2000.

\bibitem[CCPA()]{ccpa}
CCPA.
\newblock California consumer privacy act (ccpa).
\newblock https://oag.ca.gov/privacy/ccpa.

\bibitem[Chien et~al.(2024)Chien, Wang, Chen, and Li]{chien2024langevin}
Eli Chien, Haoyu Wang, Ziang Chen, and Pan Li.
\newblock Langevin unlearning: A new perspective of noisy gradient descent for
  machine unlearning.
\newblock \emph{arXiv preprint arXiv:2401.10371}, 2024.

\bibitem[Chourasia et~al.(2023)Chourasia, Shah, and
  Shokri]{chourasia2022forget}
Rishav Chourasia, Neil Shah, and Reza Shokri.
\newblock Forget unlearning: Towards true data-deletion in machine learning.
\newblock In \emph{ICML}, 2023.

\bibitem[Cohen and Nissim(2020)]{cohen2020towards}
Aloni Cohen and Kobbi Nissim.
\newblock Towards formalizing the gdpr’s notion of singling out.
\newblock \emph{Proceedings of the National Academy of Sciences}, 117\penalty0
  (15):\penalty0 8344--8352, 2020.

\bibitem[Cohen et~al.(2023)Cohen, Smith, Swanberg, and
  Vasudevan]{cohen2023control}
Aloni Cohen, Adam Smith, Marika Swanberg, and Prashant~Nalini Vasudevan.
\newblock Control, confidentiality, and the right to be forgotten.
\newblock In \emph{Proceedings of the 2023 ACM SIGSAC Conference on Computer
  and Communications Security}, pages 3358--3372, 2023.

\bibitem[Cook and Weisberg(1980)]{cook1980characterizations}
R~Dennis Cook and Sanford Weisberg.
\newblock Characterizations of an empirical influence function for detecting
  influential cases in regression.
\newblock \emph{Technometrics}, 22\penalty0 (4):\penalty0 495--508, 1980.

\bibitem[Cover and Hellman(1970)]{memory_3}
T~Cover and M~Hellman.
\newblock The two-armed-bandit problem with time-invariant finite memory.
\newblock \emph{IEEE Transactions on Information Theory}, 16\penalty0
  (2):\penalty0 185--195, 1970.

\bibitem[Cover(1969)]{memory_0}
Thomas~M Cover.
\newblock Hypothesis testing with finite statistics.
\newblock \emph{The Annals of Mathematical Statistics}, 40\penalty0
  (3):\penalty0 828--835, 1969.

\bibitem[Cover et~al.(1976)Cover, Freedman, and Hellman]{memory_2}
Thomas~M Cover, Michael~A Freedman, and Martin~E Hellman.
\newblock Optimal finite memory learning algorithms for the finite sample
  problem.
\newblock \emph{Information and Control}, 30\penalty0 (1):\penalty0 49--85,
  1976.

\bibitem[Du et~al.(2019)Du, Chen, Liu, Oak, and Song]{du2019lifelong}
Min Du, Zhi Chen, Chang Liu, Rajvardhan Oak, and Dawn Song.
\newblock Lifelong anomaly detection through unlearning.
\newblock In \emph{CCS}, pages 1283--1297, 2019.

\bibitem[Dukler et~al.(2023)Dukler, Bowman, Achille, Golatkar, Swaminathan, and
  Soatto]{dukler2023safe}
Yonatan Dukler, Benjamin Bowman, Alessandro Achille, Aditya Golatkar, Ashwin
  Swaminathan, and Stefano Soatto.
\newblock {SAFE}: Machine unlearning with shard graphs.
\newblock \emph{arXiv preprint arXiv:2304.13169}, 2023.

\bibitem[Dwork et~al.(2006)Dwork, McSherry, Nissim, and Smith]{DworkMNS06}
Cynthia Dwork, Frank McSherry, Kobbi Nissim, and Adam Smith.
\newblock Calibrating noise to sensitivity in private data analysis.
\newblock In \emph{Proceedings of the 3rd Conference on Theory of
  Cryptography}, TCC '06, pages 265--284, Berlin, Heidelberg, 2006. Springer.

\bibitem[Eisenhofer et~al.(2022)Eisenhofer, Riepel, Chandrasekaran, Ghosh,
  Ohrimenko, and Papernot]{eisenhofer2022verifiable}
Thorsten Eisenhofer, Doreen Riepel, Varun Chandrasekaran, Esha Ghosh, Olga
  Ohrimenko, and Nicolas Papernot.
\newblock Verifiable and provably secure machine unlearning.
\newblock \emph{arXiv preprint arXiv:2210.09126}, 2022.

\bibitem[El-Yaniv and Wiener(2012)]{el2012active}
Ran El-Yaniv and Yair Wiener.
\newblock Active learning via perfect selective classification.
\newblock \emph{The Journal of Machine Learning Research}, 13\penalty0
  (1):\penalty0 255--279, 2012.

\bibitem[El-Yaniv et~al.(2010)]{el2010foundations}
Ran El-Yaniv et~al.
\newblock On the foundations of noise-free selective classification.
\newblock \emph{Journal of Machine Learning Research}, 11\penalty0 (5), 2010.

\bibitem[Eldan and Russinovich(2023)]{eldan2023s}
Ronen Eldan and Mark Russinovich.
\newblock Who's harry potter? approximate unlearning in llms.
\newblock \emph{arXiv preprint arXiv:2310.02238}, 2023.

\bibitem[Flajolet and Martin(1985)]{flajolet1985probabilistic}
Philippe Flajolet and G~Nigel Martin.
\newblock Probabilistic counting algorithms for data base applications.
\newblock \emph{Journal of computer and system sciences}, 31\penalty0
  (2):\penalty0 182--209, 1985.

\bibitem[Floyd and Warmuth(1995)]{floyd1995sample}
Sally Floyd and Manfred Warmuth.
\newblock Sample compression, learnability, and the vapnik-chervonenkis
  dimension.
\newblock \emph{Machine learning}, 21\penalty0 (3):\penalty0 269--304, 1995.

\bibitem[Garg et~al.(2020)Garg, Goldwasser, and Vasudevan]{garg2020formalizing}
Sanjam Garg, Shafi Goldwasser, and Prashant~Nalini Vasudevan.
\newblock Formalizing data deletion in the context of the right to be
  forgotten.
\newblock In \emph{Annual International Conference on the Theory and
  Applications of Cryptographic Techniques}, pages 373--402. Springer, 2020.

\bibitem[Garg et~al.(2018)Garg, Raz, and Tal]{GargRT18}
Sumegha Garg, Ran Raz, and Avishay Tal.
\newblock Extractor-based time-space lower bounds for learning.
\newblock In \emph{Proceedings of the 50th Annual ACM Symposium on the Theory
  of Computing}, STOC '18, pages 990--1002, New York, NY, USA, 2018. ACM.

\bibitem[GDPR()]{GDPR16}
GDPR.
\newblock Regulation {(EU)} 2016/679 of the {European} parliament and of the
  council of 27 {April} 2016.
\newblock \emph{Official Journal of the European Union}.

\bibitem[Ghazi et~al.(2023)Ghazi, Kamath, Kumar, Manurangsi, Sekhari, and
  Zhang]{ghazi2023ticketed}
Badih Ghazi, Pritish Kamath, Ravi Kumar, Pasin Manurangsi, Ayush Sekhari, and
  Chiyuan Zhang.
\newblock Ticketed learning-unlearning schemes, 2023.

\bibitem[Ginart et~al.(2019)Ginart, Guan, Valiant, and Zou]{ginart2019making}
Antonio Ginart, Melody Guan, Gregory Valiant, and James~Y Zou.
\newblock Making {AI} forget you: Data deletion in machine learning.
\newblock In \emph{Advances in Neural Information Processing Systems 32},
  NeurIPS '19, pages 3518--3531. Curran Associates, Inc., 2019.

\bibitem[Godin and Lamontagne(2021)]{godin2021deletion}
Jonathan Godin and Philippe Lamontagne.
\newblock Deletion-compliance in the absence of privacy.
\newblock In \emph{2021 18th International Conference on Privacy, Security and
  Trust (PST)}, pages 1--10. IEEE, 2021.

\bibitem[Goel et~al.(2022)Goel, Prabhu, Sanyal, Lim, Torr, and
  Kumaraguru]{goel2022towards}
Shashwat Goel, Ameya Prabhu, Amartya Sanyal, Ser-Nam Lim, Philip Torr, and
  Ponnurangam Kumaraguru.
\newblock Towards adversarial evaluations for inexact machine unlearning.
\newblock \emph{arXiv preprint arXiv:2201.06640}, 2022.

\bibitem[Goel et~al.(2024)Goel, Prabhu, Torr, Kumaraguru, and
  Sanyal]{goel2024corrective}
Shashwat Goel, Ameya Prabhu, Philip Torr, Ponnurangam Kumaraguru, and Amartya
  Sanyal.
\newblock Corrective machine unlearning.
\newblock \emph{arXiv preprint arXiv:2402.14015}, 2024.

\bibitem[Golatkar et~al.(2020{\natexlab{a}})Golatkar, Achille, and
  Soatto]{Golatkar_2020_CVPR}
Aditya Golatkar, Alessandro Achille, and Stefano Soatto.
\newblock Eternal sunshine of the spotless net: Selective forgetting in deep
  networks.
\newblock In \emph{Proceedings of the IEEE/CVF Conference on Computer Vision
  and Pattern Recognition (CVPR)}, 2020{\natexlab{a}}.

\bibitem[Golatkar et~al.(2020{\natexlab{b}})Golatkar, Achille, and
  Soatto]{golatkar2020forget}
Aditya Golatkar, Alessandro Achille, and Stefano Soatto.
\newblock Forgetting outside the box: Scrubbing deep networks of information
  accessible from input-output observations.
\newblock \emph{arXiv:2003.02960}, 2020{\natexlab{b}}.

\bibitem[Golatkar et~al.(2021)Golatkar, Achille, Ravichandran, Polito, and
  Soatto]{GolatkarARPS21}
Aditya Golatkar, Alessandro Achille, Avinash Ravichandran, Marzia Polito, and
  Stefano Soatto.
\newblock Mixed-privacy forgetting in deep networks.
\newblock In \emph{Proceedings of the 2021 IEEE Computer Society Conference on
  Computer Vision and Pattern Recognition}, CVPR '21, pages 792--801. IEEE
  Computer Society, 2021.

\bibitem[Graves et~al.(2021)Graves, Nagisetty, and Ganesh]{graves2021amnesiac}
Laura Graves, Vineel Nagisetty, and Vijay Ganesh.
\newblock Amnesiac machine learning.
\newblock In \emph{Proceedings of the AAAI Conference on Artificial
  Intelligence}, volume~35, pages 11516--11524, 2021.

\bibitem[Guo et~al.(2020)Guo, Goldstein, Hannun, and Van Der~Maaten]{GuoGHV20}
Chuan Guo, Tom Goldstein, Awni Hannun, and Laurens Van Der~Maaten.
\newblock Certified data removal from machine learning models.
\newblock In \emph{Proceedings of the 37th International Conference on Machine
  Learning}, ICML '20, pages 3832--3842. JMLR, Inc., 2020.

\bibitem[Gupta et~al.(2021)Gupta, Jung, Neel, Roth, Sharifi-Malvajerdi, and
  Waites]{GuptaJNRSW21}
Varun Gupta, Christopher Jung, Seth Neel, Aaron Roth, Saeed Sharifi-Malvajerdi,
  and Chris Waites.
\newblock Adaptive machine unlearning.
\newblock In \emph{Advances in Neural Information Processing Systems 34},
  NeurIPS '21, pages 16319--16330. Curran Associates, Inc., 2021.

\bibitem[Hanneke(2009)]{hanneke2009theoretical}
Steve Hanneke.
\newblock \emph{Theoretical foundations of active learning}.
\newblock Carnegie Mellon University, 2009.

\bibitem[Hanneke(2024)]{hanneke2024eluder}
Steve Hanneke.
\newblock The star number and eluder dimension: Elementary observations about
  the dimensions of disagreement.
\newblock \emph{J. Mach. Learn. Res.}, 247, 2024.

\bibitem[Hanneke and Yang(2015)]{hanneke2015minimax}
Steve Hanneke and Liu Yang.
\newblock Minimax analysis of active learning.
\newblock \emph{J. Mach. Learn. Res.}, 16\penalty0 (1):\penalty0 3487--3602,
  2015.

\bibitem[Hartline et~al.(2005)Hartline, Hong, Mohr, Pentney, and
  Rocke]{history_4}
Jason~D Hartline, Edwin~S Hong, Alexander~E Mohr, William~R Pentney, and
  Emily~C Rocke.
\newblock Characterizing history independent data structures.
\newblock \emph{Algorithmica}, 42:\penalty0 57--74, 2005.

\bibitem[Hellman(1969)]{memory_1}
Martin~Edward Hellman.
\newblock \emph{Learning with finite memory}.
\newblock Stanford University, 1969.

\bibitem[Huang and Canonne(2023)]{huang2023tight}
Yiyang Huang and Cl{\'e}ment~L Canonne.
\newblock Tight bounds for machine unlearning via differential privacy.
\newblock \emph{arXiv:2309.00886}, 2023.

\bibitem[Ippolito et~al.(2022)Ippolito, Tram{\`e}r, Nasr, Zhang, Jagielski,
  Lee, Choquette-Choo, and Carlini]{ippolito2022preventing}
Daphne Ippolito, Florian Tram{\`e}r, Milad Nasr, Chiyuan Zhang, Matthew
  Jagielski, Katherine Lee, Christopher~A Choquette-Choo, and Nicholas Carlini.
\newblock Preventing verbatim memorization in language models gives a false
  sense of privacy.
\newblock \emph{arXiv preprint arXiv:2210.17546}, 2022.

\bibitem[Izzo et~al.(2021)Izzo, Anne~Smart, Chaudhuri, and Zou]{izzo2021}
Zachary Izzo, Mary Anne~Smart, Kamalika Chaudhuri, and James Zou.
\newblock Approximate data deletion from machine learning models.
\newblock In \emph{Proceedings of The 24th International Conference on
  Artificial Intelligence and Statistics (AISTATS)}, 2021.

\bibitem[Jang et~al.(2022)Jang, Yoon, Yang, Cha, Lee, Logeswaran, and
  Seo]{jang2022knowledge}
Joel Jang, Dongkeun Yoon, Sohee Yang, Sungmin Cha, Moontae Lee, Lajanugen
  Logeswaran, and Minjoon Seo.
\newblock Knowledge unlearning for mitigating privacy risks in language models.
\newblock \emph{arXiv preprint arXiv:2210.01504}, 2022.

\bibitem[Kac and Cheung(2002)]{qbinom}
Victor Kac and Pokman Cheung.
\newblock \emph{q-Binomial Coefficients and Linear Algebra over Finite Fields},
  pages 21--26.
\newblock Springer New York, New York, NY, 2002.
\newblock ISBN 978-1-4613-0071-7.
\newblock \doi{10.1007/978-1-4613-0071-7_7}.
\newblock URL \url{https://doi.org/10.1007/978-1-4613-0071-7_7}.

\bibitem[Karasuyama and Takeuchi(2010)]{karasuyama2010multiple}
Masayuki Karasuyama and Ichiro Takeuchi.
\newblock Multiple incremental decremental learning of support vector machines.
\newblock \emph{IEEE Transactions on Neural Networks}, 21\penalty0
  (7):\penalty0 1048--1059, 2010.

\bibitem[Kol et~al.(2017)Kol, Raz, and Tal]{kol2017time}
Gillat Kol, Ran Raz, and Avishay Tal.
\newblock Time-space hardness of learning sparse parities.
\newblock In \emph{Proceedings of the 49th Annual ACM SIGACT Symposium on
  Theory of Computing}, pages 1067--1080, 2017.

\bibitem[Krishna et~al.(2023)Krishna, Ma, and Lakkaraju]{krishna2023towards}
Satyapriya Krishna, Jiaqi Ma, and Himabindu Lakkaraju.
\newblock Towards bridging the gaps between the right to explanation and the
  right to be forgotten.
\newblock In \emph{ICML}, 2023.

\bibitem[Kurmanji et~al.(2023)Kurmanji, Triantafillou, and
  Triantafillou]{kurmanji2023towards}
Meghdad Kurmanji, Peter Triantafillou, and Eleni Triantafillou.
\newblock Towards unbounded machine unlearning.
\newblock \emph{arXiv preprint arXiv:2302.09880}, 2023.

\bibitem[Li et~al.(2022)Li, Kamath, Foster, and Srebro]{li2022understanding}
Gene Li, Pritish Kamath, Dylan~J Foster, and Nati Srebro.
\newblock Understanding the eluder dimension.
\newblock \emph{Advances in Neural Information Processing Systems},
  35:\penalty0 23737--23750, 2022.

\bibitem[Littlestone(1988)]{littlestone1988learning}
Nick Littlestone.
\newblock Learning quickly when irrelevant attributes abound: A new
  linear-threshold algorithm.
\newblock \emph{Machine learning}, 2:\penalty0 285--318, 1988.

\bibitem[Littlestone and Warmuth(1986)]{littlestone1986relating}
Nick Littlestone and Manfred Warmuth.
\newblock Relating data compression and learnability.
\newblock \emph{Unpublished manuscript}, 1986.

\bibitem[Marsden et~al.(2022)Marsden, Sharan, Sidford, and
  Valiant]{optimization_2}
Annie Marsden, Vatsal Sharan, Aaron Sidford, and Gregory Valiant.
\newblock Efficient convex optimization requires superlinear memory.
\newblock In \emph{Conference on Learning Theory}, pages 2390--2430. PMLR,
  2022.

\bibitem[McGregor(2014)]{mcgregor2014graph}
Andrew McGregor.
\newblock Graph stream algorithms: a survey.
\newblock \emph{ACM SIGMOD Record}, 43\penalty0 (1):\penalty0 9--20, 2014.

\bibitem[Misra and Gries(1982)]{misra1982finding}
Jayadev Misra and David Gries.
\newblock Finding repeated elements.
\newblock \emph{Science of computer programming}, 2\penalty0 (2):\penalty0
  143--152, 1982.

\bibitem[Moran and Yehudayoff(2016)]{moran2016sample}
Shay Moran and Amir Yehudayoff.
\newblock Sample compression schemes for vc classes.
\newblock \emph{Journal of the ACM (JACM)}, 63\penalty0 (3):\penalty0 1--10,
  2016.

\bibitem[Moshkovitz and Moshkovitz(2017)]{moshkovitz2017mixing}
Dana Moshkovitz and Michal Moshkovitz.
\newblock Mixing implies lower bounds for space bounded learning.
\newblock In \emph{Conference on Learning Theory}, pages 1516--1566. PMLR,
  2017.

\bibitem[Moshkovitz and Moshkovitz(2018)]{MoshkovitzM18}
Dana Moshkovitz and Michal Moshkovitz.
\newblock Entropy samplers and strong generic lower bounds for space bounded
  learning.
\newblock In \emph{Proceedings of the 9th Conference on Innovations in
  Theoretical Computer Science}, ITCS '18, pages 28:1--28:20, Dagstuhl,
  Germany, 2018. Schloss Dagstuhl--Leibniz-Zentrum fuer Informatik.

\bibitem[Mou et~al.(2020)Mou, Wen, and Chen]{mou2020sample}
Wenlong Mou, Zheng Wen, and Xi~Chen.
\newblock On the sample complexity of reinforcement learning with policy space
  generalization.
\newblock \emph{arXiv preprint arXiv:2008.07353}, 2020.

\bibitem[Munro and Paterson(1980)]{munro1980selection}
J~Ian Munro and Mike~S Paterson.
\newblock Selection and sorting with limited storage.
\newblock \emph{Theoretical computer science}, 12\penalty0 (3):\penalty0
  315--323, 1980.

\bibitem[Muthukrishnan et~al.(2005)]{muthukrishnan2005data}
Shanmugavelayutham Muthukrishnan et~al.
\newblock Data streams: Algorithms and applications.
\newblock \emph{Foundations and Trends{\textregistered} in Theoretical Computer
  Science}, 1\penalty0 (2):\penalty0 117--236, 2005.

\bibitem[Naor and Teague(2001)]{naor2001anti}
Moni Naor and Vanessa Teague.
\newblock Anti-persistence: History independent data structures.
\newblock In \emph{Proceedings of the thirty-third annual ACM symposium on
  Theory of computing}, pages 492--501, 2001.

\bibitem[Naor et~al.(2008)Naor, Segev, and Wieder]{history_3}
Moni Naor, Gil Segev, and Udi Wieder.
\newblock History-independent cuckoo hashing.
\newblock In \emph{Automata, Languages and Programming: 35th International
  Colloquium, ICALP 2008, Reykjavik, Iceland, July 7-11, 2008, Proceedings,
  Part II 35}, pages 631--642. Springer, 2008.

\bibitem[Neel et~al.(2021)Neel, Roth, and Sharifi-Malvajerdi]{NeelRS21}
Seth Neel, Aaron Roth, and Saeed Sharifi-Malvajerdi.
\newblock Descent-to-delete: Gradient-based methods for machine unlearning.
\newblock In \emph{Proceedings of the 32nd International Conference on
  Algorithmic Learning Theory}, ALT '21. JMLR, Inc., 2021.

\bibitem[Nguyen et~al.(2020)Nguyen, Low, and Jaillet]{nguyen2020variational}
Quoc~Phong Nguyen, Bryan Kian~Hsiang Low, and Patrick Jaillet.
\newblock Variational {B}ayesian unlearning.
\newblock In \emph{NeurIPS}, pages 16025--16036, 2020.

\bibitem[Nguyen et~al.(2022)Nguyen, Huynh, Nguyen, Liew, Yin, and
  Nguyen]{nguyen}
Thanh~Tam Nguyen, Thanh~Trung Huynh, Phi~Le Nguyen, Alan Wee-Chung Liew,
  Hongzhi Yin, and Quoc Viet~Hung Nguyen.
\newblock A survey of machine unlearning.
\newblock \emph{arXiv:2209.02299}, 2022.

\bibitem[Pawelczyk et~al.(2023)Pawelczyk, Neel, and
  Lakkaraju]{pawelczyk2023context}
Martin Pawelczyk, Seth Neel, and Himabindu Lakkaraju.
\newblock In-context unlearning: Language models as few shot unlearners.
\newblock \emph{arXiv preprint arXiv:2310.07579}, 2023.

\bibitem[Pawelczyk et~al.(2024)Pawelczyk, Di, Lu, Kamath, Sekhari, and
  Neel]{pawelczyk2024machine}
Martin Pawelczyk, Jimmy~Z Di, Yiwei Lu, Gautam Kamath, Ayush Sekhari, and Seth
  Neel.
\newblock Machine unlearning fails to remove data poisoning attacks.
\newblock \emph{arXiv preprint arXiv:2406.17216}, 2024.

\bibitem[Peng and Rubinstein(2023)]{peng2023near}
Binghui Peng and Aviad Rubinstein.
\newblock Near optimal memory-regret tradeoff for online learning.
\newblock In \emph{2023 IEEE 64th Annual Symposium on Foundations of Computer
  Science (FOCS)}, pages 1171--1194. IEEE, 2023.

\bibitem[Phillips and Phillips(2021)]{phillips2021big}
Jeff~M Phillips and Jeff~M Phillips.
\newblock Big data and sketching.
\newblock \emph{Mathematical Foundations for Data Analysis}, pages 261--281,
  2021.

\bibitem[Ravfogel et~al.(2022{\natexlab{a}})Ravfogel, Twiton, Goldberg, and
  Cotterell]{ravfogel2022linear}
Shauli Ravfogel, Michael Twiton, Yoav Goldberg, and Ryan~D Cotterell.
\newblock Linear adversarial concept erasure.
\newblock In \emph{International Conference on Machine Learning}, pages
  18400--18421. PMLR, 2022{\natexlab{a}}.

\bibitem[Ravfogel et~al.(2022{\natexlab{b}})Ravfogel, Vargas, Goldberg, and
  Cotterell]{ravfogel2022adversarial}
Shauli Ravfogel, Francisco Vargas, Yoav Goldberg, and Ryan Cotterell.
\newblock Adversarial concept erasure in kernel space.
\newblock In \emph{Proceedings of the 2022 Conference on Empirical Methods in
  Natural Language Processing}, pages 6034--6055, 2022{\natexlab{b}}.

\bibitem[Raz(2016)]{Raz16}
Ran Raz.
\newblock Fast learning requires good memory: A time-space lower bound for
  parity learning.
\newblock In \emph{Proceedings of the 57th Annual IEEE Symposium on Foundations
  of Computer Science}, FOCS '16, pages 266--275. IEEE Computer Society, 2016.

\bibitem[Romero et~al.(2007)Romero, Barrio, and
  Belanche]{romero2007incremental}
Enrique Romero, Ignacio Barrio, and Llu{\'\i}s Belanche.
\newblock Incremental and decremental learning for linear support vector
  machines.
\newblock In \emph{ICANN}, pages 209--218, 2007.

\bibitem[Russo and Van~Roy(2013)]{russo2013eluder}
Daniel Russo and Benjamin Van~Roy.
\newblock Eluder dimension and the sample complexity of optimistic exploration.
\newblock \emph{Advances in Neural Information Processing Systems}, 26, 2013.

\bibitem[Sekhari et~al.(2021)Sekhari, Acharya, Kamath, and
  Suresh]{sekhari2021remember}
Ayush Sekhari, Jayadev Acharya, Gautam Kamath, and Ananda~Theertha Suresh.
\newblock Remember what you want to forget: Algorithms for machine unlearning.
\newblock In \emph{NeurIPS}, pages 18075--18086, 2021.

\bibitem[Shamir(2014)]{shamir2014fundamental}
Ohad Shamir.
\newblock Fundamental limits of online and distributed algorithms for
  statistical learning and estimation.
\newblock \emph{Advances in Neural Information Processing Systems}, 27, 2014.

\bibitem[Sharan et~al.(2019)Sharan, Sidford, and Valiant]{optimization_3}
Vatsal Sharan, Aaron Sidford, and Gregory Valiant.
\newblock Memory-sample tradeoffs for linear regression with small error.
\newblock In \emph{Proceedings of the 51st Annual ACM SIGACT Symposium on
  Theory of Computing}, pages 890--901, 2019.

\bibitem[Shokri et~al.(2017)Shokri, Stronati, Song, and
  Shmatikov]{shokri2017membership}
Reza Shokri, Marco Stronati, Congzheng Song, and Vitaly Shmatikov.
\newblock Membership inference attacks against machine learning models.
\newblock In \emph{S \& P}, pages 3--18, 2017.

\bibitem[Srinivas et~al.(2022)Srinivas, Woodruff, Xu, and
  Zhou]{srinivas2022memory}
Vaidehi Srinivas, David~P Woodruff, Ziyu Xu, and Samson Zhou.
\newblock Memory bounds for the experts problem.
\newblock In \emph{Proceedings of the 54th Annual ACM SIGACT Symposium on
  Theory of Computing}, pages 1158--1171, 2022.

\bibitem[Steinhardt and Duchi(2015)]{steinhardt2015minimax}
Jacob Steinhardt and John Duchi.
\newblock Minimax rates for memory-bounded sparse linear regression.
\newblock In \emph{Conference on Learning Theory}, pages 1564--1587. PMLR,
  2015.

\bibitem[Steinhardt et~al.(2016)Steinhardt, Valiant, and
  Wager]{steinhardt2016memory}
Jacob Steinhardt, Gregory Valiant, and Stefan Wager.
\newblock Memory, communication, and statistical queries.
\newblock In \emph{Conference on Learning Theory}, pages 1490--1516. PMLR,
  2016.

\bibitem[Suriyakumar and Wilson(2022)]{suriyakumar2022algorithms}
Vinith~M Suriyakumar and Ashia~C Wilson.
\newblock Algorithms that approximate data removal: New results and
  limitations.
\newblock In \emph{NeurIPS}, 2022.

\bibitem[Thudi et~al.(2022)Thudi, Deza, Chandrasekaran, and
  Papernot]{thudi2022unrolling}
Anvith Thudi, Gabriel Deza, Varun Chandrasekaran, and Nicolas Papernot.
\newblock Unrolling {SGD}: Understanding factors influencing machine
  unlearning.
\newblock In \emph{EuroS\&P}, pages 303--319, 2022.

\bibitem[Tveit et~al.(2003)Tveit, Hetland, and Engum]{tveit2003incremental}
Amund Tveit, Magnus~Lie Hetland, and H{\aa}avard Engum.
\newblock Incremental and decremental proximal support vector classification
  using decay coefficients.
\newblock In \emph{DaWak}, pages 422--429, 2003.

\bibitem[Valiant(1984)]{Valiant84}
Leslie~G. Valiant.
\newblock A theory of the learnable.
\newblock \emph{Communications of the ACM}, 27\penalty0 (11):\penalty0
  1134--1142, 1984.

\bibitem[Vapnik(1995)]{vcdim}
Vladimir~N Vapnik.
\newblock The nature of statistical learning theory, 1995.

\bibitem[Wang et~al.(2023)Wang, Chen, Yuan, Zeng, Wong, and Yin]{wang2023kga}
Lingzhi Wang, Tong Chen, Wei Yuan, Xingshan Zeng, Kam-Fai Wong, and Hongzhi
  Yin.
\newblock Kga: A general machine unlearning framework based on knowledge gap
  alignment.
\newblock \emph{arXiv preprint arXiv:2305.06535}, 2023.

\bibitem[Wiener et~al.(2015)Wiener, Hanneke, and
  El-Yaniv]{wiener2015compression}
Yair Wiener, Steve Hanneke, and Ran El-Yaniv.
\newblock A compression technique for analyzing disagreement-based active
  learning.
\newblock \emph{J. Mach. Learn. Res.}, 16:\penalty0 713--745, 2015.

\bibitem[Woodworth and Srebro(2019)]{optimization_1}
Blake Woodworth and Nathan Srebro.
\newblock Open problem: The oracle complexity of convex optimization with
  limited memory.
\newblock In \emph{Conference on Learning Theory}, pages 3202--3210. PMLR,
  2019.

\bibitem[Wu et~al.(2020)Wu, Dobriban, and Davidson]{wu2020deltagrad}
Yinjun Wu, Edgar Dobriban, and Susan Davidson.
\newblock Deltagrad: Rapid retraining of machine learning models.
\newblock In \emph{International Conference on Machine Learning (ICML)}, 2020.

\bibitem[Zanella-B{\'e}guelin et~al.(2020)Zanella-B{\'e}guelin, Wutschitz,
  Tople, R{\"u}hle, Paverd, Ohrimenko, K{\"o}pf, and
  Brockschmidt]{zanella2020analyzing}
Santiago Zanella-B{\'e}guelin, Lukas Wutschitz, Shruti Tople, Victor R{\"u}hle,
  Andrew Paverd, Olga Ohrimenko, Boris K{\"o}pf, and Marc Brockschmidt.
\newblock Analyzing information leakage of updates to natural language models.
\newblock In \emph{CCS}, pages 363--375, 2020.

\bibitem[Zhang and Zhang(2021)]{zhang2021rethinking}
Rui Zhang and Shihua Zhang.
\newblock Rethinking influence functions of neural networks in the
  over-parameterized regime.
\newblock In \emph{Proceedings of the AAAI Conference on Artificial
  Intelligence (AAAI)}, 2021.

\bibitem[Zhang et~al.(2024)Zhang, Lin, Bai, and Mei]{zhang2024negative}
Ruiqi Zhang, Licong Lin, Yu~Bai, and Song Mei.
\newblock Negative preference optimization: From catastrophic collapse to
  effective unlearning.
\newblock \emph{arXiv preprint arXiv:2404.05868}, 2024.

\end{thebibliography}

\clearpage 
\appendix 

\renewcommand{\contentsname}{Contents of Appendix}
\tableofcontents 
\addtocontents{toc}{\protect\setcounter{tocdepth}{3}} 
\clearpage

\setlength{\parindent}{0pt}
\setlength{\parskip}{0.1em} 

\section{Discussion of Related Works} \label{app:related_works} 
\textbf{Machine unlearning and the ``Right to be Forgotten''.} Over the past few years, there has been tremendous research on  
    formalizing the notion of ``right to be forgotten'' and ``data deletion'' under different circumstances \citep{dukler2023safe, krishna2023towards, cohen2023control, eisenhofer2022verifiable, garg2020formalizing, cohen2020towards,godin2021deletion}. 
    Taking inspiration from cryptography, \cite{garg2020formalizing} define the notion of deletion-compliance, using a ``leave no trace'' approach, such that post  deletion, the state of the world should be  indistinguishable from the state of the world if the data point was never in the data set at all. 
       \cite{cohen2023control} define a notion called deletion-as-control which conceptually relaxes the notion of deletion-compliance by only requiring the effects of the data that requested to be deleted to be absent from the updated model. Our unlearning setting is closer to that of \cite{cohen2023control}.    
   
In terms of the objectives of data deletion, many works focus on removing the influence of a particular subset of training data points~\citep{ginart2019making,wu2020deltagrad,Golatkar_2020_CVPR,golatkar2020forget,BourtouleCCJTZLP21,izzo2021,NeelRS21,sekhari2021remember,jang2022knowledge,huang2023tight,wang2023kga}, while others try to remove specific concepts~\citep{ravfogel2022linear,ravfogel2022adversarial,belrose2023leace}, or corrupted/poisoned samples \citep{goel2024corrective, pawelczyk2024machine}. A detailed comparison between different definitions and objectives can be found in \cite{cohen2023control}.

\paragraph{Reducing the cost of retraining.}  There is a vast literature on algorithms and techniques that can be unlearned without incurring the cost of retraining from scratch. On one hand, we have rigorous methods for exact unlearning  \citep{BourtouleCCJTZLP21, thudi2022unrolling}, that exactly recover the retrained-from-scratch hypothesis; such methods are typically quite memory intensive and store multiple models trained on different subsets of the training dataset.  On the other hand, there are many algorithms for approximate unlearning  \citep{ginart2019making,sekhari2021remember,NeelRS21, GuptaJNRSW21, chourasia2022forget}, where the goal is to recover the retrained-from-scratch hypothesis probabilistically under a definition inspired from differential privacy~\citep{DworkMNS06}; such algorithms are generally space efficient but are restricted to specialized classes like convex models \citep{GuoGHV20, sekhari2021remember, NeelRS21, suriyakumar2022algorithms, huang2023tight}, linear regression \citep{cook1980characterizations, GuoGHV20,izzo2021}, support vector machines \citep{cauwenberghs2001incremental, tveit2003incremental, karasuyama2010multiple, romero2007incremental}, kernel methods \citep{zhang2021rethinking}, random forests~\citep{BrophyL21}, etc. In addition to these theoretically rigorous works, there are also many empirical approaches, that  theoretical guarantees but are implementable for various large scale deep learning settings \citep{du2019lifelong, Golatkar_2020_CVPR,goel2022towards,kurmanji2023towards,ravfogel2022linear,ravfogel2022adversarial,belrose2023leace, du2019lifelong,GolatkarARPS21, nguyen2020variational, graves2021amnesiac, zanella2020analyzing, eldan2023s, pawelczyk2023context, zhang2024negative}. 
            
    \paragraph{Learning and unlearning under memory constraints.}  
 A few recent works have considered the memory footprint of machine unlearning, and the associated tradeoffs due to memory constraints \citep{sekhari2021remember, ghazi2023ticketed}. 
 \cite{ghazi2023ticketed} introduced the notion of central and ticketed learning-unlearning schemes, and mergeable hypothesis classes, and provided space-efficient unlearning schemes for various specific hypothesis classes. 
Building on their work, our work provides a characterization of the memory complexity of unlearning in terms of natural combinatorial notions of a hypothesis class.  

The memory complexity of learning and hypothesis testing problems have been studied for several decades in the statistics and information theory communities \citep{memory_0,memory_1, memory_2, memory_3}. 
There was a resurgence in interest in the memory complexity of learning in the past decade \citep{shamir2014fundamental, steinhardt2015minimax, steinhardt2016memory}, leading to a breakthrough result of \cite{Raz16}, who showed that learning parities needs either quadratic memory or exponentially many samples.
Subsequently, these infeasibility results under memory constraints have been extended to various problems in learning \citep{ moshkovitz2017mixing,kol2017time, MoshkovitzM18,GargRT18, beame2018time,srinivas2022memory,peng2023near} and optimization \cite{optimization_1,optimization_3,optimization_2,optimization_4}.

\paragraph{Connection to history independent data structures.} A more traditional concept that is related to notion of machine unlearning is \textit{history independent data structures} \citep{naor2001anti,history_1, history_2, history_3, history_4, history_5, history_6}. 
At an intuitive level, our notion of a learning-unlearning scheme can be seen as enforcing history independence between the state obtained by first adding samples followed by removal, and recomputing the data structure from scratch. However, we are not aware of any direct applications of the existing literature on history independence towards machine unlearning. %
    \paragraph{Connection to privacy.} There is a rich literature in  ML on protecting privacy of individuals, with the most popular notion being that of Differential Privacy (DP) \citep{DworkMNS06}. Under 
    DP, our objective is to develop learning algorithms whose output model is probabilistically indistinguishable from a model that we would have obtained by training without a particular data point. Thus, DP implies machine unlearning in the sense that differentially private learning guarantees that any single data point is already forgotten from the trained model. However, one can hope to do better since DP guarantees privacy for all points simultaneously, whereas in machine unlearning the updated hypothesis may depend on the samples that requested to be removed \citep{sekhari2021remember}. 
    That being said, the connection between DP and machine unlearning remains largely unexplored, with only a few recent works \citep{GuptaJNRSW21, chien2024langevin}. 
        
\paragraph{Streaming algorithms and data deletion.} Another related line of work is on streaming algorithms which process data samples one at a time  while only using a small amount of memory. 
Starting from the early works of \cite{munro1980selection, flajolet1985probabilistic,misra1982finding, alon1996space}, streaming algorithms have become an rich and extensive area under which several problems have been studied (See \cite{muthukrishnan2005data,mcgregor2014graph,phillips2021big} and references therein). 
Though conceptually related to the setting we are considering, we are not aware of any direct application of techniques in this setting to the problems considered in our paper.
One concrete difference comes from the fact that we restrict ourselves to a one-shot unlearning setting, where all the deletion requests arrive at the same time and there is a single round of unlearning.
It would be interesting to extend our techniques to streaming deletion requests, e.g.~as considered in \citet{GuptaJNRSW21, chien2024langevin}.

\section{Relation Between  Various  Combinatorial Dimensions}

\begin{definition}[Littlestone dimension] 
    \label{def:littlestone_app} 
Given a hypothesis class \(\cH\) that maps elements from a domain \(\cX\) to \(\crl{0, 1}\), the  littlestone tree is a rooted binary tree, where each node is labeled by a point \(x \in \cX\),  the left edge is labeled by \(y = 0\), and the right eight is labeled by \(y = 1\). The tree is said to be shattered by a hypothesis class \(\cH\) if, for every branch in the tree there exists a concept \(h \in \cH\) which is consistent with the edges of the branch (i.e.~for every \(x\) and the decision \(y\) on this branch, we have \(h(x) = y\)). 

The Littlestone dimension of the hypothesis class \(\cH\), denoted by \(\litl(\cH)\),  is the largest \(l \in \bbN \cap \crl{0}\) for which there exists a complete littlestone tree of depth \(\ell\) all of whose branches are shattered by \(\cH\). 
\end{definition}

When clear from the context, we use the notation \(\litl\) to denote the \(\litl(\cH)\). 
When clear from the context, we use the notation \(\litl\) to denote the \(\litl(\cH)\). 

\begin{lemma} [{\cite{li2022understanding, hanneke2024eluder}}] 
\label{lem:relate_dimension} 
For any hypothesis class \(\cH\), 
\begin{enumerate}[label=\(\bullet\)] 
	\item $d(\cH) \leq \starno(\cH) \leq \Eld(\cH) \leq \abs{\cH} - 1,$ 
        \item \(\litl(\cH) \leq \Eld(\cH)\), 
	\item $\log\prn*{\abs{\cH}} \leq \Eld(\cH), $
\end{enumerate}
where \(d(\cH)\), \(\starno(\cH)\), \(\litl(\cH)\), and \(\Eld(\cH)\) denote the \VC-dimension, star number, Littlestone dimension and eluder dimension of \(\cH\), respectively. 
\end{lemma} 

The next result relates the hollow-star number for a hypothesis class, to its star number. 
\begin{lemma}
    \label{lem:hstar_star}
    For any hypothesis class $\cH$, the star number is always larger than the hollow star number, i.e.~
    \begin{equation*}
        \hstarno(\cH) - 1 \leq \starno(\cH). 
    \end{equation*}
\end{lemma}
\begin{proof}
    We will show that for any hollow star set of size $l$, there exists a star set of size $l - 1$. Let $S = \{(x_1, y_1), \dots, (x_l, y_l)\}$ be a hollow star set. By definition, $S$ is not realizable and for any $i \in [l]$:
    \begin{equation*} 
        S_i = \{(x_j, y_j)\}_{j \neq i} \bigcup \{(x_i, \bar{y}_i)\}
    \end{equation*}
    is realizable. Consider the set $S^* = \{(x_1, y_1), \dots, (x_{l - 1}, y_{l - 1})\}$. From the fact that $S_n$ is realizable, $S^*$ is realizable. Furthermore, from the fact that $S_i$ is realizable for $i \neq n$, $S^*$ is also a star set. Hence, there exists a star set of size $l - 1$. This concludes the proof.
\end{proof}

The following technical result shows that the minimum identification set and the eluder dimension are incomparable. 

\begin{proposition} 
    \label{prop:mssvseluder}
    For any $n \in \mathbb{N}$,     \begin{enumerate}[label=\(\bullet\)] 
        \item There exists a domain, $\cX_1$, and hypothesis set $\cH_1$ such that $\abs{\cX_1} = n, \abs{\cH_1} = 2n$ such that:
        \begin{equation*}
            \mis (\cH_1) = n \text{ and } \Eld (\cH_1) \leq 32 \log (n).
        \end{equation*}
        \item There exists a domain, $\cX_2$, and hypothesis set $\cH_2$ such that $\abs{\cX_2} = n + \log (n), \abs{\cH_2} = n$ such that:
        \begin{equation*}
            \mis (\cH_1) = \log (n) \text{ and } \Eld (\cH_1) \geq n - 1.
        \end{equation*}
    \end{enumerate}
\end{proposition}

\begin{proof}
    We first establish the sub-claim that the minimum identification set may be substantially larger than eluder dimension. We will use a randomized construction to show the existence of such a set. Our domain will correspond to the set of integers up to $n$; i.e $\mc{X} \coloneqq [n]$, and the hypothesis set will consist of $2n$ functions indexed by $[n] \times \{0, 1\}$, $\mc{H} \coloneqq \{h_{i, l}\}_{i \in [n], l \in \{0, 1\}}$. For each $i \in [n]$, $h_{i, 0}, h_{i, 1}$ are constructed as follows:
    \begin{gather*}
        h_{i, 0} (i) = 0 \text{ and } h_{i, 1} (i) = 1 \\
        \forall j \neq i: h_{i, 0} (j) = h_{i, 0} (j) = W_{i, j} \text{ where } W_{i, j} \overset{\text{iid}}{\thicksim} \mathrm{Unif} (\{0, 1\}).  
    \end{gather*}
    We will show that with non-zero probability, the constructed hypothesis set has eluder dimension bounded above by $O(\log (n))$ and minimum identification set bounded below by $\Omega(n)$. We start with the simpler lower bound on $\mis$.
    \begin{claim}
        \label{claim:lb_bnd_mss_mss_ed_comp}
        We have:
        \begin{equation*} 
            \mis = n.
        \end{equation*}
    \end{claim}
    \begin{proof}
        Suppose $S \subset [n]$ with $S \neq [n]$ be a minimum identification set. This is a contradiction as for any $i \notin S$, $h_{i, 0}$ and $h_{i, 1}$ realize the same values on the elements in $\mc{X}$.
    \end{proof}
    We next show that the eluder dimension is small. Here, we assume that in addition to the random variables $W_{i, j}$ drawn in the construction of $\mc{H}$, we also assume for the sake of convenience that $W_{i, i}$ is drawn. We prove the following simple
    \begin{claim}
        \label{claim:w_i_bdd_equal}
        We have:
        \begin{equation*}
            \Pr \lbrb{\exists S_{\cX}, S_\cH: \abs{S_{\cX}}, \abs{S_{\cH}} \geq 4 \log (n) \text{ and } \forall i \in S_\cX, \forall j, j' \in S_{\cH}, W_{i, j} = W_{i, j'}} < 1.
        \end{equation*}
    \end{claim}
    \begin{proof}
         Let $l = 4 \log (n)$. We will show this via a union bound over all possible subsets $S_{\cX}, S_\cH$. First, fix two sets $S_\cX$ and $S_\cH$ of size $l$. Now, we have that
         \begin{equation*}
             \Pr \lbrb{W_{i, j} = W_{i, j'} \text{ for all } i \in S_\cX, j,j' \in S_\cH} = 2^{-l (l - 1)}.
         \end{equation*}
         Hence, we get by a union bound over all such possible $S_\cX, S_\cH$ that the probability that there exists such a pair of sets is upper bounded by:
         \begin{equation*}
             \binom{n}{l} \cdot \binom{n}{l} \cdot 2^{-l (l - 1)} \leq n^{2l} \cdot 2^{-l (l - 1)} < 1 
         \end{equation*}
         for our value of $l$.
    \end{proof}
    We will condition on the events of \Cref{claim:w_i_bdd_equal}. Suppose for the sake of contradiction that there existed an eluder sequence over the sequence of elements $\{i_1, \dots, i_l\}$ in $[n]$ with $l = 32 \log (n)$. Let the base function of the sequence be $h_0$ and $h_1, \dots, h_l$ be the deviations from $h_0$ with $h_j (i_j) \neq h_0 (i_j)$ and $h_j (i_k) = h_0 (i_k)$ for all $k < j$. Now, observe the following:
    \begin{equation*}
        \forall j > 4 \log (n), k \leq 4 \log (n): h_{j} (i_k) = h_0 (i_k).
    \end{equation*}
    Now, define the sets $\cS'_\cX, S_\cH$ as follows:
    \begin{equation*}
        \cS'_\cH = \{h_j\}_{j > 4 \log (n)} \setminus \{h_{i_k, y}\}_{k \in [4\log (n)], y \in \{0, 1\}} \text{ and } S_\cX = \{i_k\}_{k \leq 4 \log (n)}.
    \end{equation*}
    That is, $\cS'_\cH$ consists of the hypotheses whose values are \emph{entirely} determined by the random variables $W_{i, j}$ on the indices in $S_\cX$. We further prune $\cS'_{\cH}$ to remove duplicates as follows:
    \begin{equation*}
        S_\cH = \cS'_{\cH} \setminus \{h_{i, 1}: h_{i, 0} \in \cS'_\cH\}.
    \end{equation*}
    Since this process removes at most one hypothesis of a pair already existing in $\cS'_\cH$, we have:
    \begin{equation*}
        \abs{S_{\cH}} \geq \frac{\abs{\cS'_\cH}}{2} \geq 10 \log (n).
    \end{equation*}
    \begin{equation*}
        h \in S_\cH, i \in S_{\cX}: h(i) = W_{j_h, i} \text{ for some } j_h.
    \end{equation*}
    Observing that the index $j_h$ is \emph{unique} for each element in $S_\cH$ as the duplicates are removed, we have obtained $S_\cX, S_\cH$ with size larger than $4\log (n)$ such that for all $i \in S_{\cX}, j, j' \in S_{cH}$:
    \begin{equation*}
        W_{i, j} = W_{i, j'}
    \end{equation*}
    yielding a contradiction with \Cref{claim:w_i_bdd_equal}.

    We now show that the eluder dimension may be much larger than the minimum identification set. We will define a hypothesis set of size $n$ over a domain of size $n + \log (n)$. Denoting the domain as $\cX = \{a_i\}_{i \in [n]} \cup \{b_j\}_{j \in \log (n)}$, the hypotheses $\cH = \{h_i\}_{i \in [n]}$ are defined as follows:
    \begin{equation*}
        h_i (a_j) = 
        \begin{cases}
            1 & \text{if } j = i \\
            0 & \text{otherwise}
        \end{cases} 
        \text{ and }
        h_i (b_j) = \mathrm{Parity}\lprp{\floor*{\frac{i}{2^j}}}
    \end{equation*}
    where $\mathrm{Parity}$ denotes the parity of the input ($0$ if the input is even and odd otherwise). That is, on the inputs $b_j$, $h_i$ evaluates to the elements of binary representation of $i$. Hence, the set $\{b_j\}_{j \in [\log (n)]}$ is a minimum identification set for $\cH$. The minimality follows from the fact that the minimum identification set must be of size at least $\log (n)$. Finally, observe that $\cH$ has an eluder sequence of size $n - 1$ on the elements $\{a_1, \dots, a_{n - 1}\}$ with $h_n$ as the base function and $h_1, \dots, h_{n - 1}$ being the deviations from the base function on the sequence.
\end{proof}

\section{Missing Details from \pref{sec:VCnotenough} } 

\subsection{Proof of \pref{theorem:LU=>VC}} 

\begin{proof}[Proof of \pref{theorem:LU=>VC}] %
Let $d=\VCd(\mcH)$. Consider the collection of datasets $\mathcal{D} = \{D_z : z \in \{ 0,1 \}^d\}$; for any maximal set of points shattered by $\mathcal{H}$, $\{ x_1,\cdots,x_d \}$, $D_z$ is constructed by placing two coincident points on each $x_i$, one labeled $1$ and the other labeled as $z_i$. \\ 

\noindent\textit{Unlearning queries.} For all $i \in [d]$, an unlearning query $U_i$ is a collection of points $\{ (x_j,1) : j \in [d] \setminus \{i\} \}$. Namely, one copy of all the points with label $1$ are removed, except for at the point $x_i$. It is important to note that the points being deleted in the unlearning query are not a function of the $z_i$'s which are a priori unknown.
For every starting dataset $D_z$, since $\{ x_1,\dots,x_d \}$ is a shattered set, the surviving dataset is $\mathcal{H}$-realizable if and only if there are no coincident points with opposite labels. Under the unlearning query $U_i$, this is possible only if $z_i = 1$. Thus, the outputs of the $\unlearn$ algorithm on these queries can be used to infer information about the starting dataset. 

Formally, let $\vZ$ be a random variable with uniform distribution over $\{0,1\}^d$. Let $\vD$ be a random variable over datasets, taking value $D_z$ when $\vZ=z$. Let $(\learn,\unlearn)$ be an $(\veps,\delta)$-\LU scheme for $\mcH$-realizability testing. As for the learning task of realizability testing, $\cW=\{\yes,\no\}$, $\learn(D)$ always outputs $\yes$ if $D$ is $\mcH$-realizable and $\no$ otherwise. Let $\vA$ be a random variable for the auxiliary information outputted by $\learn$ algorithm on dataset $\vD$. By \Cref{def:LU}, for all datasets $D_z\in \mathcal{D}$ and unlearning queries $\{U_i\}_{i\in[d]}$, if $D_z\setminus U_i$ is realizable -- $z_i=1$, then
\[\Pr[\unlearn(U_i,\aux(D_z))=\no]\le e^{\veps} \cdot 0 +\delta=\delta.\]
Similarly, if $D_z\setminus U_i$ is not realizable -- $z_i=0$, then
\[\Pr[\unlearn(U_i,\aux(D_z))=\yes]\le \delta.\]

Using Bayes' rule, it's easy to see that probability of $z_i=1$ (or $0$) conditioned on $\unlearn(U_i,\aux(D_z))=\yes$ (or $\no$) is greater than $1-\delta$. Thus, for $\delta<1/2$,

\begin{equation}\label{eq:VC1}\Ent\left(\vZ_i\mid\unlearn(U_i,\vA)\right)\le \Ent(1-\delta)\implies \I(\vZ_i;\unlearn(U_i,\vA))\ge (1-\Ent(1-\delta)).
\end{equation}
We will use this to show that $\vA$ contains $\Omega(1)$ bits of information about $\vZ_i$ for all $i\in[d]$, and must use $\Omega(d)$ bits of space. We use $|\vA|$ to denote the size of auxiliary information on datasets in $\mathcal{D}$. Formally, 
\begin{align*}
|\vA|&\ge \I(\vZ;\vA)\tag{as $\Ent(\vA)\le |\vA|$}\\
&=\sum_{i=1}^d\I(\vZ_i;\vA\mid \vZ_{<i})\tag{Chain rule}\\
&\ge \sum_{i=1}^d\I(\vZ_i;\vA)\tag{as $\I(\vZ_i;\vZ_{<i})=0$}\\
&=\sum_{i=1}^d\left(\I(\vZ_i;\vA,\unlearn(U_i,\vA))-\I(\vZ_i;\unlearn(U_i,\vA)\mid \vA)\right)\\
&\ge \sum_{i=1}^d\left(\I(\vZ_i;\unlearn(U_i,\vA))-\I(\vZ_i;\unlearn(U_i,\vA)\mid \vA)\right)\\
&\ge \sum_{i=1}^d(1-\Ent(1-\delta))\tag{Explained below}\\
&=d\cdot (1-\Ent(\delta))
\end{align*}
 The second last inequality follows from Equation \eqref{eq:VC1} and the fact that $\I(\vZ_i;\unlearn(U_i,\vA)\mid \vA)=0$. This is because $\unlearn(U_i,\vA)$ is completely determined by $\vA$ and the random bits used by the $\unlearn$ algorithm, which are independent of $\vZ_i$ (and $\vA$). Hence, any $(\veps,\delta)$-\LU scheme must have space complexity of at least $d(1-\Ent(\delta))$ bits on datasets of size at most $2d$.
\end{proof}

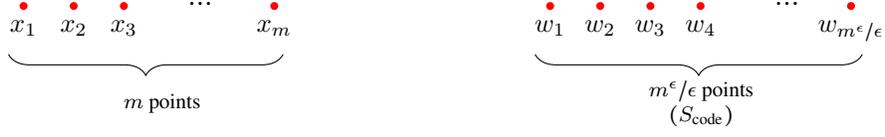
\begin{figure}
    \centering
    \begin{tikzpicture}[auto,node distance=2cm]

  \foreach [evaluate={\m=int(mod(\x,2))};] \x in {0,1,2,5} {
    \ifthenelse{\m = 0}{\node[circle, fill=red, inner sep=0pt, minimum size =3pt] (newred\x) at (-3+\x/1.5,1.5) {};}{\node[circle, fill=red, inner sep=0pt, minimum size =3pt] (newblue\x) at (-3+\x/1.5,1.5) {};}
    }
    \node[below = 0cm of newred0] {\small $x_1$};
    \node[below = 0cm of newblue1] {\small $x_2$};
    \node[below = 0cm of newred2] {\small $x_3$};
    \node[below = 0cm of newblue5] {\small $x_m$};
    
    \node (dots) at (-0.65,1.5) {$\cdots$};
    \draw [decorate,decoration={brace,mirror,amplitude=8pt,raise=4ex}]
  (-3.2,1.55) -- (0.45,1.55) node[left,yshift=-3.5em,xshift=-2.5em] {$\substack{m \text{ points}\\}$};

    \foreach [evaluate={\m=int(mod(\x,2))};] \x in {0,1,2,3,6} {
    \ifthenelse{\m = 0}{\node[circle, fill=red, inner sep=0pt, minimum size =3pt] (red\x) at (4+\x/1.5,1.5) {};}{\node[circle, fill=red, inner sep=0pt, minimum size =3pt] (blue\x) at (4+\x/1.5,1.5) {};}
    }
    \node[below = 0cm of red0] {\small $w_1$};
    \node[below = 0cm of blue1] {\small $w_2$};
    \node[below = 0cm of red2] {\small $w_3$};
    \node[below = 0cm of blue3] {\small $w_4$};
    \node[below = 0cm of red6] {\small $w_{m^\epsilon/\epsilon}$};
    
    \node (dots) at (7.15,1.5) {$\cdots$};
    \draw [decorate,decoration={brace,mirror,amplitude=8pt,raise=4ex}]
  (3.8,1.55) -- (8.15,1.55) node[left,yshift=-3.5em,xshift=-3.4em] {$\substack{m^\epsilon/\epsilon \text{ points}\\ (\Sc)}$};
    \end{tikzpicture}
    \caption{Lower bound for VC classes}
    \label{fig:lb-VC-central}
\end{figure}

\subsection{Proof of \pref{theorem:VClb}} 

\begin{proof}[Proof of \pref{theorem:VClb}] 
We will construct the lower bound by considering a collection of datasets and arguing that any \LU / \TiLU scheme which supports unlearning queries on $1/\epsilon$ points can be used to essentially infer the entire dataset. For the ticketed lower bound, we furthermore show that it suffices to select the $1/\epsilon$-sized unlearning query from a small subset of $\approx n^\epsilon/\epsilon$ points. In particular, the total information contained within the tickets of these few points, put together, can recover the entire dataset, which will imply that the maximum ticket size will have to be large.\\

\noindent\textit{Dataset.} The collection of datasets we consider is denoted $\mathcal{D} = \{ D(z) : z \in \{ 0,1 \}^m \}$ and illustrated in \Cref{fig:lb-VC-central}. All of the datasets in $\mathcal{D}$ contain a fixed collection of $m^\epsilon/\epsilon$ points denoted $\Sc$, all labelled $0$. As a function of the hidden vector $z \in \{ 0,1\}^m$, the dataset contains additional points. If $z_i = 1$, the dataset contains an additional point $x_i$ labelled $0$. The total number of points contained in datasets in $\mathcal{D}$ are between $m^\epsilon/\epsilon$ and $m+m^\epsilon/\epsilon$ which are both $O (m)$ for constant $\epsilon \in (0,1]$.\\

\noindent\textit{Hypothesis class.} The hypothesis class $\mathcal{H}$ we consider has $m$ functions, $\{ h_i : i \in [m] \}$. The function $h_i$, satisfies the following properties, 
\begin{enumerate}
    \item[(i)] For any $j \ne i$, $h_i$ labels $x_j$ as $0$. 
    \item[(ii)] For $j = i$, $h_i$ labels $x_i$ as $1$.
    \item[(iii)] For the points in $\Sc$, consider any injective function $\sigma$ from $[m] \to \binom{\Sc}{1/\epsilon}$, the $1/\epsilon$-sized subsets of $\Sc$. Such an injective function exists since $|\Sc| = m^\epsilon/\epsilon$ and therefore the size of the co-domain $\binom{m^\epsilon/\epsilon}{1/\epsilon} \ge m$ is larger than that of the domain. $h_i$ labels the points in $\Sc$ as $0$ except for all the points in $\sigma (i)$, which are labelled as $1$.
\end{enumerate}

\noindent \textit{Littlestone dimension of $\mathcal{H}$.} The Littlestone dimension of $\mathcal{H}$ is at most $1/\epsilon+1$. Consider the online learning algorithm which labels every point in the dataset with the label $0$. Notice that any labeling of the dataset consistent with some hypothesis $h \in \mathcal{H}_i$ for some $i$, labels at most $1/\epsilon+1$ points as $1$. This means that the online learning algorithm we consider makes at most $1/\epsilon+1$. The Littlestone dimension of the class, which captures the optimal mistake bound of realizable online learning for the class is therefore at most $1/\epsilon+1$.\\

\noindent\textit{Unlearning queries.} The unlearning queries we consider correspond to choosing an arbitrary subset of $1/\epsilon$ points from $\Sc$. The learner iterates over all such subsets to get the set of unlearning queries.\\

\noindent\textit{Unlearning $\implies$ $z$ can be recovered.} Consider an unlearning query $U_i$ of $1/\epsilon$ points from $\Sc$ corresponding to $\sigma(i)$ for some $i \in [m]$. The hypothesis $h_i$ is potentially the only one which can label all the remaining points in $\Sc$ correctly. The only points labeled incorrectly by $h_i$ is potentially the point $x_i$. In particular, since $h_i$ labels this point as $1$, after the unlearning query $U_i$ is made, the dataset is $\mathcal{H}$-realizable if and only if $x_i$ did not exist in it to begin with. In particular, under the unlearning query $U_i$, the realizability of the dataset answers whether $z_i = 0$ or $z_i = 1$. Therefore, the hidden vector $(z_1,\cdots,z_m)$ can be recovered by considering all such unlearning queries $\{ U_i : i \in [m] \}$.

We now formalize this argument. 
Let $\vZ$ be a random variable with uniform distribution over $\{0,1\}^m$. Let $\vD$ be a random variable over datasets, taking value $D_z$ when $\vZ=z$. 
Let $(\learn,\unlearn)$ be an $(\veps,\delta)$-\LU scheme for $\mcH$-realizability testing. As for the learning task of realizability testing, $\cW=\{\yes,\no\}$, $\learn(D)$ always outputs $\yes$ if $D$ is $\mcH$-realizable and $\no$ otherwise. Let $\vA$ be a random variable for the auxiliary information outputted by $\learn$ algorithm on dataset $\vD$. By \Cref{def:LU}, for all datasets $D_z\in \mathcal{D}$ and unlearning queries $\{U_i\}_{i\in[m]}$, if $D_z\setminus U_i$ is realizable -- $z_i=1$, then
\begin{align}
\Pr[\unlearn(U_i,\aux(D_z))=\no]\le e^{\veps} \cdot 0 +\delta=\delta.
\end{align}
Similarly, if $D_z\setminus U_i$ is not realizable -- $z_i=0$, then
\begin{align}
\Pr[\unlearn(U_i,\aux(D_z))=\yes]\le \delta.
\end{align}
Thus, using Bayes' rule, it's easy to see that probability of $z_i=1$ (or $0$) conditioned on \\ $\unlearn(U_i,\aux(D_z))=\yes$ (or $\no$) is greater than $1-\delta$. Thus, for $\delta<1/2$,
\begin{align}
\Ent\left(\vZ_i\mid\unlearn(U_i,\vA)\right)\le \Ent(1-\delta)\implies \Ent\left(\vZ_i\mid\vA\right)\le \Ent(1-\delta),
\end{align}
where the second implication follows from data processing inequality.
Thus, $\vA$ contains $\Omega(1)$ bits of information about $\vZ_i$ for all $i\in[m]$, and must use $\Omega(m)$ bits of space. Formally,
\begin{align*}
|\vA|&\ge \I(\vZ;\vA)\tag{as $\Ent(\vA)\le |\vA|$}\\
&=\sum_{i=1}^m\I(\vZ_i;\vA\mid \vZ_{<i})\tag{Chain rule}\\
&\ge \sum_{i=1}^m\I(\vZ_i;\vA)\tag{as $\I(\vZ_i;\vZ_{<i})=0$}\\
&=\sum_{i=1}^m(\Ent(\vZ_i)-\Ent(\vZ_i\mid \vA))\\
&\ge \sum_{i=1}^m(1-\Ent(1-\delta))\\
&=m\cdot (1-\Ent(\delta))
\end{align*}
This implies that any $(\veps,\delta)$-\LU scheme must have space complexity of at least $m(1-\Ent(\delta))$ bits on datasets of size at most $m+m^\beta/\beta$.

Moving on to the \TiLU scheme, the lower bound is obtained by considering an additional argument. 
Note that the unlearning queries considered are only of points in $\Sc$, which are $m^\beta/\beta$ in number.
In particular, by collecting the tickets of all the points in $\Sc$, the \TiLU scheme should have enough information to recover the entire hidden vector $z \in \{ 0,1 \}^d$. 
Formally, introduce the random variable $\vT$ which is the collection of tickets of all the points in $\Sc$.
\begin{align*}
  |\vA| + \sum_{i \in \Sc  }   | \vT_i | &\ge \I (\vZ; \vA, \vT) \\ 
  & \ge \sum_{i=1}^{m} \I ( \vZ_i ; \vA, \vT \mid \vZ_{<i} ) \\
  & \ge \sum_{i=1}^{m} \I ( \vZ_i ; \vA, \vT ) \\
  & = \sum_{i=1}^{m} ( \Ent(\vZ_i) - \Ent(\vZ_i \mid \vA, \vT) ) \\
  & \ge \sum_{i=1}^{m} ( 1 - \Ent(1-\delta) ) \\
  & = m (1 - \Ent(\delta)).
\end{align*}
Thus, we have that $|\vA| + \sum_{i \in \Sc  }   | \vT_i | \ge m (1 - \Ent(\delta))$ which implies that $\max \{ |\vA|, \max_i |\vT_i| \} \ge\frac{m}{2|\Sc|} = \Omega (\beta m^{1-\beta})$. Rewriting in terms of $n$ $(= m + m^\beta/\beta)$ results in the statement of the theorem.

\end{proof}

\section{Missing Details from \pref{sec:Eluder}}

\subsection{Proof of \pref{thm:lbeluder}} 
\begin{proof}[Proof of \pref{thm:lbeluder}] 
Let $\{ (x_i,y_i) \in \mathcal{X} \times \{ 0,1 \} \}_{i=1}^\eluder$ denote the longest eluder sequence for the hypothesis class $\mathcal{H}$. Namely, for each $i \le \eluder$, there exists hypotheses $h,h' \in \mathcal{H}$ such that $h(x_j) = h'(x_j) = y_j$ for all $j < i$, but $h(x_i) \ne h'(x_i)$.\\

\noindent\textit{Datasets.}~Let $m=\min\{n/2,\eluder\}$. Consider a collection of datasets of size at most $n$, $\mathcal{D} = \{ D_z : z \in \{ 0,1\}^m \}$. Namely, the datasets in $\mathcal{D}$ are indexed by a $m$-length binary string. The dataset $D_z$ is defined with having $z_i$ copies of the point $(x_i,y_i)$ and one copy of the point $(x_i, \neg y_i)$. Note that the size of the datasets in $\mathcal{D}$ is at most $2m\le n$.\\

\noindent\textit{Recovering $z$ using unlearning queries.}~Informally, when the underlying dataset is chosen as $D_z$ for an arbitrary $z \in \{0,1\}^m$, by making unlearning queries, we show that $z$ can be recovered. This means that the \LU scheme must store $\Omega(m)$ bits of information. The unlearning queries we consider peel away $z$ revealing it one coordinate at a time. Unlike the proof of \pref{theorem:LU=>VC} and \ref{theorem:VClb}, our unlearning queries would depend on partial information about $z$. 

As before, let $\vZ$ be a random variable with uniform distribution over $\{0,1\}^m$. Let $\vD$ be a random variable over datasets, taking value $D_z$ when $\vZ=z$. 
Let $(\learn,\unlearn)$ be an $(\veps,\delta)$-\LU scheme for $\mcH$-realizability testing.  Let $\vA$ be a random variable for the auxiliary information outputted by $\learn$ algorithm on dataset $\vD$. For all $i\in[m]$, let $\vU_i$ be a random variable representing an unlearning query, which is a collection of datapoints $(x_j,\neg y_j)_{j<i}$ and all data points on features $\{x_j:j>i\}$. Formally, when $\vZ=z$, $\vU_i=U_{i,z}=\{(x_j,\neg y_j: j\neq i)\}\cup \{(x_j,y_j): j>i \text{ and }z_j=1\}$. Note that $\forall i\in [m], \vU_i\subseteq \vD$. 

Fix $z$ and consider the dataset $D_z$ and the unlearning query $U_{i,z}$. The $i$th unlearning query removes all the points labeled $(x_j,\neg y_j)$ for $j < i$ and all the points with features in $\{x_j:j>i\}$. Note that, if $z_i=1$, the dataset remains $\mathcal{H}$-unrealizable since there are coincident points at $x_i$ labeled $y_i$ and $\neg y_i$. If $z_i=0$, then the dataset $D_z\setminus U_{i,z}$ becomes realizable, since by definition of an eluder sequence, there must be a hypothesis in $\mathcal{H}$ which agrees with all the surviving labels $\{ (x_j,y_j) : j < i \}$ while labeling $x_i$ as either $0$ or $1$, whichever equates to $\neg y_i$. Thus, the $\learn$ algorithm outputs $\yes$ on the dataset $D_z\setminus U_{i,z}$ if and only if $z_i=0$. By \Cref{def:LU}, for all $z,i$, if $D_z\setminus U_{i,z}$ is realizable -- $z_i=0$, then

\[\Pr(\unlearn(U_{i,z}, \aux(D_z))=\no)\le e^{\veps}\cdot 0 +\delta=\delta.\]

Similarly, if $D_z\setminus U_{i,z}$ is not realizable -- $z_i=1$, then

\[\Pr(\unlearn(U_{i,z}, \aux(D_z))=\yes)\le e^{\veps}\cdot 0 +\delta=\delta.\]

Using Bayes' rule, it is easy to see that probability of $z_i=0$ (or $z_i = 1$) conditioned on the output of  $\unlearn(U_{i,z},\aux(D_z))$ being $\yes$ (or $\no$) is greater than $1-\delta$. Thus, for $\delta<1/2$, 
\begin{equation}\label{eq:El1}\Ent\left(\vZ_i\mid\unlearn(\vU_i,\vA)\right)\le \Ent(1-\delta)\implies \I(\vZ_i;\unlearn(\vU_i,\vA))\ge (1-\Ent(1-\delta)).
\end{equation}
We will use this to show that $\vA$ contains $\Omega(m)$ bits of information about $\vZ$. We use $|\vA|$ to denote the size of auxiliary information on datasets in $\mathcal{D}$. Formally, 
\begin{align*}
|\vA|&\ge \I(\vZ;\vA)\tag{as $\Ent(\vA)\le |\vA|$}\\
&=\sum_{i=1}^d\I(\vZ_i;\vA\mid \vZ_{>i})\tag{Chain rule}\\
&=\sum_{i=1}^d\left(\I(\vZ_i;\vA,\unlearn(U_i,\vA)\mid \vZ_{>i})-\I(\vZ_i;\unlearn(\vU_i,\vA)\mid \vA,\vZ_{>i})\right)\\
&\ge \sum_{i=1}^d\left(\I(\vZ_i;\unlearn(\vU_i,\vA)\mid \vZ_{>i})-\I(\vZ_i;\unlearn(\vU_i,\vA)\mid \vA,\vZ_{>i})\right)\\
&= \sum_{i=1}^d\I(\vZ_i;\unlearn(\vU_i,\vA)\mid \vZ_{>i})\tag{Explained below}\\
&\ge \sum_{i=1}^d\I(\vZ_i;\unlearn(\vU_i,\vA))\tag{as $\I(\vZ_i;\vZ_{>i})=0$}\\
&\ge \sum_{i=1}^d(1-\Ent(1-\delta))\tag{Equation \eqref{eq:El1}}\\
&=d\cdot (1-\Ent(\delta))
\end{align*}
 The second last equality follows from the fact that $\I(\vZ_i;\unlearn(\vU_i,\vA)\mid \vA,\vZ_{>i})=0$. This is because the unlearning query $\vU_i$ is deterministic conditioned on $\vZ_{>i}$. Furthermore, the output of the $\unlearn$ algorithm on this query is completely determined by $\vA$ and the random bits used by the $\unlearn$ algorithm, which are independent of $\vZ_{\ge i}$ (and $\vA$). Hence, any $(\veps,\delta)$-\LU scheme must have space complexity of at least $m(1-\Ent(\delta))$ bits on datasets of size  at most $2m$. The theorem follows as $m=\min(n/2,\eluder(\mcH))$.
\end{proof}

\subsection{Proof of \pref{thm:tiluUB}} 
\begin{proof}[Proof of \pref{thm:tiluUB}]We define the hypothesis class and corresponding \TiLU scheme below.\\

\noindent\textit{Hypothesis class.} Consider the domain $\mathcal{X} = [|\mathcal{X}|]$ and the hypothesis class $\mathcal{H}$ which captures every possible labeling on $[d]$ and every $h \in \mathcal{H}$ labels every $x > d$ as $0$. We show that there exists a \TiLU scheme with space complexity upper bounded by $O( \max \{ \log (d), \log (n), \log (|\mathcal{X}|) \})$.
Consider an arbitrary data set on $\mathcal{X}$, defined by the $n_x^0$ points labeled $0$ and the $n_x^1$ points labeled $1$ at each $x \in \mathcal{X}$. Together, let us denote $n_x = (n_x^0, n_x^1)$.\\

\noindent\textit{TiLU Scheme.} The \TiLU scheme we consider is as follows: let $S$ denote the sequence of $x \in [d]$, for which the dataset contains at least one copy of $(x,0)$ and $(x,1)$, and $x > d$ for which the dataset contains at least one copy of $(x,1)$, with all the points arranged in increasing order. The dataset is realizable if and only if for each $x \in S$ such that $x \le d$, either all $n_x^0$ copies of $(x,0)$ or all $n_x^1$ copies of $(x,1)$ are in the unlearn set, and for each $x > d$, all $n_x^1$ copies of $(x,1)$ are in the unlearn set. 

The server stores in $\aux$, $x = S_1$ and $x=S_2$ as well as $n_x$ for both of these points. The ticket of each point $(x,y) \in D$ such that $x = S_i $ stores $S_i$, $n_{S_i}^0$ and $n_{S_i}^1$ as well as $S_{i+1}$, $n_{S_{i+1}}^0$ and $n_{S_{i+1}}^1$ (unless $i = |S|$). The size of the tickets are therefore upper bounded by $O( \max \{ \log (d), \log (n), \log (|\mathcal{X}|) \} )$. The tickets of the remaining points are empty.\\

\noindent\textit{Correctness of the scheme.} Suppose the unlearning query is composed of $k_x^0$ copies of $(x,0)$ and $k_x^1$ copies of $(x,1)$ being deleted for each $x \in \mathcal{X}$. The \TiLU scheme answers,
\begin{enumerate}[label=\(\bullet\)] 
    \item $\no$ if for $x= S_1$ or $x = S_2$, at least one copy of $(x,0)$ and $(x,1)$ survive (which can be computed since the server stores $n_x$ for these points).
    \item In the complementary case, at least one point of the form $(S_2,y)$ must have been in the unlearn set. In this case, the dataset is unrealizable (and the scheme returns $\no$) unless all $n_{x}^1$ copies of $(x,1)$ (or $n_{x}^0$ copies of $(x,0)$, if $x \in [d]$) are in the unlearn set for $x = S_2$ and $x = S_3$ which is computable from the ticket of $(S_2,y)$.
    \item In general we may recurse this argument to consider the complimentary case where some $(S_i,y)$ must have been in the unlearn set. In this case, the dataset is unrealizable (and the scheme returns $\no$) unless all $n_{x}^1$ copies of $(x,1)$ (or $n_{x}^0$ copies of $(x,0)$, if $x \in [d]$) are in the unlearn set for both $x = S_i$ and $x = S_{i+1}$ which can be computed using information stored in the ticket of $(S_i,y)$.
\end{enumerate}
Thus using information stored in the tickets, the \TiLU scheme can determine whether the unlearning query covers the set of all points which must be unlearned to make the dataset realizable.
\end{proof}

\subsection{Proof of \pref{lem:vstomergeable}}
\begin{proof}[Proof of \pref{lem:vstomergeable}]  
Let  \(\oldEnc\) and \(\oldDec\) be the encoding and decoding functions corresponding to the  \(C\)-bit version-space compression scheme for \(\cH\). Recall that a set of hypothesis \(\cH' \subseteq \cH\) is said to be a version space if there exists some dataset \(\bar{\cS} \in \cZ^*\)  such that \(\cH' = \cH(\bar{\cS})\). We next define a function  \(\canonical: 2^{\cH} \mapsto \cZ^*\) that maps subsets \(\cH' \subseteq \cH\) to datasets \(\cS'\) such that (a) if \(\cH'\) is a version space then \(\canonical(\cH')\) returns a canonical dataset \(\cS'\) such that \(\cH' = \cH(\cS')\), and (b) if \(\cH'\) is not a version space then \(\canonical(\cH') = \bot\). Note that for any dataset \(\cS\), we have that \(\cH(\cS) = \cH(\cS')\) for \(\cS'=\canonical(\cH(\cS))\). For the sake of completeness, we provide an implementation of \(\canonical\) function in \Cref{app:canonical}.

We next provide an implementation of the \(\Encode\), \(\Decode\) and \(\Merge\) methods.\\ 

\noindent\textit{$\Encode$:}  
Given an input \(\cS\), check if \(\cS\) is realizable via the class \(\cH\), and  
\begin{enumerate}[label=\(\bullet\)] 
	\item If \(\cS\) is not realizable, let \(\cS' =\canonical(\emptyset)\). 
	\item If \(\cS\) is realizable, then, compute the version space \(\cH' = \crl{h \in \cH \mid h(x) = y~\text{for all}~(x, y) \in \cS}\), and let  \(\cS'=\canonical(\cH')\). 
\end{enumerate}

Return \(\Encode(\cS) = \oldEnc(\cS')\). \\

\noindent\textit{\(\Decode\):} Given an encoding \(E\)  compute the version space \(\cH' \leftarrow \oldDec(E)\), and return \(\no\) if \(\cH' = \emptyset\), and \(\yes\) otherwise. \\

\noindent\textit{\Merge:} Given encodings \(E_1\) and \(E_2\), compute the version spaces  \(\cH_1 \leftarrow  \oldDec(E_1)\) and \(\cH_2 \leftarrow \oldDec(E_2)\), and define \(\cH'' = \cH_1 \cap \cH_2\). Compute the dataset \(\cS'' = \canonical(\cH'')\) and return \(\Merge(E_1, E_2) = \oldEnc(\cS'')\). \\

\noindent\textit{Showing that \(\cH\) is a mergeable hypothesis class.} We next prove that the above defined procedures satisfy \Cref{def:mergeable_testing}. Let \(E\) be the output of \(\Encode\) for a dataset \(\cS\). By definition of the \(\canonical\) function, we have that \(\oldDec(E) = \cH(\cS') = \cH' = \cH(\cS)\). Thus, \(\Decode(\Encode(\cS)) = \) \(\yes\)~iff \(\cS\) is realizable. 

We next establish the mergeability property. Note that for any sets \(\cS_1\) and \(\cS_2\), the output of \(\Encode(\cS_1 \cup \cS_2) = \oldEnc(\cS')\)  where \(\cS' = \canonical(\cH(\cS_1 \cup \cS_2))\). On the other hand, we compute \(\Merge(\Encode(\cS_1), \Encode(\cS_2)) = \oldEnc(\cS'')\)   where \(\cS'' = \canonical(\cH(\cS_1) \cap \cH(\cS_2))\). However, since \(\cH(\cS_1) \cap \cH(\cS_2) = \cH(\cS_1 \cup \cS_2)\) by definition of the version spaces, we have that \(\cS' = \cS''\) and thus \(\Merge(\Encode(\cS_1), \Encode(\cS_2)) = \Encode(\cS_1 \cup \cS_2)\). 
Since all generated encodings are at most \(C\)-bits, we immediately get that \(\cH\) is \(C\)-bit mergeable. 
\end{proof}

\subsection{Proof of \pref{thm:eludercompression}} \label{app:eludercompression_proof}

\begin{proof}[Proof of \pref{thm:eludercompression}] We will build up to the full version of the result in parts. First we will prove the following claim, which is a weaker version of \pref{thm:eludercompression} which replaces $\starno (\cH)$ by the smaller $\Eld(\cH)$. Then, we will argue the stronger result by extending the proof.

\medskip
\textbf{Claim 1.} Any hypothesis class $\mathcal{H}$ has an $O(\Eld(\cH) \log(\abs{\cZ}))$-bit version-space compression scheme, where $\Eld(\cH)$ denotes the Eluder dimension of $\cH$.

\medskip
Toward the goal of proving this claim, we first define the \(\enc\) and \(\dec\) methods, and then show that they satisfy \Cref{def:vs_compression} with \(C = O(\Eld(\cH) \log(\abs{\cZ}))\). \\

\noindent\textit{\enc:}  Given a set of samples $\cS$, the encoding function creates an eluder subsequence $\cS'$ such that $\mcH(\cS')=\mcH(\cS)$. Initialize \(\cS' = \emptyset\) . Iterate over samples \((x, y) \in \cS\): 
\begin{itemize}
	\item Let \(\cH(\cS')\) be the set of all the hypothesis that are consistent with the samples in the set \(\cS'\) so far. 
	\item Update \(\cS' = \cS'\cup\crl{(x, y)}\) if there exists two hypothesis \(h_1, h_2 \in \cH(\cS')\) such that \(h_1(x) \neq h_2(x)\). 
 \item Otherwise, we are in the case that for all $h\in\cH(\cS')$, $h(x)=b$ for some $b\in\{0,1\}$. If $b=y$, then go to the next sample. Otherwise, add $(x,y)$ to $\cS'$ and exit the iteration. Note that in this case, both $\cS$ and $\cS'$ are not realizable.  
\end{itemize}
Return \(\cS'\) as the output of \(\Encode(\cS)\). \\

\noindent\textit{\dec:} Given the set of samples \(\cS'\) as the  encoding, return the version space \(\cH(\cS')\). \\

\noindent\textit{Proof of version space compression.}
We next show that for any set \(\cS\), \(\dec(\enc(\cS)) = \cH(\cS)\). For this, we prove that \(\cH(\cS) = \cH(\cS')\), using induction on number of samples that have been iterated over in the encoding protocol $\enc$. We prove that for all iterations indexed by $i$, \(\cH(\cS_i') =\cH(\cS_i)\). Here, $\cS'_i $ denotes the set $\cS'$ after $i$ iterations and $\cS_i$ denotes the set of first $i$ samples in $\cS$. The statement is trivially true for $i=0$.
To prove the induction step, we assume that  \(\cH(\cS_i') = \cH(\cS_i)\) after $i$ iterations, and show that the statement is true for the $(i+1)$th iteration. The forward direction is straightforward, as by construction, \(\cS'_{i+1} \subseteq \cS_{i+1}\) and thus \(\cH(\cS_{i+1}) \subseteq \cH(\cS'_{i+1})\). We next prove the converse direction, that is \(\cH(\cS_{i+1}') \subseteq \cH(\cS_{i+1})\).

Suppose there exists a hypothesis \(h \in \cH(\cS'_{i+1}) \setminus \cH(\cS_{i+1})\). Since  \(h \notin \cH(\cS_{i+1})\) but $h\in\cH(\cS_{i})=\cH(\cS_{i}')$, for the sample in the $(i+1)$th iteration (say \((x_{i+1}, y_{i+1})\)), \(h( x_{i+1}) \neq y_{i+1}\). This implies that $(x_{i+1}, {y}_{i+1})\notin \cS_{i+1}'$. Under the $\enc$ protocol, we don't $(x_{i+1},y_{i+1})$ to $\cS'$ only when $\forall h'\in \cH(\cS'_i)$, $h'(x_{i+1})=y_{i+1}$. This implies that $h\notin  \cH(\cS'_i)\supseteq\cH(\cS'_{i+1})$, which is a contradiction. Hence, \(\cH(\cS_{i+1}') \subseteq \cH(\cS_{i+1})\).\\

\noindent\textit{Bound on \(|\enc(\cS)|\).} We next argue that for any input set \(\cS \in \cZ^\star\), the encoding \(\cS' = \enc(\cS)\) contains at most \((\Eld(\cH)+1)\) samples. Let \(m\) denote the number of samples in \(\cS'\). Note that for all but last sample added to $\cS'$, the first $m-1$ samples in $\cS'$ are realizable via \(\cH\). Thus, there exists a function \(h'\) such that \(h'(x_{i_j})  = y_{i_j}\)  for all samples \((x_{i_j}, y_{i_j}) \in \cS'\) where \(j \leq m-1\). Next, note that, by construction of the set \(\cS'\) in the \(\enc\) protocol, the sample \((x_{i_j}, y_{i_j})\) is added to \(\cS'\) iff there exist two hypothesis \(h_1,h_2 \in \cH(\cS'_{i_j-1})\subseteq\cH(\cS'_{i_{j-1}})\) such that \(h_1(x_{i_j}) = y_{i_j}\) and \(h_2(x_{i_j}) \neq y_{i_j}\).  Thus, the first $m-1$ samples in \(\cS'\) form an eluding sequence w.r.t.~the hypothesis class \(\cH\). Thus, we must have that \(m-1 \leq \Eld(\cH)\). This implies that the number of compression bits \(C\) used by $\enc$ satisfy \(C \leq O\prn*{\Eld(\cH) \log(\abs{\cZ})}\). 

\medskip
Having proved Claim 1, we will establish how to improve this result to only depend on the star number of $\cH$, $\starno (\cH)$.
\medskip

\noindent\textit{Proof of \pref{thm:eludercompression}.} By Claim 1, we know that the version-space compression of $\cH$ is of size at most $O(\Eld(\cH) \log (\abs{\cZ}))$ bits. Furthermore, by construction of $\enc(\cdot)$, we know that this is implied by the compression of a dataset $\cS$ which is simply its corresponding eluder subsequence $\cS'$. In particular, we know that $\cH (\cS) = \cH (\cS')$ where $\cH(\cdot)$ computes the version space of (i.e., hypotheses consistent with) its argument. We will show how to prune $\cS'$ in a way which preserves this property. Let's order the eluder set $\cS' = (z_1,\cdots,z_{\Eld{(\cH)}})$ in the canonical order induced by the eluder dimension and iteratively prune the dataset as follows,
\begin{enumerate}
    \item Instantiate $\cS''_1 \gets \cS'$.
    \item Iterating over $i \in [\Eld(\cH)]$ in order, if $\cH (\cS''_i \setminus \{ z_i \}) = \cH (\cS''_i)$, then update $\cS''_{i+1} \gets \cS''_i \setminus \{ z_i \}$. That is, in a sequential manner, prune all those datapoints which do not increase the version space of the dataset $\cS'$.
\end{enumerate}
By construction, at the end of this process we are left with a subset $\cS_{\text{final}}'' = \cS''_{\Eld(\cH)}$ which has the same version space as that of $\cS'$ (and $\cS$). What remains to be shown is that the resulting dataset is of size at most $\starno(\cH)$. If in some iteration, for the datapoint $z_i = (x_i,y_i)$, if $\cH (\cS''_i \setminus \{ z_i \}) \ne \cH (\cS''_i)$, this implies that there exist two hypotheses $h_i,h'_i \in \cH (\cS''_i \setminus \{ z_i \})$ such that $h_i(x_i) \ne h'_i(x_i)$. That is, $h_i$ and $h'_i$ agree on all ``future'' points after $x_i$, but not $x_i$ itself. On the other hand, by the property of the eluding sequence, $h_i$ and $h'_i$ must agree on all ``prior'' points of $\cS'$ before $z_i$ which happened to survive in $\cS''_i$. Namely, if $h_i(x_{i'}) \ne h'_i(x_{i'})$ for some $i' < i$ such that $z_{i'} \in \cS''$, then $h$ and $h'$ cannot belong to $\cH (\cS''_i \setminus \{ z_i \})$ because they disagree at the point $z_{i'} \in \cS''_i \setminus \{ z_i \}$. Thus $h_i$ and $h'_i$ realize the dataset $\cS''_{\text{final}} \setminus \{ (x_i,y_i) \}$ and satisfy $h_i (x_i) \ne h_i'(x_i)$.

\smallskip
Overall, this implies that each surviving point $z = (x,y) \in \cS_{\text{final}}''$ the datasets $ \cS_{\text{final}}''$ and $\cS_{\text{final}}'' \setminus \{ (x,y) \} \cup \{ (x,\bar{y}) \}$ are both realizable. This implies that $|\cS_{\text{final}}''| \le \starno(\cH)$, proving the result.
\end{proof}

\subsection{Proof of \pref{lem:mergeable_compression}} 
\begin{proof}[Proof of \pref{lem:mergeable_compression}]
Let a set of functions $\Encode,\Merge$ and $\Decode$ make a $C$-bit mergeable compression scheme for $\mcH$. We show that there exists a function $\dec:\{0,1\}^C\rightarrow 2^{\mcH}$ such that the pair of functions $(\Encode,\dec)$ is a $C$-bit version-space compression scheme for $\mcH$. Give a $C$-bit compression string $E$, $\dec$ outputs a subset $\mcG\subseteq\mcH$:
\begin{enumerate}
\item Initialize $\mcG=\emptyset$.
\item For all $h\in \mcH$, execute the following steps
    \begin{enumerate}
        \item \label{itm:mtovs1} Let $\mcS_h=\{(x,y)\mid x\in\mcX\text{ and }y=h(x)\}$. 
\item Let $F=\Encode(\mcS_h)$.
\item \label{itm:mtovs2} If $\Decode(\Merge(E,F))=\yes$, then add $h$ to $\mcG$, otherwise go to the next hypothesis.
    \end{enumerate}
\item Output $\mcG$.
\end{enumerate}
To prove that $(\Encode,\dec)$ is a version-space compression scheme, we want to show that for all sample sets $\mcS\in(\mcX\times\mcY)^*$, $\dec(\Encode(\mcS))=\mcH(\mcS)$. Let $E=\Encode(\mcS)$ and $\mcG$ be the subset outputted when the above $\dec$ algorithm is run on $E$. Below, we prove that $$h\in \mcG\text{ if and only if }h\in \mcH(\mcS).$$ Recall that $\mcH(\mcS)=\{h\in\mcH\mid  h(x)=y \text{~for all~} (x,y)\in \mcS \}.$\\

\noindent\textit{Case 1 ($h\in \mcH({\mcS})$).} Consider the set $\mcS\cup \mcS_h$ (where $\mcS_h$ is defined in Step \ref{itm:mtovs1} of the $\dec$ algorithm). As $h\in \mcH(\mcS$), all samples in $\mcS$ agrees with $h$ and hence, the set $\mcS\cup \mcS_h$ is realizable. By the definition of mergeable compression $\Merge(\Encode(\mcS),\Encode(\mcS_h))=\Encode(\mcS\cup\mcS_h)$. As $\mcS\cup\mcS_h$ is realizable, $\Decode(\Merge(E,\Encode(\mcS_h)))=\yes$ and thus, $h$ is added to $\mcG$.\\

\noindent\textit{Case 2 ($h\notin \mcH({\mcS})$).} Similar to Case 1, consider the set $\mcS\cup \mcS_h$. We will prove that $\mcS\cup \mcS_h$ is not realizable. This would further imply that $\Decode(\Merge(E,\Encode(S_h)))=\Decode(\Encode(\mcS\cup\mcS_h))=\no$ and hence $h$ is not added to $\mcG$ at Step \ref{itm:mtovs2} of the $\dec$ algorithm. As $h\notin \mcH({\mcS})$, there exists a sample $(x,y)\in \mcS$ such that $y\neq h(x)$. As $\mcS_h$ contains the sample $(x,h(x))$, both $(x,y)$ and $(x,h(x))$ belongs to the set $\mcS\cup\mcS_h$ for a $y\neq h(x)$, implying that it is not realizable.  
\end{proof} 

\subsection{Proof of  \pref{thm:mergeablealg}}  \label{app:mergeablealg}  

The following proof is a straightforward extension of the proof of \cite[Theorem 5]{ghazi2023ticketed} and is included here for completeness. 

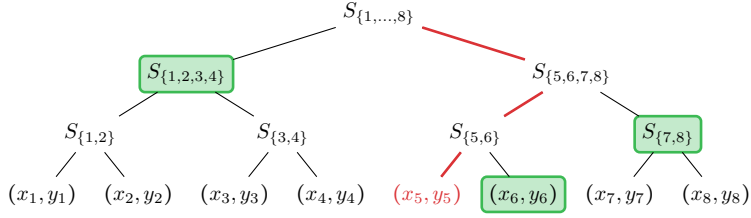
\begin{figure} 
\centering \small
\begin{tikzpicture}[
  grow=down,
  level 1/.style={sibling distance=64mm},
  level 2/.style={sibling distance=32mm},
  level 3/.style={sibling distance=16mm},
  level 4/.style={sibling distance=8mm},
  level distance=10mm,
  highlight/.style={rectangle,rounded corners=2pt,draw=Ggreen,fill=Ggreen!30,line width=1pt},
  every node/.style = {outer sep=0pt},
  scale=0.8, transform shape
]
\node (1) {$S_{\crl{1,\ldots,8}}$}
  child {
    node[highlight] {$S_{\crl{1,2,3,4}}$}
    child {
      node {$S_{\crl{1,2}}$}
      child {
        node {$(x_1, y_1)$}
      }
      child {
        node {$(x_2, y_2)$}
      }
    }
    child {
      node {$S_{\crl{3,4}}$}
      child {
        node {$(x_3, y_3)$}
      }
      child {
        node {$(x_4, y_4)$}
      }
    }
  }
  child {
    node (3) {$S_{\crl{5,6,7,8}}$}
    child {
      node (6) {$S_{\crl{5,6}}$}
      child {
        node[Gred] (12) {$(x_5, y_5)$}
      }
      child {
        node[highlight] {$(x_6, y_6)$}
      }
    }
    child {
      node[highlight] {$S_{\crl{7,8}}$}
      child {
        node {$(x_7, y_7)$}
      }
      child {
        node {$(x_8, y_8)$}
      }
    }
  };
\path[Gred, line width=1pt]
(1) edge (3)
(3) edge (6) 
(6) edge (12);
\end{tikzpicture}
\caption{Picture taken from \cite{ghazi2023ticketed}. Illustration of TiLU scheme underlying the proof of Theorem \ref{thm:mergeablealg}. The ticket $t_5$ for the example $(x_5, y_5)$ are the index $i$ and the outputs of $\Encode$ applied on the subsets $S_{\crl{1,2,3,4}}$, $S_{\crl{7,8}}$, and $S_{\crl{6}}$.}  
\label{fig:merkle}
\end{figure}

\begin{proof} 
For simplicity, assume that $n$ is a power of $2$; the argument generalizes to all $n$ easily. Let the samples in \(S\) be \(\crl{(x_i, y_i)}_{i \leq n}\) (arranged in any  order), and consider a full binary tree of depth $d = \log_2 n$ with leaf $i$ corresponding to the sample $(x_i, y_i)$. 
For each internal node $v$, let $S_v \ldef{} \crl{(x_i,y_i) \mid v~\text{is an ancestor of the leaf}~i}$ be the dataset consisting of examples corresponding to the leaf nodes in the subtree under $v$.
For any leaf $i$, let $v_1, \dots, v_{d - 1}$ be the nodes on the path from the root of the tree to leaf $i$, and for each $j \in \crl{2,\ldots, d}$ let $\tilde{v}_j$ be the child of $v_{j-1}$ and sibling of $v_j$. \Cref{fig:merkle} shows an example.
Let the ticket corresponding to example $i$ be given as
\[
\ticket_i \ldef{} \prn{i, \Encode(S_{\tilde{v}_2}), \dots, \Encode(S_{\tilde{v}_{d}})}.
\]

It is immediate to see that the number of bits in $\ticket_i$ is $d + C\cdot (d-1)$, where \(d = \log(n)\). Define $\learn(S)$ to return $\aux =  \Decode(\Encode(S))$ and tickets $\ticket_i$ as specified above. 

We define $\unlearn$ as follows. If $I = \emptyset$, then simply return $\aux$. Given a non-empty $I\subseteq [n]$, let $R$ be the set of all nodes $v$ such that no leaf in the sub-tree under $v$ belongs to $I$, but the same is not true of the parent of $v$. It is easy to see that $S \smallsetminus S_I$ is precisely given as $\bigcup_{v \in R} S_v$, and moreover, $S_{v}$ and $S_{v'}$ are disjoint for distinct $v, v' \in R$. 
For all $v \in R$, we can recover $\Encode(S_v)$ from ticket $\ticket_i$ for any leaf $i \in I$ in the subtree under the sibling of $v$. Thus, by repeated applications  of $\Merge$, we can recover $\Encode(\bigcup_{v \in R} S_v) = \Encode(S \smallsetminus S_I)$. Thus, we can compute the  \(\crl{\yes, \no}\) answer after unlearning using $\Decode(\Encode(\bigcup_{v \in R} S_v)))$. 
\end{proof} 

\subsection{Missing Details from  \pref{sec:compression}}  \label{app:canonical}  

In the following, we provide an implementation of the \(\canonical\) function, used in the proof of Lemma \ref{lem:vstomergeable}. 

\vspace{3mm} 

\noindent 
\canonical: 
Given a subset \(\cH' \subseteq \cH\) as the input, if $\cH'=\emptyset$, return $\{(x,0),(x,1)\}$ for a fixed $x\in \cX$, otherwise initialize \(S' = \emptyset\), and grow it as follows:  For \(x \in \sorted(\cX)\): 
\begin{itemize}[label=\(\bullet\), leftmargin=16mm] 
\item Update \(S' = S' \cup \crl{(x, y)}\) if \(h'(x) = y\) for all \(h' \in \cH'\) but there exists an \(h \in \cH\) such that \(h(x) \neq y\). Else, continue to the next sample. 
\end{itemize} 

Check if \(\cH(S') = \cH'\). If yes return \(S'\), otherwise, return \(\bot\). 

\vspace{3mm} 

Here, we prove a bound on the size of a version space compression scheme for parities which is relevant to the results of \Cref{rem:parity}.

\begin{lemma}
    \label{lem:bd_version_compression_parities}
    Any $C$-bit version-space compression scheme for the class of parities $\Hparity$ in dimension $d$ must satisfy:
    \begin{equation*}
        C \geq \frac{d(d - 6)}{2}.
    \end{equation*}
    Furthermore, this is also true when restricted to datasets of size at most $d / 2$.
\end{lemma}
\begin{proof}
    Consider a dataset generated through the following process:
    \begin{itemize}
        \item First choose a subspace $V$ of $\{0, 1\}^d$
        \item Choose a set of $d / 2$ linearly independent vectors, $\{x_i\}_{i = 1}^{d / 2}$ in $V$
        \item Generate dataset $S = \{(x_i, 1)\}_{i = 1}^{d / 2}$.
    \end{itemize}
    Let $w^* \in \{0, 1\}^d$ be any solution to the dataset. Then, observe that:
    \begin{equation*}
        V_\perp = \{w - w^*: w \in \mc{H} (S)\}
    \end{equation*}
    where $V_\perp$ denotes the orthogonal complement of the sub-space $V$ which is also a subspace of dimension $d / 2$. Hence, $C + d$ bits corresponding to $\Encode (S)$ with a canonical solution, $w^*$ with the decoding function, $\Decode$, suffice to recover $S$. However, the choice of $S$ is arbitrary and there exist \cite{qbinom}:
    \begin{equation*}
        \binom{d}{d / 2}_2 \coloneqq \frac{(1 - 2^d)(1 - 2^{d - 1}) \dots (1 - 2^{d / 2 + 1})}{(1 - 2) \dots (1 - 2^{d / 2})}
    \end{equation*} 
    possible choices for $S$. From the following observation,
    \begin{equation*}
        \frac{1 - q^{t}}{1 - q^{t - d / 2}} \geq 2^{d / 2 - 1},
    \end{equation*}
    we get that:
    \begin{equation*}
        \binom{d}{d / 2}_2 \geq 2^{d(d - 2) / 4}. 
    \end{equation*}
    Hence, we must have:
    \begin{equation*}
        d + C \geq \frac{d (d - 2)}{4} \implies C \geq \frac{d(d - 6)}{2}
    \end{equation*}
    concluding the proof.
\end{proof}

 \subsection{Version-space Compression is Necessary for Unlearning in the Central Model} \label{app:version_space_necessary} 
 
 \pref{thm:vscentral}  shows that version-space compression is necessary for unlearning in the central model. Before we state and prove the theorem, we define a minimum identification set for a hypothesis class.

\begin{definition}
[Minimum Identification Set]\label{def:mis} 
For any hypothesis class $\mcH$, minimum identification set $\mcI\subseteq \mcX$ is the minimum sized subset such that knowing the labels on this set, identifies the hypothesis $h$ that generated them. Formally,
\[\mcI=\argmin {|\cS|} \;\;s.t.\;\; {\forall h_1,h_2\in\mcH, (h_1=h_2)\text{ or } (\exists x\in \cS, h_1(x)\neq h_2(x))}.\]
Let $\Idd(\mcH) $ denote the size of the minimum identification set.  
\end{definition}

The above definition of minimum identification set is closely related to the notion of minimal specifying set in active learning \citep{hanneke2009theoretical, hanneke2024eluder}, with the main difference being that we do not restrict ourselves to a particular labeling function on the set \(\cT\). 

Note that the entire set $\mcX$ is an identification set, but for certain hypothesis class, the size of the minimum identification set can be much smaller. One such example is the class of parity functions defined on $n$-bit strings, that is, $\mcX=\{0,1\}^n$. A parity function is represented by another $n$-bit string $a$ and computes the parity of the inner product of $a$ with input $x$, that is, $f_a(x)=\left(\sum_{i=1}^na_ix_i\right)\mod 2$. Thus, the class of parity functions contains all such functions --  $\Hparity=\{f_a\mid a\in\{0,1\}^n\}$. Knowing the output of a parity function on indicator strings (that is, $(1,0,\ldots,0), (0,1,0,\ldots,0),\ldots$) determines the function and hence, this is an indenfication set. Therefore, for the class of parity functions, $\Idd(\Hparity)\le n \ll 2^n=|\cX|$.

In the following theorem, we show that for classes with bounded eluder dimension and small sized identification sets, $\LU$ schemes require large storage even for small sized datasets. \pref{thm:lbeluder} beats the bound achieved by following theorem when $\Idd(\mcH)\approx |\cX|$.

\begin{theorem} 
\label{thm:vscentral}
For any hypothesis class $\mcH$, if there exists a $(0,0)-\LU$ scheme (under the central model) with space complexity $\s (n)$, then there exists a version-space compression scheme for $\mcH$, which uses $\s(\Eld(\mcH)+2\Idd(\mcH)+1)$ bits of compression.
Here, $\Eld(\mcH)$ and $\Idd(\mcH)$ represent the eluder dimension and the size of minimum identification set for $\mcH$, respectively\footnote{Note that, both these quantities are incomparable to each other. We give examples in \pref{prop:mssvseluder}.}.
\end{theorem}

\begin{remark}\label{rem:parity}
As a concrete application, for class of parity functions, where $\Eld(\Hparity)=\Idd(\Hparity)=n$, $(0,0)-\LU$ schemes requires $\Omega(d^2)$ bits of storage even for datasets of sample size $\Theta(d)$. This is because, version-space compression for the class of parity functions requires $\Omega(d^2)$ bits. We give proof of the last fact in \Cref{lem:bd_version_compression_parities}.
\end{remark}

\begin{proof}[Proof of \pref{thm:vscentral}]
Let $(\learn,\unlearn)$ be an $\LU$ scheme for the hypothesis class $\mcH$ with space complexity $s(n)$. We construct the pair of functions $(\enc,\dec)$ for version-space compression for the class $\mcH$ as follows:\\

\noindent\textit{$\enc$:} Given a set of samples $\mcS$, the encoding algorithm constructs an eluder subsequence $\mc\cS'$ as in the proof of \pref{thm:eludercompression}, which is a subset of samples in $\mcS$, such that $|\cS'|\le \Eld(\mcH)+1$ and $\mcH(\mcS)=\mcH({\mc\cS'})$. Let $\mcI$ be the minimum identification set for the hypothesis class $\mcH$. Let $D$ be a dataset that contains $\cS'$ plus two samples per $x\in \mcI$, that $(x,0)$ and $(x,1)$. Thus, $D$ contains at most $\Eld(\mcH)+1+2\Idd(\cH)$ samples. $\enc$ outputs the auxiliary information that $\LU$ scheme uses for this dataset -- $\learn(D)$. Using the space complexity of the $\LU$ scheme, we can bound the compression bits used by the encoding algorithm by $\s(|D|)$.\\

\noindent\textit{$\dec$:} Given the auxiliary information as an encoding $E$, the decoding algorithm constructs a version-space $\cH'$ as follows:
\begin{itemize}
\item Initially $\cH'=\emptyset$.
\item For all $h\in\cH$, we construct an unlearning request:
\begin{itemize}
\item Let $\cS_h=\{(x,1-h(x))\mid x\in\mcI\}$.
\item If $\unlearn(\cS_h,E)$ answers $\yes$, then add $h$ to $\cH'$.
\end{itemize}
\item Output $\cH'$.
\end{itemize}

To prove the $(\enc,\dec)$ is a version-space compression for $\cH$, we need to show that for all sample sets $\cS$, $\dec(\enc(\cS))=\cH(\cS)$. It suffices to show that $\dec(\enc(\cS))=\cH(\cS')$ where $\cS'$ is the eluder subsequence of $\cS$.
Firstly, we note that the unlearning request for each hypothesis is valid because the dataset $D$ used for encoding contains both $(x,0)$ and $(x,1)$ for all $x\in\mcI$. After deletion, the updated dataset $D_h=D\setminus \cS_h=\cS'\cup \{(x,h(x)\mid x\in\mcI\}$. By the learning-unlearning guarantee of the pair of functions $(\learn,\unlearn)$, $\unlearn(\cS_h,E)$ is $\yes$ if and only if $D_h$ is realizable. So, it suffices to show that $D_h$ is realizable if and only if $h\in\cH(\cS')$.

The backward direction is straightforward -- if $h\in \cH(\cS')$, then $h$ realizes all the samples in $D_h$, and hence the dataset is realizable. 
To prove the other direction, we use the fact that $\mcI$ is an identification set. If $D_h$ is realizable, then there exists $h'\in\cH$ such that $\forall (x,y)\in D_h, \;h'(x)=y$. As $\forall x\in\mcI,\; (x,h(x))\in D_h$, this implies that $h(x)=h'(x),\forall x\in\mcI$. By \pref{def:mis}, as $\mcI$ is an identification set, $h=h'$. As $D_h$ contains all samples in $\cS'$, $h$ realizes these samples too. Hence, $h\in \mcH(\cS')$.
\end{proof}

\section{Missing Details from \pref{sec:bdd_deletions}} 

\begin{proof}[Proof of \pref{lem:bdd_unrealizable_set}]
    The proof of the lemma will follow a greedy construction. We start by defining $S_0 \coloneqq S$ and construct $S_i$ for $i > 0$ as follows:
    \begin{enumerate}
        \item If there exists $(x, y) \in S_{i - 1}$ such that $S_{i - 1} \setminus \{(x, y)\}$ is unrealizable, construct $S_i \coloneqq S_{i - 1} \setminus \{(x, y)\}$.
        \item Otherwise, stop and let $S' \coloneqq S_{i - 1}$
    \end{enumerate}
    First, observe that the process must output an unrealizable set by definition as the process starts with an unrealizable set at step $0$, and $S_{i}$ is unrealizable if $S_{i - 1}$ is unrealizable. Now, to bound the size of $S'$, note that, by the termination condition, the removal of any point from $S'$ renders it realizable. Hence, $S'$ is a hollow star set and its size is bounded above by $\hstarno$.
\end{proof}

\begin{proof}[Proof of \pref{thm:bdd_deletions_star_no}] 
    To prove the theorem, we start by bounding the number of $l$-critical sets for $l \in [k]$. First, fix $l \in [k]$. We will maintain a list of prefixes of $l$-critical sets; that is, these are subsets which may be expanded to construct $l$-critical sets. To start, let $C_0 = \{()\}$ be the set of $0$-length prefixes of $l$-critical sets. For $0 < i \leq l$, we construct $C_i$ from $C_{i - 1}$ as follows:
    \begin{itemize}
        \item Let $P = ((x_1, y_1), \dots, (x_{i - 1}, y_{i - 1})) \in C_{i - 1}$ be a prefix of an $l$-critical set.
        \item Since, $P$ is a prefix of an $l$-critical set, $S \setminus P$ is unrealizable. 
        \item From \Cref{lem:bdd_unrealizable_set}, there exists a subset $S' \subset S \setminus P$ which is also unrealizable and has $\abs{S'} \leq \hstarno$
        \item Any completion of $P$ to a $l$-critical set must feature one of the elements of $S'$
        \item For all $(x, y) \in S'$ such that $P \bigcup \{(x, y)\}$ is extendible to an $l$-critical set, add $P \cup \{(x, y)\}$ to $C_i$
    \end{itemize}
    It is clear from the above construction that $\abs{C_i} \leq \hstarno \abs{C_{i - 1}}$. Furthermore, note that any $l$-critical set $((x_1, y_1), \dots, (x_l, y_l))$ is a member of $C_l$ by applying induction on the observation that if $((x_1, y_1), \dots, (x_i, y_i)) \in C_i$, then $((x_1, y_1), \dots, (x_{i + 1}, y_{i + 1})) \in C_{i + 1}$. Therefore, we obtain:
    \begin{equation*}
        \abs{C_l} \leq \hstarno^l. 
    \end{equation*}
    To conclude the proof of the theorem, observe that for any dataset $S = \{(x_i, y_i)\}_{i = 1}^n$, it suffices to store the set of critical sets of sizes up to $k$. The answer to any query may be produced simply by checking whether the points in the query contain a critical set. This produces a valid answer to the unlearning query as any set of deletions that render a dataset realizable must contain a critical set of size at most $k$ by noting that points may be removed from the deletion query until it reduces to a critical set. The number of critical sets of size less than $k$ is now given by:
    \begin{equation*}
        \sum_{i = 1}^k \hstarno^{i} \leq \hstarno^{k + 1}.
    \end{equation*}
    This concludes the proof of the theorem.
\end{proof}

\section{Missing Details from \pref{sec:halfspaces}}  \label{app:halfspace_proof} 
\subsection{Proof of \pref{lem:critic_set_size_linear}} 
\begin{proof}[Proof of \pref{lem:critic_set_size_linear}] 
    Let $S = \{(x_1, y_1), \dots, (x_l, y_l)\}$ be a hollow star set and $S = S_p \cup S_n$ be a decomposition of $S$ into its positively and negatively labeled points. Since $S$ is not realizable by any linear classifier, there exists a point $z \in \conv(S_p) \cap \conv (S_n)$. For this point $z$. Suppose $z = \sum_{x \in S_p} w_x x = \sum_{y \in S_n} w'_y y$ such that $w_x \geq 0$ for all $x \in S$ and $\sum_{x \in S_p} w_x = \sum_{y \in S_n} w_y = 1$. Let $Q = Q_p \cup Q_n$ with $Q_p$ and $Q_n$ denoting the supports of $w$ and $w'$ respectively; i.e $Q_p = \supp (w)$ and $Q_n = \supp (w')$. We will show that there exists a $Q \subset S$ such that $\abs{Q} \leq d + 2$. 
    
    \begin{claim}
        \label{clm:q_supp_linear_class}
        We have:
        \begin{equation*}
            \abs{Q} \leq d + 2.
        \end{equation*}
    \end{claim}
    \begin{proof}
        Suppose for the sake of contradiction that $\abs{Q} > d + 2$. Then, we have that there exists $\wt{w}, \wt{w}' \neq 0$ such that the following hold:
        \begin{gather*}
            \sum_{x \in Q_p} \wt{w}_x x = \sum_{y \in Q_n} \wt{w}'_y y \\
            \sum_{x \in Q_p} \wt{w}_x = 0 \text{ and } \sum_{y \in Q_n} \wt{w}_y = 0. 
        \end{gather*}
        Consider the vector, $(w_\gamma, w'_\gamma) = (w + \gamma \wt{w}, w' + \gamma \wt{w}')$ and the largest $\gamma$ such that all the entries of $w_\gamma$ and $w'_\gamma$ are non-negative and the corresponding $\gamma^*$. Observe that one of the $w_{\gamma^*}, w'_{\gamma^*}$ must have a zero-coordinate. Furthermore, the vectors obtained using the weights $w_{\gamma^*}, w'_{\gamma^*}$ also lie in the intersection of the convex hulls of $S_p$ and $S_n$. The size of the support of $w_{\gamma^*}, w'_{\gamma^*}$ is $\abs{Q} - 1$. Iterating this process, we arrive at a support set of size $d + 2$. 
    \end{proof}
    Since $S$ is a hollow star set, $S \setminus \{(x, y)\}$ is realizable for any $(x, y) \in S$. However, any $(x, y) \in S$ must also belong to $Q$ as $Q$ itself is unrealizable. \Cref{clm:q_supp_linear_class} concludes the proof.
\end{proof}

\subsection{Proof of \pref{thm:linear_central_lb}}  

\begin{proof}[Proof of \pref{thm:linear_central_lb}]
    We start by constructing a family of datasets and show that the central \LU scheme \emph{must} use a large amount of memory for at least one member of the family. Our family of datasets will be indexed by subsets of the set $\mc{I}$ which consists of all size-$k$ tuples of elements from $[d]$; i.e.: 
    \begin{equation*}
        \mc{I} \coloneqq \lbrb{\lbrb{i_1, \dots, i_k}: \forall j \in [k], i_j \in [d], \forall l \neq j: i_j \neq i_l}.
    \end{equation*}
    Now, for any subset $S \subseteq \mc{I}$, our dataset, denoted by $D_S$, is defined as follows:
    \begin{enumerate}
        \item Add $(e_i, 1)$ to $D_S$ for all $i \in [d]$
        \item For all $L = \{i_1, \dots, i_k\} \in S$, add:
        \begin{equation*}
            \lprp{x = \frac{1}{d - k} \sum_{j \notin L} e_{j}, y = 0}
        \end{equation*}
        to the dataset. That is, $x$ corresponds to the mean of the standard basis vectors corresponding to the elements not in $L$, and the corresponding label is $0$. 
    \end{enumerate}
    Hence, the dataset is of size $\abs{S} + d$. We now establish the following structural property on the critical sets of $D_S$ that enables identifying $S$ from the set of critical sets. Denoting by $Q_S$ the set of size-$k$ critical sets of $D_S$, we show the following.
    \begin{claim}
        \label{claim:critic_set_ident}
        We have:
        \begin{gather*}
            \forall \{i_1, \dots, i_k\} \in S: \{(e_{i_1}, 1), \dots, (e_{i_k}, 1)\} \notin Q_S \\
            \forall \{i_1, \dots, i_k\} \notin S: \{(e_{i_1}, 1), \dots, (e_{i_k}, 1)\} \in Q_S 
        \end{gather*}
    \end{claim}
    \begin{proof}
        Let $L = \{i_1, \dots, i_k\}$. We will prove each of the sub-claims in turn. For the first, assume the contrary that there exists a linear classifier $(w, b)$ which correctly classifies the dataset after the removal of $e_{i_1}, \dots, e_{i_k}$. Therefore, we simultaneously have:
        \begin{equation*}
            \forall i \notin L: \ip{e_i}{w} \geq b 
        \end{equation*}
        for the positive labels and
        \begin{equation*}
            \ip{\frac{1}{d - k} \sum_{i \notin L} e_i}{w} < b
        \end{equation*}
        for a one of the negative data points in the dataset. This is a contradiction which establishes the first sub-claim. For the second sub-claim, consider the linear classifier defined by $w, b$:
        \begin{equation*}
            w \coloneqq \sum_{i \notin L} e_i \text{ and } b = 1 - \frac{1}{2(d - k)}. 
        \end{equation*}
        We have for any $i \notin L$:
        \begin{equation*}
            \ip{w}{e_i} = 1 > b.
        \end{equation*}
        Now, we have for any $L' = \{i'_1, \dots, i'_k\} \neq L$:
        \begin{equation*}
            \ip{w}{\frac{1}{d - k} \sum_{i \notin L'} e_i} = 1 - \frac{\abs{L' \setminus L}}{d - k} < b.
        \end{equation*}
        This proves the second sub-claim.
    \end{proof}

    Let by $S$ be a uniformly random subset of $ \mc{I} $ (represented in binary).
    Let $D_S$ denote the dataset corresponding to $S$.
    Note that \cref{claim:critic_set_ident}, if the answer of the unlearning query was exactly correct then 
    \begin{equation*}
        \lbrb{L = \lbrb{i_1, \dots, i_k} \in \mc{I}: \unlearn \lprp{\lbrb{e_{i_1}, \dots, e_{i_k}}, p_S} = \no{}} = S.
    \end{equation*}
    Accounting for errors, we have that 
    \begin{align*}
        \Pr[  L \in S \mid{} \unlearn( L , \learn(D_S)  ) = \no  ] \geq 1 - \delta
    \end{align*}
    This implies
    \begin{align*}
        \Ent(Z_L \mid{} \unlearn( L , \learn(D_S)  )) \leq 1 -  \Ent(\delta)
    \end{align*}
    As in the proof of \cref{theorem:VClb}, we have that the 
    \begin{align*}
        |   \learn(D) | \ge (1 - \Ent(\delta)) \mc{I}   
    \end{align*}
    The lower bound follows noticing that $| \mc{I} | = \binom{d}{k}$.
\end{proof}

\subsection{Proof of \pref{thm:distributional_halfspace}} 

\begin{proof}[{Proof of \pref{thm:distributional_halfspace}}] \textbf{Notation.} Fix $2 \le k' \le d - 2$ and let $\rho_1$ denote the uniform distribution on the vertices of the simplex, $\{ \textbf{0} \} \cup \{ e_i \}_{i=1}^d$, and let $\rho_2$ denote the uniform distribution on the centroids of the $(d-k')$ faces of the simplex. Namely, points of the form $\frac{1}{d-k'} \sum_{i \notin L} e_{i}$ for $L = (i_1,\cdots,i_{k'})$ sampled without replacement from $[d]$. 

\medskip
In order to prove \pref{thm:distributional_halfspace}, we will first construct two auxiliary distributions over datasets, $\mathcal{D}^0_n$ and $\mathcal{D}^1_n$ which are parameterized by $n$ (but not in the way that datasets in its support are not constrained to have at most $n$ points almost surely). In particular the number of points a dataset drawn from $\mathcal{D}_n^0$, $\mathcal{D}_n^1$ will be distributed as $N \sim \Poisson (n/2)$ for $n = \binom{d}{k'}$. The Poisson sampling model will allow the proof to be simpler, by making certain random variables which will appear in the proof independent (specifically related to the counts of the number of occurrences of a given point in the support of the random dataset). Moreover, as we will later argue, the size of datasets drawn from these distributions are bounded by $n$ with high probability.

\paragraph{Construction of $\mathcal{D}_n^0$.} Sample $N \sim \Poisson (n/2)$. Then, $N$ points are sampled from the mixture distribution $\rho = p \cdot \rho_1 + (1-p) \cdot \rho_2$ where $p = 3\log (d) / \binom{d}{k'}$. The label distribution is as follows: for $x \sim \rho_1$, the label of the point is $1$ with probability $1$ and for $x \sim \rho_2$, the label of the point is $0$ with probability $1$. Note that this corresponds to a legal distribution since the supports of $\rho_1$ and $\rho_2$ are disjoint. Let $\vD^0$ be a generic random variable over datasets induced by the above sampling process.

\medskip
Note that the number of points at the centroid of any particular $(d-k')$-face of the simplex is marginally distributed as $\Poisson ( (n/2) (1-p)/\binom{d}{k'}) \equiv \Poisson ( (1-p)/2)$. More generally, the composition of the dataset can be computed using a balls in bins approach with $B = \binom{d}{k'}+d$ bins. The first $d$ bins correspond to points on the simplex and with probability $p$ one such bin is chosen uniformly at random and a ball is dropped into it, corresponding to a point on the simplex with label $1$. The last $\binom{d}{k'}$ bins correspond to the $(d-k')$-faces of the simplex, and with probability $1-p$, a uniformly random one out of these bins is chosen and a ball is dropped into it, corresponding to a point labeled $0$ at the centroid of that face. Define the event $\mathcal{E}$ that all of the following occur,
\begin{enumerate}[label=\(\bullet\)]
    \item every point in the support of $\rho_1$ appears at least once and at most $10 \log(d)$ times in $\vD$, and,
    \item every point in the support of $\rho_2$ appears at most $\log \binom{d}{k'} + \log(d^3) \le (k'+3) \log (d)$ times in $\vD$.
\end{enumerate}
Since $p = 3\log(d)/\binom{d}{k'}$, by union bounding, $\mathcal{E}$ occurs with probability at least $1 - O(1/d^3)$. We will also let $X_i$ denote the number of balls dropped into bin $i \in [B]$, the collection of which is a sufficient statistic for $\vD^0$.

\begin{claim} \label{claim:F3}
$\Pr (\mathcal{E}) \ge 1 - O (1/d^3)$.
\end{claim}
\begin{proof}
This is a direct consequence of the exponential tails of Poisson random variables, and a union bound.
\end{proof}

\paragraph{Construction of $\mathcal{D}_n^1$.} This distribution over datasets is defined as $\mathcal{D}_n^0 | \mathcal{E}$. We will denote a generic dataset drawn from this distribution as $\vD^1$. We have the following claim on the size of the dataset $\vD^1$.

\begin{claim} \label{claim:F4}
$\Pr (|\vD^1| \ge n) \le e^{-\Omega(n)} = O(1/d^3)$.
\end{claim}
\begin{proof}
First observe that since $|\vD^0| \sim \Poisson (n/2)$, by Poisson tail bounds we have that,
\begin{align*}
    \Pr (|\vD^0| \ge n ) \le e^{-\Omega(n)} \le O(1/d^3).
\end{align*}
Note that since $\mathcal{E}$ occurs with probability at least $1 - O (1/d^3)$, there exists a coupling $\Pi$ between $\vD^0$ and $\vD^1$ such that $\Pr_{\Pi} (\vD^0 = \vD^1) = 1 - O(1/d^3)$. By union bounding,
\begin{align*}
    \Pr_\Pi (|\vD^1| \le n) &\ge \Pr_\Pi (|\vD^1| = |\vD^0| \text{ and } |\vD^0| \le n) \\
    \implies \Pr_\Pi (|\vD^1| \ge n) &\le \Pr_\Pi (|\vD^1| \ne |\vD^0| \text{ or } |\vD^0| \ge n) \\
    &\le \Pr_\Pi (|\vD^1| \ne |\vD^0| ) + \Pr (|\vD^0| \ge n) \\
    &\le O(1/d^3).
\end{align*}
\end{proof}

\paragraph{Construction of final dataset distribution $\mathcal{D}_n$.} Note that $\mathcal{D}_n^0$ and $\mathcal{D}_n^1$ are defined purely for the purpose of the proof. The terminal distribution over datasets we will consider in the proof of \Cref{thm:distributional_halfspace} is simply the dataset constructed by drawing $n$ points independently from $\rho = p \cdot \rho_1 + (1-p) \cdot \rho_2$. We will denote a dataset drawn from this distribution as $\vD$. Note that $\mathcal{D}_n$ is a distribution on datasets supported on $n$ points.

\medskip
Next we will show that an LU scheme under the final distribution $\mathcal{D}_n$ can be translated to an LU scheme with essentially the same space complexity on $\mathcal{D}_n^1$.

\begin{claim} \label{claim:F5}
Consider any $(\varepsilon,\delta)$-\LU scheme $(\learn,\unlearn)$ with auxiliary memory $\aux$ for the task of $\mcH$-realizability testing: for each $n_1 > 0$ suppose the space complexity of the \LU scheme against the distribution over datasets $\mathcal{D}_{n_1}$ is,
\begin{align*}
    \mathbb{E} [ |\aux (\vD)| \}] \le f(n_1,k)
\end{align*}
supporting unlearning queries of size $k$. Then, for some value of $n_0 \in [n/2,n]$, there exists another $(\varepsilon,\delta)$-\LU scheme, $(\widetilde{\learn},\widetilde{\unlearn})$ with auxiliary memory $\widetilde{\aux}$ for the same task against the distribution over datasets $\mathcal{D}^1_n$ satisfying,
\begin{align*}
    \mathbb{E} [ |\widetilde{\aux} (\vD^1)| \}] \le f(n_0,k) + 1 + O(n\log(d)/d^2)
\end{align*}
which also supports unlearning queries of size at most $k$.
\end{claim}
\begin{proof}
The scheme is as follows,
\begin{equation*}
    \widetilde{\aux}(\vD^1) = \begin{cases}
        ( \aux(\vD^1) , 0 ) \qquad &\text{if } |\vD^1| \le n \text{ and } |\aux(\vD^1)| \le |\vD^1|\\
        ( \vD^1 , 1 ) &\text{otherwise.}
    \end{cases}
\end{equation*}
The final bit $\in \{ 0,1\}$ is simply there to indicate to the unlearning scheme which case we are in. 


\medskip
While this specifies the usage of auxiliary memory, the corresponding $\widetilde{\learn}$ and $\widetilde{\unlearn}$ subroutines are those induced naturally. In the former case, we instantiate them as $\learn (\vD^1)$ and $\unlearn ( \cdot,\aux (\vD^1))$ respectively. In the latter case, we may carry out exact unlearning since the entire dataset is present in auxiliary memory. The space complexity of this scheme is bounded as follows,
\begin{align}
    \mathbb{E} \big[ |\widetilde{\aux} (\vD^1)| ]
    &\le \mathbb{E} [ |\widetilde{\aux} (\vD^1)| \cdot \mathbb{I} ( |\vD^1| \le n) \big] + \mathbb{E} [ |\widetilde{\aux} (\vD^1)| \ \big| \ |\vD^1| \ge n] \Pr (|\vD^1| \ge n) \nonumber\\
    &= \mathbb{E} [ (\min \big\{ |\aux (\vD^1)|, |\vD^1| \big\} + 1) \cdot \mathbb{I} ( |\vD^1| \le n) \big] + \mathbb{E} [ (|\vD^1| + 1) \ \big| \ |\vD^1| \ge n] \Pr (|\vD^1| \ge n) \nonumber\\
    &= \sum_{n'=1}^n \mathbb{E} [\min \big\{ |\aux (\vD^1)|, n' \big\} \cdot \mathbb{I} ( |\vD^1| = n') \big] + 1 + O(1/d^3) \cdot \mathbb{E} [ |\vD^1| \ \big| \ |\vD^1| \ge n] \nonumber\\
    &\le \sum_{n'=1}^n \mathbb{E} [ \min \big\{ |\aux (\vD^1)|, n' \big\} \cdot \mathbb{I} ( |\vD^1| = n') \big] + 1 + O(n \log(d)/d^2). \label{eq:0001}
\end{align}
The last inequality uses the fact that $|\vD^1|$ (by virtue of conditioning on $\mathcal{E}$) has at most $10 (d + 1)\log(d) + (k'+2) \log(d) \binom{d}{k'} \le O(n d \log (d))$ datapoints in it almost surely. Recall from the proof of \pref{claim:F4} the existence of a coupling $\Pi$ between $\vD^0$ and $\vD^1$ such that $\Pr_\Pi (\vD^0 = \vD^1) = 1 - O(1/d^3)$. This implies,
\begin{align*}
    &\sum_{n'=1}^n \mathbb{E} [ \min \big\{ |\aux (\vD^1)|, n' \big\} \cdot \mathbb{I} ( |\vD^1| = n') \big] \\
    &\le \sum_{n'=1}^n \big( \mathbb{E}_\Pi [ \min \big\{ |\aux (\vD^1)|, n' \big\} \cdot \mathbb{I} ( |\vD^1| = n', \vD^0 = \vD^1) \big] + \mathbb{E}_\Pi [ n' \cdot \mathbb{I} ( |\vD^0| = n', \vD^0 \ne \vD^1) \big] \big) \\
    &\le \sum_{n'=1}^n \big(\mathbb{E} [ \min \big\{ |\aux (\vD^0)|, n' \big\} \cdot \mathbb{I} ( |\vD^0| = n') \big] + \mathbb{E}_\Pi [ n' \cdot \mathbb{I} ( |\vD^0| = n', \vD^0 \ne \vD^1) \big] \big)
\end{align*}
With this,
\begin{align*}
    &\sum_{n'=1}^n \mathbb{E} [ \min \big\{ |\aux (\vD^1)|, n' \big\} \cdot \mathbb{I} ( |\vD^1| = n') \big] \\
    &\le \sum_{n'=1}^n \big( \mathbb{E} [ \min \big\{ |\aux (\vD^0)|, n' \big\} \cdot \mathbb{I} ( |\vD^0| = n') \big] + \mathbb{E}_\Pi [ n' \cdot \mathbb{I} ( |\vD^0| = n', \vD^0 \ne \vD^1) \big] \big)\\
    &\le \sum_{n'=1}^n \mathbb{E} [ \min \big\{ |\aux (\vD^0)|, n' \big\} \cdot \mathbb{I} ( |\vD^0| = n') \big] + n \cdot \Pr_\Pi ( \vD^0 \ne \vD^1 )\\
    &\le \sum_{n'=1}^n \mathbb{E} [ \min \big\{ |\aux (\vD^0)|, n' \big\}  \ \big| \ |\vD^0| = n' \big] \cdot \Pr (|\vD^0| = n') + O(n/d^3) \\
    &\le \sum_{n'=n/2}^n \mathbb{E} [ \min \big\{ |\aux (\vD^0)|, n' \big\}  \ \big| \ |\vD^0| = n' \big] \cdot \Pr (|\vD^0| = n') + n e^{-\Omega(n)} + O(n/d^3)
\end{align*}
where the last inequality uses the fact that $\Pr (|\vD^1| \le n')$ for $n' \le n/2$ is at most $e^{-\Omega(n)}$. Let $n_0$ denote the maximizer of $\mathbb{E} [ |\aux (\vD^0)| \ \big| \ |\vD^0| = n' \big]$ over $n' \in [n/2,n]$ for a small constant to be determined later. Then, we have the upper bound,
\begin{align*}
    \sum_{n'=1}^n \mathbb{E} [ \min \big\{ |\aux (\vD^1)|, n' \big\} \cdot \mathbb{I} ( |\vD^1| = n') \big] &\le \mathbb{E} [ |\aux (\vD^0)| \ \big| \ |\vD^0| = n_0 \big] + O(n/d^3) \\
    &\le \mathbb{E} [ |\aux (\vD^0)| \ \big| \ |\vD^0| = n_0 \big] + O(n/d^3) \\
    &\le f(n_0,k) + O(n/d^3).
\end{align*}
where the last inequality uses the fact that $\vD^0 \big| |\vD^0| = n_0$ is precisely distributed as $\vD \sim \mathcal{D}_{n_0}$ and the space complexity bound on the black-box \LU scheme. Combining with \pref{eq:0001} completes the proof.
\end{proof}

With this claim it suffices to argue space complexity lower bounds for $(\varepsilon,\delta)$-LU schemes for the task of $\mcH$-realizability testing against the distribution $\mathcal{D}_{n_0}^1$. We will precisely establish this next.

\begin{claim} \label{claim:F6}
Any $(\varepsilon,\delta)$-\LU scheme $(\widetilde{\learn},\widetilde{\unlearn})$ for $\mcH$-realizability testing against the distribution $\mathcal{D}_n^1$ satisfies,
\begin{align*}
    \mathbb{E} [ |\widetilde{\aux} (\vD^1)| ] \ge n \cdot ( c_0 - \Ent (\delta)) - O(d \log \log (d))
\end{align*}
\end{claim}
\begin{proof}
Recall that for the learning task of realizability testing, $\cW=\{\yes,\no\}$ and $\learn(D)$ always outputs $\yes$ if $D$ is $\mcH$-realizable and $\no$ otherwise. Let $\vA$ be a random variable for the auxiliary information outputted by $\learn$ algorithm on dataset $\vD^1 \sim \mathcal{D}_n^1$. In the distributional setting (\Cref{def:LU-dist}), given a realization of $\vD^1 = \{ (x_i,y_i) \}_{i=1}^{N}$ and subsets $I \subseteq [N]$ if $\vU_I = \{ (x_i,y_i) : i \in  I \}$ is realizable, then
\begin{align} \label{eq:l1}
\Pr[\unlearn(\vU_I,\aux(\vD^1))=\no] \le e^{\veps} \cdot 0 +\delta = \delta.
\end{align}
Similarly, if $\vU_I$ is not realizable, then
\begin{align} \label{eq:ul1}
\Pr[\unlearn(\vU_I,\aux(\vD^1))=\yes]\le \delta.
\end{align}
Now, conditioned on $\vD^1$, consider the unlearning query denoted $\vD^1_L$ which corresponds to choosing a subset of $k'$ indices, $L = (i_1,\dots,i_{k'}) \in \binom{[d]}{k'}$ and deleting $X_i$ copies of $(e_i,1)$ for each $i \in L$. 
The overall size of the unlearning query is at most $k = 10 k' \log (d)$ points.

Consider the point,
\begin{equation*}
    z_L = \lprp{x = \frac{1}{d - k} \sum_{j \notin L} e_{j}, y = 0}
\end{equation*}
Note that as in the proof of \Cref{claim:critic_set_ident}, the dataset (upon making the unlearning query corresponding to points $L$) is linearly separable if and only if the dataset $\vD^1$ contains no copies of the above point; i.e., $N_{z_L} = 0$ (the number of occurrences of $z_L$ in $\vD^1$). By the approximate correctness of the unlearning scheme (c.f.~ eqs.~\pref{eq:l1},\pref{eq:ul1}), and Bayes rule, the probability that $\mathbb{I} (N_{z_L} = 0) = \vZ_L = 0$ (or $\vZ_L = 1$) conditioned on $\unlearn(\vD^1_L,\aux(\vD^1))=\no$ (or $\yes$) is greater than $1-\delta$. Thus, for $\delta<1/2$,
\begin{align}
&\Ent\left( \vZ_L \mid\unlearn(\vD^1_L,\vA(\vD^1))\right)\le \Ent(1-\delta) \nonumber\\
\implies &\I( \vZ_L ; \unlearn(\vD^1_L,\vA(\vD^1)) \ge c_0 -\Ent(1-\delta)).\label{eq:El1-dist}
\end{align}
where the last implication uses \pref{claim:F5} which lower bounds $\Ent(\vZ_L)$.

\medskip
We will use this to show that $\mathbb{E}[\vA (\vD^1)] \ge \Omega(n)$. Consider the collection of subsets $\vL = \binom{[d]}{k'}$, and define an arbitrary canonical total ordering $\{ >,< \}$ of these subsets. Let $\vZ = \{ \vZ_L : L \in \vL \}$ and $\vZ_{> L} = \{ \vZ_L : L' > L  \}$. For brevity we will drop the argument and let $\aux \equiv \aux(\vD^1)$. Furthermore, let $\vN = \{ N_{(x,1)} : x \in \{ \textbf{0}, e_1,\cdots,e_d \} \}$ count the number of occurrences of points in the support of $\rho_1$. Then,
\begin{align*}
|\vA| &\ge \I(\vZ;\vA) \tag{as $\Ent(\vA)\le |\vA|$}\\
&=\sum_{L \in \vL} \I( \vZ_L ;\vA \mid \vZ_{>L})\tag{Chain rule}\\
&=\sum_{L \in \vL} \left(\I(\vZ_L;\vA,\unlearn(\vD^1_L,\vA)\mid \vZ_{>L})-\I(\vZ_L;\unlearn(\vD^1_L,\vA)\mid \vA,\vZ_{>L})\right)\\
&\ge \sum_{L \in \vL} \left(\I(\vZ_L;\unlearn(\vD_L^1,\vA)\mid \vZ_{>L})-\I(\vZ_L;\unlearn(\vD_L^1,\vA)\mid \vA,\vZ_{>L})\right)\\
&= \sum_{L \in \vL} \I(\vZ_L;\unlearn(\vD_L^1,\vA)\mid \vZ_{>L}) - O(d \log \log (d)) \tag{Explained below}\\
&\ge \sum_{L \in \vL} \I(\vZ_L;\unlearn(\vD_L^1,\vA)) - O(d \log \log (d)) \tag{$\vZ_L \perp \vZ_{> L}$ by Poisson independence}\\
&\ge \sum_{L \in \vL} (c_0-\Ent(1-\delta)) - O(d \log \log (d)) \tag{Equation \eqref{eq:El1-dist}}\\
&=n \cdot (c_0-\Ent(\delta)) - O(d \log \log (d))
\end{align*}
The second last equality follows from the fact that $\I(\vZ_L;\unlearn(\vD_L^1,\vA)\mid \vA,\vZ_{>L})\le O(d \log \log (d))$. This is by the following chain of arguments,
\begin{align*}
\I(\vZ_L;\unlearn(\vD_L^1,\vA)\mid \vA,\vZ_{>L}) &\le \I(\vZ_L;\unlearn(\vD_L^1,\vA)\mid \vA, \vN,\vZ_{>L}) + \Ent(\vN) \tag{Chain rule}\\
&\le O ( d \log\log(d))
\end{align*}
The last inequality is because of two facts, $(i)$ $\vN$ encodes $d+1$ random variables which take values in the range $\{ 1 ,\dots, 10\log (d) \}$ (cf. the event $\mathcal{E}$), implying that $\Ent (\vN) \le O(d \log \log (d))$, and $(ii)$ the unlearning query $\vD_L^1$ is deterministic given $\vN$ and the output of the $\unlearn$ algorithm on this query is completely determined by $\vA$ and the random bits used by the $\unlearn$ algorithm, which are independent of $\vZ_{\ge L}$ (and $\vA$). 
\end{proof}

\begin{claim} \label{claim:F7}
There exists an absolute constant $c_0 > 0$ such that for any $L$, $\Ent(\vZ_L) \ge c_0$.
\end{claim}
\begin{proof}
Note by Poisson independence, $X_i$ which counts the number of balls in the $i^{\text{th}}$ bin (i.e., number of occurrences of a datapoint in $\vD^0$) are independent across different values of $i$. Since $\mathcal{E}$ is a ``rectangular'' event (each $X_i$ is constrained to fall in some interval) this implies that $\vZ_L = \mathbb{I} (N_{z_L} = 0)$ is distributed as $\mathbb{I} (X_i = 0)$ conditioned on $ X_i \le \log \binom{d}{k'} + \log (d^3)$ where $i$ is the index of the point $z_L$. Since $X_i \sim \Poisson ((1-p) / 2)$,
\begin{align*}
    \Pr (X_i = 0) = 1 - e^{- (1-p)/2} \in \big[ 1-e^{-1/4}, 1-e^{-1/2} \big],
\end{align*}
which assumes that $d$ is sufficiently large that $3 \log (d) / \binom{d}{k'} \le \frac{1}{2}$. Note that $\Pr (X_i \le \log \binom{d}{k'} + \log (d^3)) \ge 1 - O(1/d^2)$. This implies that,
\begin{align*}
    1 - e^{-1/4} \le \Pr (X_i = 0) \le \Pr (N_{z_L} = 0) \le \frac{\Pr (X_i = 0)}{\Pr (X_i \le \log\binom{d}{k'} + \log (d^3))} \le \frac{1-e^{-1/2}}{1 - O(1/d^2)}
\end{align*}
Since the LHS is bounded away from $0$ and the RHS is bounded away from $1$, the entropy of the indicator random variable $\vZ_L = \mathbb{I} (N_{z_L} \ge 0)$ is at least some absolute constant $c_0$.
\end{proof}

\paragraph{Proof of \pref{thm:distributional_halfspace}.} Combining \pref{claim:F5} and \pref{claim:F6}, for any $(\varepsilon,\delta)$-\LU scheme $(\learn,\unlearn)$ with auxiliary memory $\aux$ for the task of $\mcH$-realizability testing against the distribution over datasets $\vD \sim \mathcal{D}_{n_0}$,
\begin{align*}
    \mathbb{E} [ |\aux (\vD)| \}] \ge n (c_0 - \Ent(\delta)) - O(n \log (d)/d^2) + d\log\log(d))
\end{align*}
for some $n_0 \in [n/2,n]$. Note that $n \ge \Omega(d^2)$ assuming that $2 \le k' \le d-2$ and therefore, assuming $d$ is at least a sufficiently large constant, we have that,
\begin{align*}
    \mathbb{E} [ |\aux (\vD)| \}] \ge n (c_1 - \Ent(\delta))
\end{align*}
For a different absolute constant $c_1 > 0$. This completes the proof.
\end{proof}

\section{From Realizability Testing to Unlearning}
\label{app:reduction} 

In this section, we discuss the connection of our results on \LU / \TiLU schemes for realizability testing, to designing unlearning schemes for ERM under unbounded deletions. We focus on $(0,0)$-\LU and \((0, 0)\)-$\TiLU$ schemes in this section, but the lower bounds hold for any constant $\veps\in[0,1]$ and $\delta\in[0,1/2)$. 

\begin{definition}[Empirical risk minimization] Let $\mathcal{H}$ be a learning task. The learning task of empirical risk minimization operates on the range space $\cW = \cH$ and is defined as,
\begin{align}
    f(D) = \argmin_{h \in \cH} \sum_{(x,y) \in D} \mathbb{I} (h(x) \ne y).
\end{align}
Ties in the $\argmin ( \cdot )$ may be resolved arbitrarily.
\end{definition}

The space complexity of \LU schemes for \ERM are denoted by $\s_{\ERM} (\cdot)$ to make the distinction with the notation for realizability testing clear.

\medskip

In the case where the underlying dataset $D$ is $\mathcal{H}$-realizable, the existence of \LU / \TiLU schemes for realizability testing is trivial since any subset of $D$ is always going to remain $\mathcal{H}$-realizable. However, under the same constraint, the existence of \LU / \TiLU schemes for ERM is non-trivial. Indeed, the challenge arises from the fact that correctness of the scheme requires that $\unlearn (D_I, \aux)$ (with $\aux$ computed on the input dataset $D$) is required to match with the hypothesis that would have been returned by $\learn (D \setminus D_I)$. Since the unlearning query $D_I$ is arbitrary, this is a non-trivial requirement - the unlearning scheme has to output the same hypothesis whether (i) the dataset $D$ was learned and $D_I$ was subsequently unlearned, or (ii) whether the smaller dataset $D \setminus D_I$ was directly learned. The key technical challenge in proving the correctness of any such unlearning scheme is to exhibit that the hypothesis output after unlearning a subset of points is ``canonical'' in a sense.

\medskip
In the case where $D$ is $\mathcal{H}$-realizable to begin with, we show that it is possible to devise an ticketed unlearning scheme for \ERM with space complexity scaling with the star number  of $\mathcal{H}$ and logarithmically in the size of $D$. We follow a similar approach as the case of realizability testing, and introduce the notion of a mergeable concept class for \ERM to build up to the main result.

\begin{definition}[{\boldmath Mergeable Concept Class for ERM, \citep{ghazi2023ticketed}}]\label{def:mergeable_ERM}
A concept class $\cH \subseteq \crl{\cX \to \cY}$ is said to be \emph{$C$-bit mergeable} if there exist methods
$\Encode, \Merge, \Decode$ such that
\begin{itemize}
\item $\Encode: \cZ^* \to \bit^C$ is a permutation-invariant encoding of its input into $C$ bits. 
\item $\Decode: \bit^C \to \cH$ such that $\Decode(\Encode(S)) \in \ERM_{\cH}(S)$ for all $\cH$-realizable $S \in \cZ^*$.
\item $\Merge: \bit^C \times \bit^C \to \bit^C$ such that for all $S_1, S_2 \in \cZ^{*}$ such that $S_1 \cup S_2$\footnote{We use $S_1 \cup S_2$ to denote the {\em concatenation} of the two datasets.} is $\cH$-realizable, it holds that $\Encode(S_1 \cup S_2) = \Merge(\Encode(S_1), \Encode(S_2))$.%
\end{itemize}
\end{definition}

The following result by \cite{ghazi2023ticketed} shows that any \(C\)-bit mergeable hypothesis class for ERM has a \TiLU scheme with bounded memory for realizable datasets. 

\begin{theorem}[{\cite{ghazi2023ticketed}}]
\label{thm:merkle_ERM} 
For any $C$-bit mergeable concept class $\cH$, there exists a TiLU scheme for ERM with both the ticket size \( O(C \log n)\) and auxiliary number of bits \(\log |\cH|\).  
\end{theorem}

\begin{theorem}[Realizable unlearning upper bound under bounded star number]  
let \(\cH\) be a hypothesis class with star number \(\starno(\cH)\). Then, \(\cH\) has a \TiLU scheme for ERM with both ticket size bounded by \(O(\starno(\cH) \log(n))\) and the  number of auxiliary bits bounded by \(\log(\cH)\).   
\end{theorem} 
\begin{proof} We first prove that the hypothesis class \(\cH\) is \(O(\Eld(\cH) \log(\abs{\cZ})\)-bit Mergeable; this part will be similar to the proof of \pref{thm:eludercompression}. We first define additional notation. Recall that a set of hypothesis \(\cH' \subseteq \cH\) is said to be a version space if there exists some dataset \(\bar{\cS} \in \cZ^*\)  such that \(\cH' = \cH(\bar{\cS})\). Next, define a function  \(\canonical: 2^{\cH} \mapsto \cZ*\) that maps subsets \(\cH' \subseteq \cH\) to datasets \(\cS'\) as follows: On input \(\cH'\), 
\begin{itemize}
    \item 
Initialize \(\cS' = \emptyset\), and for \(x \in \sorted(\cX)\):  
\begin{itemize}[label=\(\bullet\)] 
\item Update \(\cS' = \cS' \cup \crl{(x, y)}\) if \(h'(x) = y\) for all \(h' \in \cH'\) but there exists an \(h \in \cH\) such that \(h(x) \neq y\). Else, continue to the next sample. 
\end{itemize} 
\item Prune \(S'\) to remove all the samples that do not change the version space \(\cH(S')\), similar to the pruning in the proof of \pref{thm:eludercompression}. 
\item Return \(\canonical(\cH') = \cS'\). 
\end{itemize}

\begin{claim} Let \(\cS'=\canonical(\cH(\cS))\), we have \(\cH(\cS) = \cH(\cS')\).
\end{claim}
\begin{proof} The proof follows trivially from the definition of \(\canonical\). 
\end{proof}

\begin{claim} \label{claim:eluder_bound} \(\abs{\cS'} \leq \starno(\cH)\)
\end{claim}
\begin{proof}
The sequence of points \(\cS'\) form a star set w.r.t.~\(\cH\) and thus we must have that \(\abs{\cS'} \leq \Eld(\cH)\).   
\end{proof}

Finally, we define the \(\ERM\) rule such that for any dataset \(\cS\), \(h = \ERM(\cS)\) denotes the lexicographically smallest hypothesis in the version space \(\cH(\cS)\). 

\noindent 
We are now ready to provide \(\Encode\), \(\Decode\) and \(\Merge\) methods: 

\paragraph{$\Encode$:}  
Given an input \(\cS\) (that is realizable via the hypothesis class \(\cH\)), compute the version space \(\cH' = \crl{h \in \cH \mid h(x) = y~\text{for all}~(x, y) \in \cS}\), and let  \(\cS'=\canonical(\cH')\). Return \(\Encode(\cS) = \cS'\). 

\paragraph{\(\Decode\):} Given an encoding \(\cS'\),  compute the version space \(\cH' \leftarrow \cH(\cS')\), and return the lexicographically smallest hypothesis \(h'\) in \(\cH'\). 

\paragraph{\Merge:} Given encodings \(\cS'_1\) and \(\cS'_2\), compute the version spaces  \(\cH_1 \leftarrow  \cH(\cS'_1)\) and \(\cH_2 \leftarrow \cH(\cS'_2)\), and define \(\cH'' = \cH_1 \cap \cH_2\). Compute the dataset \(\cS'' = \canonical(\cH'')\) and return \(\Merge(S'_1, S'_2) = \cS''\).

\paragraph{\(\mb{\cH}\) is \(\mb{O(\starno(\cH))\log(\abs{\cZ})}\)-bit mergeable for ERM.} Let \(\cS'\) be the output of \(\Encode\) for a dataset \(\cS\). By definition of \(\canonical\), we have that \(\cH(\cS') = \cH' = \cH(\cS)\). Thus, \(\Decode(\Encode(\cS)) = \ERM(\cS)\).

We next establish the mergeability property. Note that for any sets \(\cS_1\) and \(\cS_2\), the output of \(\Encode(\cS_1 \cup \cS_2) = \cS'\)  where \(\cS' = \canonical(\cH(\cS_1 \cup \cS_2))\). On the other hand, we compute \(\Merge(\Encode(\cS_1), \Encode(\cS_2)) = \cS''\)   where \(\cS'' = \canonical(\cH(\cS_1) \cap \cH(\cS_1))\). However, since \(\cH(\cS_1) \cap \cH(\cS_1) = \cH(\cS_1 \cup \cS_2)\) by definition of the version spaces, we have that \(\cS' = \cS''\) and thus \(\Merge(\Encode(\cS_1), \Encode(\cS_2)) = \Encode(\cS_1 \cup \cS_2)\). 

Since all generated encodings have at most \(\Eld(\cH)\)-samples (due to Claim \ref{claim:eluder_bound}), we immediately get that \(\cH\) is \(\Eld(\cH) \log(\abs{\cZ})\)-bit mergeable. 
\end{proof}

So far, we have considered the realizable setting for unlearning for ERM. Unfortunately, there are simple examples where if the initial dataset is not realizable, then there cannot exist any LU scheme with sublinear memory.

\begin{remark}[{\citet[Theorem 24]{ghazi2023ticketed}}] \label{rem:24ghazi} Consider the class of \(1\)-d thresholds over the set \(\cX = \crl{1, 2, \dots, \abs{\cX}}\). Consider any \LU scheme for \ERM for this class that works with arbitrary datasets of size at most $n$. Then, there exists a worst-case dataset such that the space complexity of the \LU scheme is \(\Omega(\min\crl{n, \abs{\cZ}})\) bits. Furthermore, this lower bound holds even if we only consider unlearning queries of constant size.
\end{remark}

\subsection{A Reduction from Unlearning Schemes for ERM to Realizability Testing} 

While \Cref{rem:24ghazi} shows that \LU schemes for \ERM only exist in very restricted settings, the implications for \TiLU schemes are less clear. In this section we discuss a white-box reduction to convert space complexity lower bounds for unlearning schemes for realizability testing to corresponding lower bounds for unlearning schemes for ERM. At an intuitive level, it may seem that outputting the \ERM of a dataset after having unlearning a subset of the points is a harder task than simply checking whether the dataset is $\mathcal{H}$-realizable or not. While this intuition is certainly true if the input dataset $D$ is $\mathcal{H}$-realizable to begin with, and may also be true for many common hypothesis classes even ignoring this assumption, there are cases which this intuition breaks down. For instance, for the singleton hypothesis class $\mathcal{H} = \{ h \}$, computing the \ERM under unlearning queries can be done by an unlearning scheme with $0$ space complexity; realizability testing on the other hand is a function of the labels of the datapoints and requires at least $1$ bit of space. In spite of this pathological case, we show a way to reduce space complexity lower bounds on unlearning schemes for realizability testing to those for \ERM that is applicable for a wide variety of hypothesis classes. We first begin by describing the general method we use to derive lower bounds for the former.

\subsubsection{Unifying Lower Bounds for Unlearning Schemes for Realizability Testing} \label{sec:unifying}
The lower bounds we prove in \Cref{theorem:VClb} and \Cref{thm:lbeluder} follow a common recipe which we transcribe below. The lower bounds rely on constructing a collection of datasets $\mathcal{D} = \{ D(z) : z \in \{ 0,1 \}^p \}$ indexed by $p$-bit binary strings where $D(z) = D (\emptyset) \cup \{ (x_i, y_i) : i \in [p],\ z_i = 1 \}$ for some base dataset $D (\emptyset)$ and $(x_i,y_i) \in \mathcal{X} \times \{ 0,1 \}$. The datasets in $\mathcal{D}$ satisfy the following structure: for each $i \in [p]$, there exists a subset of points $U_i \subseteq D(\emptyset)$, such that for any $z \in \{ 0,1 \}^d$, the dataset $D(z) \setminus \left( \{ (x_j,y_j) : j \le i-1\} \cup U_i \right)$ is $\mathcal{H}$-separable if and only if $z_i = 0$. The size of datasets in $\mathcal{D}$ are upper bounded by $|D(\emptyset)| + p$. 

The lower bound on the space complexity of \LU schemes for realizability testing follow by arguing that by following the unlearning queries $\{ (x_j,y_j) : j \le i-1\} \cup U_i$ for $i=1,2,\cdots,p$, the learner can recover the bits $z_1,\cdots,z_p$ sequentially. For $n = |D(\emptyset)| + p$, this results in the lower bound,
\begin{align} \label{eq:s(n)-lb}
    \s (n) \ge p,
\end{align}
for \LU schemes for realizability testing. Likewise, for \TiLU schemes for realizability testing, we arrive at the lower bound,
\begin{align} \label{eq:s(n)-lb-tx}
    \s (n) \ge \frac{p}{| \cup_i U_i |},
\end{align}
We now discuss how to lift such lower bound instances for the case of unlearning for ERM. We will make the reduction explicit for \LU schemes; the same arguments also result in lower bounds for \TiLU schemes for \ERM.

\subsubsection{The White-Box Reduction} \label{sec:white-box}

Recall the structural assumption we make on the lower bound instances for unlearning for realizability testing:  for each $i \in [p]$, there exists a fixed unlearning query $U_i \subseteq D(\emptyset)$, such that for any $z \in \{ 0,1 \}^d$, the dataset $(D(z) \setminus \{ (x_j,y_j) : j \le i-1\} ) \setminus U_i$ is $\mathcal{H}$-separable if and only if $z_i = 0$.
From the collection of datasets $\mathcal{D}$ construct,
\begin{align} 
    \mathcal{D}' = \Big\{ D(z) \cup \{ L \times (x_i,y_i) : i \in [p],\ z_i = 1 \} : z \in \{ 0,1 \}^p,\ z_1 = 0 \Big\}
\end{align}
where the notation $L \times (x,y)$ indicates $L$ copies of the point $(x,y)$, and $L = \max_{i \ge 2} |U_i \setminus U_1|$. The datasets in $\mathcal{D}'$ have size at most $p(L+2)+|D(\emptyset)| \le (L+3)n  \le 3n^2$ and are essentially indexed by a $p-1$-bit string $(z_2,\cdots,z_p)$ since $z_1 = 0$.

Ignoring $i=1$, we will start with $i=2$ and assume that $z_2=1$. Since $z_1 = 0$, by the structural assumption on $\mathcal{D}$, the dataset $D' (z) \setminus U_1$ is realizable, and let $h \in \mathcal{H}$ be the realizing hypothesis. Therefore, on the dataset $D' (z) \setminus U_2$, the same hypothesis $h$ incurs an error of at most $|U_2 \setminus U_1| \le L$. On the other hand, note that any classifier on this dataset which classifies the point $(x_2,y_2)$ incorrectly incurs an error of at least $L+1$. These two conditions imply that when $z_2=1$, the \ERM of the dataset $D' (z) \setminus U_2$ correctly classifies $x_2$ with the label $y_2$. On the other hand, note that if $z_2 = 0$ then the \ERM of $D' (z) \setminus U_2$ must label the point $x_2$ with the label $\neg y_2$. If this weren't the case, the dataset would remain $\mathcal{H}$-realizable even upon adding $(x_2,y_2)$ to it, which contradicts the structural condition on $\mathcal{D}$. In conclusion, the \ERM of the dataset $D'(z) \setminus U_2$ evaluated on the point $x_2$ reveals whether the same dataset is $\mathcal{H}$-realizable or not. If the predicted label on $x_2$ is $y_2$, $D'(z) \setminus U_2$ is realizable, and if the predicted label is $\neg y_2$, the dataset is not realizable.

By recursing this argument, suppose $z_2,z_3,\cdots,z_i$ have been inferred using the unlearning queries made thus far. Then the learner can compute $z_{i+1}$ using a similar approach: the \ERM on the dataset $(D'(z) \setminus \{ (L+1) \times (x_j,y_j) : j \le i-1, z_i = 1 \} ) \setminus U_i$ must correctly classify the point $(x_j,y_j)$ if $z_i = 1$; on the other hand, if $z_i=1$, the \ERM on the same dataset must classify the point $x_j$ as $\neg y_j$ if $z_i = 0$. The prediction of the \ERM on the point $x_j$ predicts $z_{i+1}$. By this construction, we see that an \LU scheme for \ERM is thus able to infer the hidden string $(z_2,z_3,\cdots,z_p)$, which means that we must have $\s_{\ERM} (3 n^2) \ge p-1$. In comparison with \cref{eq:s(n)-lb}, the space complexity lower bound is now on datasets of size $O(n^2)$. While these arguments are stated for \LU schemes, extension to the ticketed case follow suite similarly.

As a result of this equivalence, we have a few corollaries in order.

\begin{corollary}
For any hypothesis class with eluder dimension $\eluder$ the space complexity of any \LU scheme for \ERM on datasets of size at most $n$ satisfies, 
\begin{align}
    \s_{\ERM} (n) = \Omega (\eluder \log (n / \eluder))
\end{align}
\end{corollary}

Note the white-box reduction we described in \Cref{sec:white-box} does not preserve the bound on size of the unlearning queries; notably, in the realizability testing lower bound (corresponding to the datasets $\mathcal{D}$), we consider unlearning queries of the form $U_1, U_2 \cup \{ (x_1,y_1) \}, U_3 \cup \{ (x_1,y_1), (x_2,y_2) \}$ and so on, while in the \ERM lower bounds (Corresponding to the datasets $\mathcal{D}'$), the unlearning queries are of the form $U_1, U_2 \cup \{ (L+1) \times (x_1,y_1) \}, U_3 \cup \{ (L+1) \times (x_1,y_1), (L+1) \times (x_2,y_2) \}$ and so on, which are roughly $L+1$ times larger. However, note that in certain special cases, such as the case of $d$-dimensional linear classifiers, the lower bounds for realizability testing only consider unlearning queries of the form $U_1,U_2,U_3,\dots$, without having to delete copies of the inferred datapoints, $\{ (x_j,y_j) : j \le i \}$ (c.f.~the proof of \Cref{thm:linear_central_lb}). In these special cases, the white-box reduction establishing space complexity lower bounds for \LU / \TiLU schemes for \ERM also consider the same kind of unlearning queries, namely, $U_1,U_2,U_3,\dots$. In these special cases, the reduction preserves the size of the unlearning queries. For \TiLU schemes under \ERM for halfspaces, we have the space complexity lower bound
\begin{align}
    \s_{\ERM} ( (L+3) n) \ge \frac{p-1}{|\cup_i U_i|}
\end{align}
In the lower bound instances we consider for halfspaces, the unlearning queries $U_i$ correspond to subsets of $k$ points drawn from $\{ (e_i,1) : i \in [d] \}$. In particular, the hidden vector is of dimension $p = \binom{d}{k}$, while $L = \max_{i \ge 2} |U_i \setminus U_1| \le \max k$, and $|\cup_i U_i| = d$. Together this results in a lower bound for the ticketed setting for ERM which closely mirrors that established for realizability testing in \Cref{thm:linear_tilu_lb}.

\begin{corollary} \label{corr:linear-ERM-lb}
Consider the hypothesis class corresponding to $d$-dimensional halfspaces on the domain $\cX$ defined in \cref{eq:linear_lb_domain} of size $d + \binom{d}{k}$. Consider any \TiLU scheme for \ERM which answers unlearning queries of size at most $k$. Then, there exists a worst-case dataset of size at most $n = (k+3) |\cX|$ such that,
\begin{align}
    \s_{\ERM} (n) = \Omega \left( \frac{1}{d} \cdot \binom{d}{k} \right).
\end{align}
\end{corollary}

\end{document}